\documentclass[11pt]{article}

\usepackage[margin=1in]{geometry}
\usepackage{amsmath,amssymb,amsthm,mathtools,bm,algorithm,algpseudocode}
\usepackage{graphicx}
\usepackage{xcolor}
\usepackage{booktabs}
\usepackage{longtable}
\usepackage{subcaption}
\usepackage{enumitem}
\usepackage{tabularx}
\usepackage{threeparttable}
\usepackage{natbib}
\usepackage{authblk}
\usepackage[colorlinks=true,linkcolor=blue,citecolor=blue,urlcolor=blue]{hyperref}

\newtheorem{example}{Example}
\newtheorem{theorem}{Theorem}
\newtheorem{lemma}[theorem]{Lemma}
\newtheorem{proposition}[theorem]{Proposition}
\newtheorem{remark}[theorem]{Remark}
\newtheorem{corollary}[theorem]{Corollary}
\newtheorem{definition}[theorem]{Definition}

\bibpunct{(}{)}{;}{a}{,}{,}
\newenvironment{keywords}{\par\noindent\textit{Keywords:}\ }{\par}

\newtheorem{assumption}{Assumption}

\def\arg{\mathop{\rm arg}}

\def\bfx{\mathop{\bf x}}
\def\bfx{\mathop{\bf X}}

\def\bfx{\boldsymbol{x}}
\def\bfd{\boldsymbol{d}}

\def\bfz{\boldsymbol{z}}
\def\bfw{\boldsymbol{w}}

\newcommand {\bfphi} {\mbox{\boldmath $\phi$}}

\title{On the Optimality of Random Features under General Conditions: Misspecified Cases and Robust Learning}
\author{Caixing Wang}
\affil{School of Statistics and Data Science\\
Southeast University\\
Nanjing, China\\
\texttt{caixingwang@seu.edu.cn}}
\date{}

\begin{document}
\maketitle

\begin{abstract}
Random features provide a standard approach to scaling kernel methods to large datasets by approximating the kernel with a finite-dimensional feature map. Existing works of random feature methods primarily focus on well-specified models and light-tailed noise conditions, leaving a gap in understanding their performance under more general conditions, such as misspecified models and heavy-tailed noise. In this paper, we provide a comprehensive and improved analysis of kernel ridge regression with random features (KRR-RF) under general conditions. In the light-tailed noise setting, we improve upon the worst-case uniform bounds of the feature map used in the previous works and lead to the minimax optimal learning rates with $2r+\gamma> \alpha_0$, where $(r,\gamma,\alpha_0)$ denote the regularity of the target function, the effective dimension of the kernel, and the embedding index of the RKHS, respectively. In the heavy-tailed noise setting, we leverage a recently developed Hilbert-valued Fuk–Nagaev inequality and extend the same optimal results under finite weak moment conditions. The additional cost to achieve robustness is a mildly stronger explicit sample-size requirement than in the light-tailed case. For both settings, our results recover the conditions random features budgets (up to log factors) required in the previous works. In the most benign case when $\alpha_0=\gamma$, KRR-RF are minimax optimal for all $r \in (0,1]$. To the best of our knowledge, this is the first work to establish optimal learning rates for KRR-RF under weak moment conditions in the misspecified setting. 
Numerical experiments and a real data analysis validate the derived learning rates and demonstrate their practical relevance in the presence of mis-specification and heavy-tailed noise.
\end{abstract}

\begin{keywords}
Random Features; Misspecified Models; Robust Learning; Weak Moments; Generalization.
\end{keywords}

\section{Introduction}
In a standard supervised learning setting, we are given a dataset $\{(\bfx_i, y_i)\}_{i=1}^n$ consisting of $n$ independent and identically distributed samples drawn from an unknown distribution $\rho$ on $\mathcal{X} \times \mathbb{R}$. The goal is to learn a function $f:\mathcal{X} \rightarrow \mathbb{R}$ that minimizes the expected risk
\begin{align}\label{expected_risk}
\mathcal{E}(f) = \mathbb{E}_{(\bfx, y) \sim \rho}[(f(\bfx) - y)^2].
\end{align}
The minimizer of \eqref{expected_risk}, called the regression function, is given by the conditional expectation $f_{\rho}(\bfx)=\int_{\mathbb{R}} y d\rho(y|\bfx)$, where $\rho(y|\bfx)$ denotes the conditional distribution of $y$ given $\bfx$. Since $\rho$ is unknown, $f_{\rho}$ cannot be
computed directly, it must be estimated from the training
samples within a pre-specified hypothesis space. 

A natural and widely adopted choice of the hypothesis space is
a reproducing kernel Hilbert space (RKHS) $\mathcal H$
associated with a continuous, symmetric, positive semi-definite
kernel $K:\mathcal X\times\mathcal X\to\mathbb R$
\citep{aronszajn1950theory,steinwart2008support}. Kernel
methods estimate the regression function within $\mathcal H$,
typically by solving a regularized empirical risk minimization
problem. Among various kernel methods, kernel ridge regression (KRR) is one of the most popular approaches due to its simplicity and strong theoretical guarantees \citep{smale2007learning,caponnetto2007optimal}. Despite its empirical success, KRR suffers from high computational complexity: it requires storing and inverting an $n \times n$ kernel matrix, which becomes prohibitive for large-scale datasets. This limitation has motivated the development of scalable techniques, which aim to reduce the computational burden while retaining the theoretical properties of kernel methods, such as distributed learning \citep{zhang2015divide,lin2017distributed,lin2020distributed}, Nystr{\"o}m subsampling \citep{williams2001using,rudi2015less,li2023optimal}, random features \citep{rahimi2007random,rahimi2008weighted,rudin2017fourier,rudi2017generalization}, stochastic gradient methods \citep{lin2016optimal,lin2020optimal}, Falkon \citep{rudi2017falkon} and EigenPro 3.0 \citep{abedsoltan2023toward}.

Among these scalable techniques, random features have become
one of the most widely used approaches, owing to their
conceptual simplicity and ease of implementation. The key idea
is to approximate the kernel by the inner product of
finite-dimensional randomized feature maps, so that KRR reduces to a linear ridge regression problem in lower dimensions \citep{rahimi2007random,rahimi2008weighted}. This brings the training cost down from $O(n^3)$ to $O(nM^2)$ and the memory cost from $O(n^2)$ to $O(M^2)$, which is a substantial saving when $M\ll n$. A significant body of research has focused on understanding the generalization properties of random feature approximations \citep{rudi2017generalization,avron2017faster,bach2017equivalence,li2021towards,lanthaler2023error}. In the milestone work by \citet{rudi2017generalization}, the learning rates of KRR-RF align with those of the exact KRR up to a logarithmic factor, provided that the number of random features $M$ is sufficiently large. Their analysis leveraged the integral operator tool and concentration inequalities to establish these results. Crucially, these rates are governed by two fundamental quantities: 
the effective dimension of the kernel characterized by parameter $\gamma$ (\emph{capacity condition}) and the regularity of the target function characterized by parameter $r$ (\emph{source condition}). However, these results are essentially confined to the
well-specified case, i.e., $f_{\rho}\in{\cal H}$ when
$r\geq 1/2$, recent research has started to explore the generalization properties of KRR-RF under this setting. However, these studies are either capacity-independent \citep{lanthaler2023error} or can only cover part of the misspecified regime when $2r+\gamma \geq 1$ \citep{liuon2023,Liijcai2023,lioptimalRF2023,wang2026adaptive}. Whether the full misspecified regime $r\in (0,1]$ can be effectively addressed remains an open question and needs new analytical techniques. 

Another important condition in the analysis of KRR-RF is the noise condition. The above-mentioned results typically assume that the noise or response is bounded or drawn from a light-tailed distribution, such as Gaussian and sub-Gaussian distributions. This Bernstein-type assumption facilitates the derivation of Bernstein concentration inequalities and generalization bounds. However, the light-tailed assumption on the noise may fail in practice. Financial time series containing stock returns during
market crashes, sensor networks subject to occasional
malfunctions, and medical records with rare but extreme patient
measurements all exhibit heavy-tailed behavior, which is
typically modeled by distributions with polynomially decaying
tails, such as the Student-$t$, Fr\'echet, and Pareto
distributions \citep{sigman1999primer,rachev2003handbook}. In
such cases, the noise possesses only a finite number of
moments, and classical Bernstein concentration inequalities underlying
the existing analyses of random features no longer apply. Despite its practical importance, the study of KRR-RF under heavy-tailed noise and misspecified settings remains relatively limited, necessitating the development of new analytical tools and techniques to handle such challenging scenarios. In this paper, we aim to fill this gap by providing a comprehensive and improved analysis of KRR-RF under general conditions including both misspecified settings and heavy-tailed noise. We focus on the following key questions:

\emph{First, in the light-tailed noise setting, what is the precise range of the source and capacity parameters $(r, \gamma)$ for which KRR-RF achieves optimal generalization performance and what are the corresponding requirements on the number of random features $M$ and the sample size $n$? Second, in the heavy-tailed noise setting, does the optimal
rate persist under weak moment conditions? If so, what are the additional costs to achieving the robust generalization performance?}

We answer these questions by establishing new theoretical results that characterize the generalization performance of KRR-RF under both light-tailed and heavy-tailed noise conditions, and provide precise conditions on the source and capacity parameters, the number of random features, and the sample size required to achieve optimal or robust generalization performance. Our contributions can be summarized as follows: 

\begin{itemize}
    \item \textbf{Comprehensive and Improved Theory.} To derive a refined understanding of KRR-RF in the misspecified setting, we use an additional embedding condition characterized by index $\alpha_0 \in [\gamma,1]$, which quantifies the $L^\infty$ embedding to interpolation space of the RKHS. In the light-tailed setting, we establish the minimax optimal learning rates of KRR-RF whenever $2r+\gamma>\alpha_0$. This extends the existing theory $2r+\gamma > 1$ to a more general setting and covers the full range of source conditions $r\in (0,1]$ when $\alpha_0 = \gamma$. This is non-trivial and needs new analytical techniques to give a refined bound on the feature map. In the heavy-tailed noise setting, we further show that the same optimal rates persist when the noise only has a finite number of moments. The price of robustness is a mildly stronger and explicit requirement on the sample size $n$ than in the light-tailed case.  In both settings, our results recover the random-feature budgets required by previous works up to logarithmic factors for both uniformly sampling and data-dependent sampling schemes. Thus the extension to the full misspecified setting and heavy-tailed noise incurs essentially no additional cost in the number of features. To the best of our knowledge, this is the first work to establish optimal learning rates for KRR-RF under weak moment conditions in the mis-specified setting.
    
    \item \textbf{Novel Analytical Techniques.} We first decompose the generalization error into a population RF error, an empirical RF fluctuation error and a noise error. For the latter two stochastic error terms, they are both determined by a leverage-type quantity associated with the feature map $\|C_{M,\lambda}^{-1/2}\phi_M(\bfx)\|_2^2$. We develop new analytical techniques to bound this leverage-type quantity by using a variational leverage identity and the embedding condition. Instead of the uniform worst-case bound $\kappa^2/\lambda$ used in previous work, we establish tighter pointwise control of order $O(\lambda^{-\alpha})$ for $\alpha>\alpha_0$ in high probability. This refinement together with Chebyshev's and Bernstein's inequalities enable us to obtain sharp bounds on the empirical RF fluctuation error and the noise error, leading to improved generalization guarantees for KRR-RF under both light-tailed noise conditions. In the heavy-tailed noise setting, we further combine these techniques with a recently developed Hilbert-valued Fuk-Nagaev inequality \citep{dkh1973some,nagaev1979large,yurinsky2006sums,mollenhauer2025regularized} to bound the noise error under weak moment conditions. We prove that once the sample size is sufficiently large (larger than that for the light-tailed case), the additional polynomial factor can be dominated by the traditional logarithmic factor, resulting in the same optimal learning rates even under heavy-tailed noise.
    
    \item \textbf{Practical Experimental Validation.} We conduct extensive experiments on synthetic data to validate our theoretical findings. The experiments demonstrate the optimal learning rates can be achieved in both light-tailed and heavy-tailed noise settings, even when the regression function lies outside the RKHS. We also apply our methods to real-world datasets to further illustrate the practical effectiveness and robustness of KRR-RF in practical scenarios.
\end{itemize}

\paragraph{Organization.}
The rest of this paper is organized as follows. Section~\ref{sec:pre} introduces the regression model, the RKHS, and its interpolation spaces. 
Section~\ref{sec:krr-rf} reviews kernel ridge regression and its random feature approximation. Section~\ref{sec:main} presents the main results: the assumptions, the optimal learning rates of KRR-RF under 
light-tailed and heavy-tailed noise, and a matching minimax lower bound. 
Section~\ref{sec:comparison} compares our results with related work on KRR, random features, and robust kernel methods. Sections~\ref{sec:sim} and \ref{sec:real} report simulated experiments and a real data application. The proofs of the main results are provided in the appendices.

\section{Preliminaries}\label{sec:pre}

\subsection{Model Setup}
Let ${\cal X}$ be a compact subset of $\mathbb{R}^d$ and
$\mathcal{Y}\subset\mathbb{R}$ be the output space. The joint
distribution $\rho$ on ${\cal X}\times\mathcal{Y}$ can be
decomposed as $\rho(\bfx,y)=\rho_{X}(\bfx)\rho(y|\bfx)$,
where $\rho_{X}$ is the marginal distribution of the
covariates and $\rho(y|\bfx)$ is the conditional distribution
of $y$ given $\bfx$. Under the standard supervised framework of \eqref{expected_risk} and the definition the regression function $f_\rho$, we can express the output $y$ as
\begin{align}
y = f_\rho(\bfx) + \varepsilon,
\end{align}
where $\varepsilon$ is the noise term with zero mean conditioned on $\bfx$. In this paper, we do not impose assumption that the noise $\varepsilon$ is bounded \citep{smale2007learning,li2021towards} or drawn from a light-tailed distribution \citep{caponnetto2007optimal,rudi2017generalization}. 

\subsection{RKHS and its Interpolation Space}

We take the hypothesis space to be a reproducing kernel Hilbert
space (RKHS). Let $K:{\cal X}\times{\cal X}\to\mathbb R$ be a
continuous, symmetric, and positive semi-definite Mercer kernel. The
associated RKHS ${\cal H}$ is defined as the completion of
$\operatorname{span}\{K(\cdot,\bfx):\bfx\in{\cal X}\}$ with
respect to the inner product satisfying
$\langle K(\cdot,\bfx),K(\cdot,\bfx')\rangle_{\cal H}=K(\bfx,\bfx')$,
and it enjoys the reproducing property
$$
f(\bfx)=\langle f,K(\cdot,\bfx)\rangle_{\cal H}$$
for all $f\in{\cal H}$. Since $K$ is a Mercer kernel and $\cal X$ is compact, there exists a constant $\kappa>0$ such that
$$
\sup_{\bfx\in{\cal X}}K(\bfx,\bfx)\le\kappa^2.
$$
We also define the space of square-integrable functions with respect to the marginal distribution $\rho_{X}$ as $L^{2}(\rho_{X})=\{f:{\cal X}\to\mathbb{R} \mid \|f\|_{L^2(\rho_{X})}^2=\int_{\cal X} |f(\bfx)|^2 d\rho_{X}(\bfx)<\infty\}$.

The spectral properties of the RKHS are characterized through
the integral operator $L_K:L^2(\rho_{X})\to L^2(\rho_{X})$,
defined by
\begin{align}\label{integral_operator}
(L_K f)(\bfx)=\int_{\cal X}K(\bfx,\bfx')f(\bfx')
d\rho_{X}(\bfx').
\end{align}
The operator $L_K$ is positive, self-adjoint, and trace-class.
By Mercer's theorem, it admits the spectral decomposition
\begin{align}\label{spectral_decomposition}
L_K=\sum_{i=1}^{\infty}\mu_i\langle\cdot,e_i\rangle_{L^2(\rho_{X})}e_i,
\end{align}
where $\{\mu_i\}_{i\geq 1}$ are the positive eigenvalues in non-increasing
order and $\{e_i\}$ are the corresponding eigenfunctions, which
form an orthonormal basis of $L^2(\rho_{X})$. Throughout this paper, we assume that $L_K$ is injective on $L^2(\rho_X)$, so that the eigenfunctions 
$\{e_j\}_{j\geq 1}$ associated with the positive eigenvalues form a complete orthonormal basis of $L^2(\rho_X)$.

We then introduce the interpolation space (or
fractional power space) associated with the RKHS
${\cal H}$. For any $s\geq 0$, we define the fractional
integral operator $L_K^{s}$ via the spectral decomposition
\eqref{spectral_decomposition} as
$L_K^{s}=\sum_{i=1}^{\infty}\mu_i^{s}\langle\cdot,e_i
\rangle_{L^2(\rho_{X})}e_i$.
\begin{definition}\label{def_interpolation_space}
Given the RKHS ${\cal H}$ induced by the kernel $K$ and its
associated integral operator $L_K$, for any $s\geq 0$, the
interpolation space $[{\cal H}]^s$ is defined as the
range of the operator $L_K^{s/2}$:
\begin{align}\label{interpolation_space}
[{\cal H}]^s
:=\mathrm{Ran}L_K^{s/2}
=\left\{\sum_{i=1}^{\infty}a_i\mu_i^{s/2}e_i:
\sum_{i=1}^{\infty}a_i^2<\infty\right\}
\subset L^2(\rho_{X}).
\end{align}
For any $f,g\in[{\cal H}]^s$ admitting the expansions
$f=\sum_{i=1}^{\infty}a_i\mu_i^{s/2}e_i$ and
$g=\sum_{i=1}^{\infty}b_i\mu_i^{s/2}e_i$, the inner product
and the induced norm on $[{\cal H}]^s$ are defined as
$\langle f,g\rangle_{[{\cal H}]^s}
=\langle L_K^{-s/2}f,L_K^{-s/2}g\rangle_{\rho}
=\sum_{i=1}^{\infty}a_ib_i$ and
$\|f\|_{[{\cal H}]^s}
=\langle f,f\rangle_{[{\cal H}]^s}^{1/2}
=\bigl(\sum_{i=1}^{\infty}a_i^2\bigr)^{1/2}$.
\end{definition}
Equipped with this inner product, $[{\cal H}]^s$ is a
separable Hilbert space with orthonormal basis
$\{\mu_i^{s/2}e_i\}_{i\geq 1}$. In particular,
$[{\cal H}]^1={\cal H}$ and
$[{\cal H}]^0=L^2(\rho_{X})$. Moreover, for any
$0\leq s_1\leq s_2$, there are continuous embeddings
$[{\cal H}]^{s_2}\hookrightarrow
[{\cal H}]^{s_1}\hookrightarrow L^2(\rho_{X})$
\citep{fischer2020sobolev}. The interpolation spaces provide a
natural framework for characterizing the regularity of the
regression function $f_\rho$: the condition
$f_\rho\in[{\cal H}]^s$ with $s>1$ corresponds to
functions smoother than those in ${\cal H}$, whereas
$0<s<1$ corresponds to a larger space of rougher functions.
In the latter regime, $[{\cal H}]^s$ can be
equivalently characterized by the real interpolation method
\citep{steinwart2012mercer}: $[{\cal H}]^s\cong
[L^2(\rho_{X}),{\cal H}]_{s,2}$, where $\cong$ denotes
isomorphism with equivalent norms and $[\cdot,\cdot]_{s,2}$ is
the real interpolation functor.

\section{Kernel Ridge Regression with Random Features (KRR-RF)}\label{sec:krr-rf}

\subsection{Kernel Ridge Regression (KRR)}

We estimate the regression function $f_\rho$ by kernel ridge
regression over the RKHS ${\cal H}$. Given the training
samples $\{(\bfx_i,y_i)\}_{i=1}^n$, the KRR estimator is
defined as the solution of the regularized empirical risk
minimization problem
\begin{align}\label{krr_estimator}
\widehat f_\lambda
=\arg\min_{f\in{\cal H}}\left\{\frac1n\sum_{i=1}^n
\big(f(\bfx_i)-y_i\big)^2+\lambda\|f\|_{\cal H}^2\right\},
\end{align}
where $\lambda>0$ is the regularization parameter. By the
representer theorem \citep{wahba1990spline}, the solution takes
the form $\widehat f_\lambda=\sum_{i=1}^n\alpha_i
K(\cdot,\bfx_i)$, with coefficient vector $\boldsymbol\alpha$ given by
\begin{align}\label{krr_solution}
\boldsymbol\alpha=(\mathbf K+\lambda n\mathbf I)^{-1}\mathbf y,
\end{align}
where $\mathbf K=[K(\bfx_i,\bfx_j)]_{i,j=1}^n$ is the Gram
matrix and $\mathbf y=(y_1,\dots,y_n)^\top$. 

Despite its appealing theoretical properties \citep{smale2007learning,caponnetto2007optimal}, the computational cost of KRR can be prohibitive for large-scale datasets due to the need to invert the $n \times n$ Gram matrix. Typically, this results in a computational complexity of $\mathcal{O}(n^3)$ and a memory requirement of $\mathcal{O}(n^2)$, which can be infeasible for large $n$. This scalability issue has motivated a
line of scalable approximations, among which random features,
introduced in the next section, are one of the most widely used
approaches.

\subsection{Kernel Ridge Regression with Random Features}

Random features provide a scalable approximation of the kernel
by randomized finite-dimensional feature maps
\citep{rahimi2007random}. We assume that the kernel admits an
integral representation
\begin{align}\label{in_re}
K(\bfx, \bfx^{\prime})=\int_{\Omega}\psi(\bfx,\bfw)\psi(\bfx^{\prime},\bfw)d\pi(\bfw),
\end{align}
where $(\Omega,\pi)$ is a probability space and
$\psi:\mathcal{X}\times \Omega \rightarrow \mathbb{R}$ is a feature map and $\pi$ is a probability measure on $\Omega$.
This covers the standard Fourier features for
shift-invariant kernels \citep{rahimi2007random} (see Example~\ref{ex1}) as well as
more general constructions. Drawing $\bfw_1,\dots,
\bfw_M$ independently from $\pi$, we define the random
feature map
\begin{align}\label{rf_map}
\phi_M(\bfx)=\frac{1}{\sqrt M}\bigl(\psi(\bfx,\bfw_1),
\dots,\psi(\bfx,\bfw_M)\bigr)^\top\in\mathbb R^M,
\end{align}
which satisfies $\mathbb E_{\bfw \sim \pi}[\phi_M(\bfx)^\top
\phi_M(\bfx')]=K(\bfx,\bfx')$. 

\begin{example}\label{ex1}
For the Gaussian kernel, $K(\bfx,{\bfx}^{\prime}) = e^{ -\frac{\|\bfx-{\bfx}^{\prime}\|_{2}^{2}}{2\sigma^2}}$ with $\sigma>0$, it is continuous, positive definite, and scaled shift-invariant. Since the inverse Fourier transform
of $e^{-\frac{\|\bfz\|_{2}^{2}}{2\sigma^2}}$ is a Gaussian, and using a basic symmetry argument, we have 
\begin{align*}
K(\bfx,{\bfx}^{\prime})=\frac{1}{2\pi C}\int\int_{0}^{2\pi}\sqrt{2} \cos (\bfw^{T} \bfx+b )\sqrt{2} \cos (\bfw^{T} {\bfx}^{\prime}+b )e^{-\frac{\sigma^2\|\bfx-{\bfx}^{\prime}\|_{2}^{2}}{2} }d\bfw db,
\end{align*}
where $C$ is a normalizing factor. It can be approximated with $\psi (\bfx, \bfw_i)=\sqrt{2} \cos (\bfw_i^{T} \bfx+b_i )$, where $\{{\bfw}_i\}_{i=1}^M$ and $\{b_i\}_{i=1}^M$ are independently drawn from distribution with density $\frac{1}{C}e^{-\frac{\sigma^2\|\bfw\|_{2}^{2}}{2}}$ and $\text{Unif}[0,2 \pi]$, respectively.

\end{example}

Replacing the kernel in \eqref{krr_estimator} with the
finite-dimensional approximation $K_{M}(\bfx, \bfx^{\prime})= \langle \bfphi_{M}( \bfx), \bfphi_{M}(\bfx^{\prime})  \rangle$ leads to the KRR-RF estimator
$\widehat f_{\lambda,M}(\bfx)=\widehat{\boldsymbol\theta}^\top
\phi_M(\bfx)$, where
\begin{align}\label{krr_rf}
\widehat{\boldsymbol\theta}
=(\widehat{C}_{M}+\lambda I)^{-1}\widehat{S}_M^*\widehat{\mathbf{y}}, 
\end{align}
where we define operator $\widehat{S}_M^*=\frac{1}{\sqrt{n}}\big(\phi_M(\bfx_1),\dots,\phi_M(\bfx_n)\big)$, $\widehat{\mathbf y}=\frac1{\sqrt n}(y_1,\dots,y_n)^\top$, and $\widehat{C}_M=\widehat{S}_M^*\widehat{S}_M=\frac1n\sum_{i=1}^n \phi_M(\bfx_i)\phi_M(\bfx_i)^\top$. Furthermore, we denote $S_{M}: \mathbb{R}^{M} \to L^{2}(\rho_{X})$, $(S_{M}\boldsymbol{\beta})(\cdot) = \phi_{M}(\cdot)^{\top}\boldsymbol{\beta}$ for any $\boldsymbol{\beta} \in \mathbb{R}^M$, and $S_M^*: L^2(\rho_X) \to \mathbb{R}^M$ as its adjoint operator. Then the final estimator can be written as 
\begin{align}\label{krr_rf_f}
\widehat f_{\lambda,M} = S_M\widehat{C}_{M,\lambda}^{-1}\widehat{S}_M^*\widehat{\mathbf{y}}, \quad \widehat{C}_{M,\lambda} = \widehat{C}_M + \lambda I.
\end{align}
The population counterpart of $\widehat C_M$ is
$C_M=\mathbb E_{\bfx \sim \rho_{X}}[\phi_M(\bfx)\phi_M(\bfx)^\top]$ and $C_{M,\lambda} = C_M + \lambda I$. Solving \eqref{krr_rf} typically needs the computation cost of $O(M^2n)$ and the memory cost of $O(M^2)$, which is significantly lower than the $O(n^3)$ computation and $O(n^2)$ memory required for the exact kernel method when $M \ll n$.

\begin{remark}[\textbf{Limitations of existing KRR-RF theory}]
Despite the approximation introduced by using a finite number of random features, the KRR-RF estimator can still achieve optimal learning rates under suitable conditions on $M$ and $\lambda$. However, its existing theory is limited in two fundamental aspects. First, most analyses are
restricted to the well-specified case, where the regression
function belongs to the RKHS, i.e., $f_\rho\in{\cal H}$,
corresponding to $r\geq 1/2$ under the source condition. Analyses that allow the misspecified case $r<1/2$ typically
impose the additional restriction $2r+\gamma\geq 1$, which
excludes a substantial part of the source range. Second, existing
results rely heavily on light-tailed noise assumptions, such as
bounded or sub-Gaussian noise, which fail in many applications
where the noise exhibits heavy tails. In the following, we aim to address these limitations by considering the full misspecified case and heavy-tailed noise scenarios.
\end{remark}

\section{Main Results}\label{sec:main}
In this section, we present the main theoretical results regarding the learning rates of the KRR-RF estimator under light-tailed and heavy-tailed noise conditions. For an estimator $\widehat f$, we are interested in bounding the excess risk, i.e.,
$
\mathcal{E}(\widehat f) - \mathcal{E}(f_\rho) = \|\widehat f - f_\rho\|_{L^2_{\rho_X}}^2.
$
We first introduce the common assumptions that will be used in both the light-tailed and heavy-tailed noise scenarios.

\subsection{Assumptions}

\begin{assumption}[Bounded random feature condition]\label{ass1}
    Assume the kernel $K$ admits the integral representation in \eqref{in_re} with $\psi$ being bounded and continuous in both variables. Specifically, there exists a constant $\kappa \geq 1$ such that $|\psi (\bfx, \bfw)| \leq \kappa$ for all $\bfx \in \mathcal{X}$ and $\bfw \in \Omega$. The associated RKHS ${\cal H}$ is separable.
\end{assumption}

Assumption \ref{ass1} requires the feature function to be
bounded and continuous, which implies that the kernel is
bounded, i.e., $\sup_{\bfx\in\mathcal X} K(\bfx,\bfx)\le\kappa^2$ by \eqref{in_re}. This assumption is standard in the analysis of random feature-based kernel methods \citep{rudi2017generalization,li2021towards}.

\begin{figure}[t]
    \centering
    \includegraphics[width=0.8\textwidth]{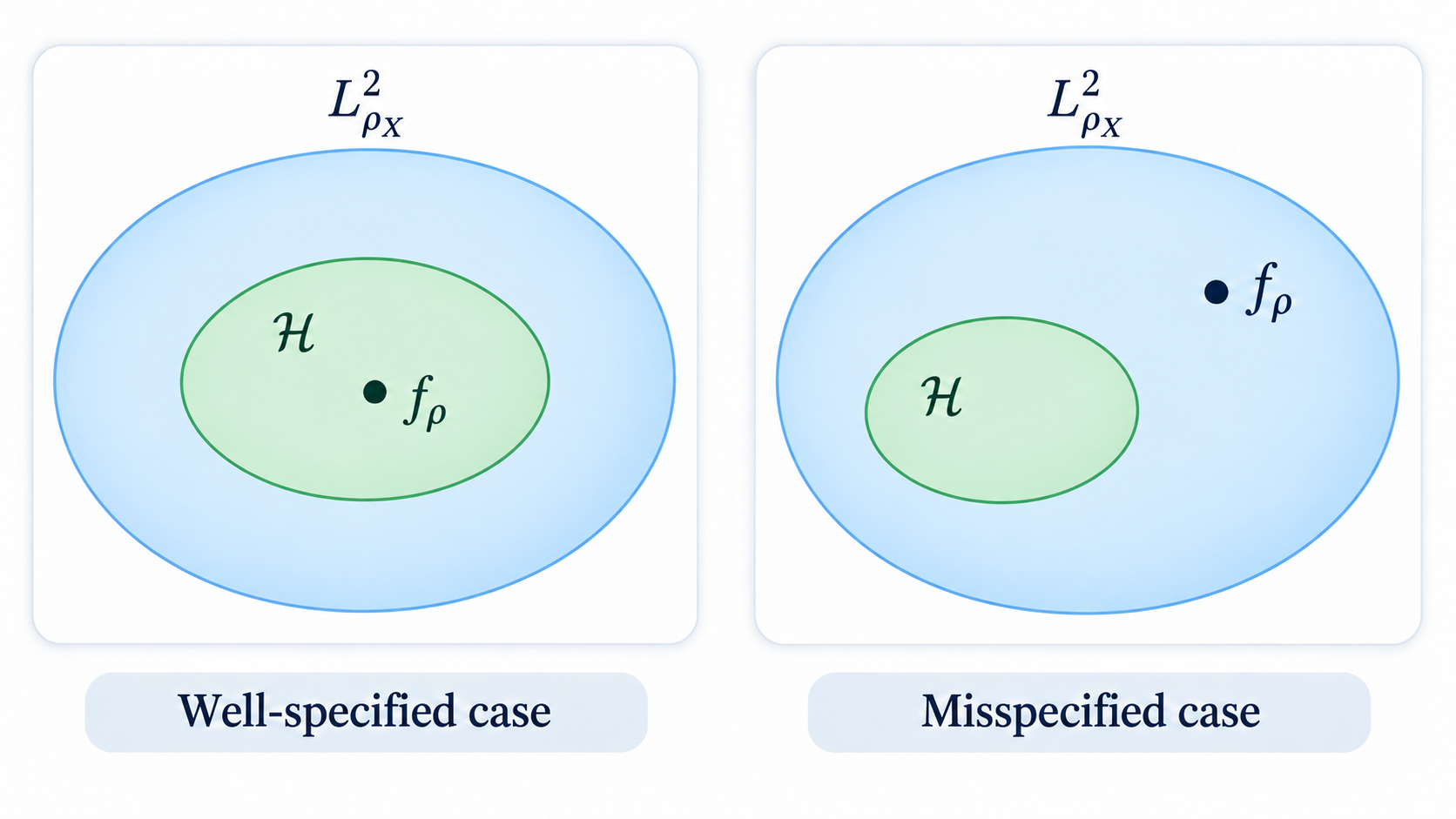}
    \caption{Illustration of the well-specified cases and misspecified case.}
    \label{fig:source_condition}
\end{figure}

\begin{assumption}[Source condition]\label{ass3}
      Suppose the regression function $f_\rho \in [{\cal H}]^{2r}$ for some $r\in(0,1]$ and $\|f_\rho\|_{[{\cal H}]^{2r}} \leq R$ for some constant $R>0$, which is also equivalent to assuming that there exists $h_{\rho} \in L^2_{\rho_{X}}$ such that 
      $$
      f_{\rho}=L_K^r h_{\rho}, \quad \|h_{\rho}\|_{L^2_{\rho_X}}\leq R
      $$
      and $L_K^r$ denotes the $r$-th power of the integral operator $L_K$.
\end{assumption}

Assumption \ref{ass3} is commonly referred to as the source condition \citep{smale2007learning,caponnetto2007optimal,rudi2017generalization}. It characterizes the regularity of the target function $f_\rho$ with respect to the RKHS $\mathcal{H}$, where a larger value of $r$ indicates a smoother target function. When $r \in [1/2,1]$, the target function $f_\rho$ belongs to the RKHS $\mathcal{H}$ itself. When $r \in (0,1/2)$, the target function $f_\rho$ may lie outside the RKHS $\mathcal{H}$, capturing the misspecified case. Existing analyses typically focus on the well-specified case where $r \geq 1/2$ \citep{rudi2017generalization,avron2017faster,li2021towards} or part of the misspecified case where $2r+\gamma >1$. Our work aims to address the full misspecified regime where $r \in (0,1/2)$.

\begin{assumption}[Capacity Condition]\label{ass4}
          For any $\lambda>0$, let the effective dimension be defined as
          ${\cal N}({\lambda}) = \mathrm{Tr}((L_K+\lambda I)^{-1} L_K).$ Assume that there exist constants $Q>0$ and $\gamma \in [0,1]$ such that $${\cal N}({\lambda}) \leq Q^2\lambda^{-\gamma}.$$
  \end{assumption}

The capacity of the RKHS is measured by the effective
dimension, defined as
\begin{align}\label{effective_dimension}
\mathcal N(\lambda)
=\mathrm{Tr}\bigl[(L_K+\lambda I)^{-1}L_K\bigr]
=\sum_{i=1}^{\infty}\frac{\mu_i}{\mu_i+\lambda},
\qquad \lambda>0,
\end{align}
where $\{\mu_i\}_{i\geq 1}$ are the eigenvalues of the integral operator
$L_K$ in \eqref{spectral_decomposition}. The effective
dimension counts the number of eigenvalues that are
significantly larger than $\lambda$, and it determines the
complexity of the hypothesis space at the regularization scale
$\lambda$: a slower eigenvalue decay leads to a larger
$\mathcal N(\lambda)$ and hence a more complex model. Assumption \ref{ass4} 
always holds when $\gamma=1$ by taking $Q=\operatorname{Tr}(L_K) \leq \kappa^2$. It is satisfied when the eigenvalues $\{\mu\}_{i\geq 1}$ of $L_K$ have a polynomial decay, i.e., $\mu_i \leq Ci^{-1/\gamma}$. 
$$
        {\cal N}(\lambda)=  \sum_{i=1}^{\infty}\frac{\mu_i}{\mu_i+\lambda}\leq \sum_{i=1}^{\infty}\frac{C}{C+i^{1/\gamma}\lambda} 
        \leq \int_{0}^{\infty}\frac{C}{C+t^{1/\gamma}\lambda}dt\leq \lambda^{-\gamma}\int_0^\infty \frac{C}{C+t^{1/\gamma}}d t
        \leq C_{u}\lambda^{-\gamma},
$$
for some constant $C_u$.

The following assumption is a key condition to establish the learning rates of KRR-RF under misspecification. Typically, an RKHS $\cal H$ has the embedding property into $L^\infty(X)$ of order $\alpha \in [\gamma,1]$, if there exists a constant $M_{\alpha}>0$ such that 
\begin{align}\label{embedding_property}
\left\|\mathcal{H}|^{\alpha} \hookrightarrow L^{\infty}(X)\right\|\leq \sqrt{M_{\alpha}}
\end{align}
\cite{fischer2020sobolev} proved that \eqref{embedding_property} holds if and only if $\sum_{k=1}^{\infty} \mu_{k}^{\alpha}e_k^2(\bfx)\leq M_{\alpha}$ for $\rho_{\cal X}$-almost all $\bfx \in \cal X$. If there is no such $M_{\alpha}$ satisfying this condition, we set $M_\alpha=\infty$. Note that 
$\sup_{\bfx\in\mathcal X} K(\bfx,\bfx)\le\kappa^2$, it is always true that $M_{\alpha}\geq \kappa$ for $\alpha \geq 1$. By the inclusion relationship of interpolation spaces, we have $\cal H$ has the embedding property of order $\alpha^{\prime}$ whenever it has the embedding property of order $\alpha$ for any $\alpha^{\prime} \geq \alpha$. Then we introduce the embedding index $\alpha_0$ to characterize the smallest order of the embedding property, which first appears in \cite{zhang23mis,zhang2024optimality}.

\begin{assumption}[Embedding condition]\label{ass5}
Suppose that there exists $\alpha_{0}>0$, such that
\[
\alpha_{0}=\inf \left\{\alpha \in\left[\gamma, 1\right]:\left\|\mathcal{H}|^{\alpha} \hookrightarrow L^{\infty}(X)\right\|<\infty\right\},
\]
where we call $\alpha_{0}$ the embedding index of an RKHS $\mathcal{H}$.
\end{assumption}

This assumption implies that for any $\alpha>\alpha_0$, the RKHS $\cal H$ has the embedding property of order $\alpha$, i.e., there exists a constant $M_{\alpha}>0$ such that \eqref{embedding_property} holds. However, the embedding property of order $\alpha_0$ itself may not hold, meaning that $M_{\alpha_0}$ could be infinite. The case $\alpha=\gamma$ corresponds to the benign RKHSs, which implies the RKHS feature space is highly homogeneous and free of extreme leverage points or spatial spikes. Benign RKHSs including RKHS with uniformly bounded eigenfunctions, Sobolev RKHS, RKHS with shift-invariant periodic kernels and so on. More examples and detailed discussions can be found in Section 4 of \cite{zhang2024optimality}.

\begin{remark}\label{remark:embedding_index}
Assumption \ref{ass5} implies the maximal capacity condition
used in our analysis, i.e., $\mathcal N_\infty(\lambda)
=\operatorname*{esssup}_{\bfx\in X}\sum_j
\frac{\mu_j}{\mu_j+\lambda}e^2_j(\bfx) \leq M_{\alpha}\lambda^{-\alpha}$ for any $\alpha>\alpha_0$, which is fundamental to our analysis in the misspecified setting. We provide its proof in Lemma \ref{lem:embed} of Appendix \ref{app:embedding_condition}. Although this condition may not be new and appears in the literature of KRR \citep{fischer2020sobolev,zhang23mis} and general spectral algorithms \citep{zhang2024optimality}, it is nontrivial to involve in the analysis of KRR-RF under misspecified settings. This is because the error decomposition and operator representation in KRR-RF is totally different from that in standard KRR. It plays a crucial role in deriving a high-probability bound on the leverage-type quantity, i.e., $\|C_{M,\lambda}^{-1/2}\phi_M(\bfx)\|_2^2 \leq 8 M_{\alpha} \lambda^{-\alpha}$ for any $\alpha>\alpha_0$ outside an exceptional set of measure at most $O(1/n)$. Its proof is highly nontrivial and relies on some technical lemmas presented in Appendix \ref{app:empirical_rf_fluctuation_error}. In this way, the embedding index $\alpha$ determines the pointwise behavior of the leverage-type quantity, which replaces the uniform worst-case bound $\kappa^2/\lambda$ used in all of the previous analyses \citep{rudi2017generalization,liu2021random,Liijcai2023,wang2026adaptive}. This refinement significantly improves the tightness of the analysis in the misspecified setting, leading to the optimal learning rates of KRR-RF when $2r+\gamma>\alpha_0$.
\end{remark}

\subsection{Learning Rates of KRR-RF in the Light-tailed Noise Settings}

We first focus on the learning rates of KRR-RF under the light-tailed noise condition specified in the following assumption. 

\begin{assumption}[Light-tailed noise condition]\label{ass2}
       Suppose the noise satisfies that $\mathbb{E}(\varepsilon|\bfx)=0$. There exist $\sigma\geq0$ and $B\geq0$ such that for all $p\geq 2$, 
       $$
       \mathbb{E}(|\varepsilon|^{p}| \bfx)\leq \frac{1}{2}p!\sigma^{2}B^{p-2}.$$
\end{assumption}

Assumption \ref{ass2} is commonly referred to as the light-tailed noise condition, which ensures that the noise has sub-exponential tails. It is satisfied by many common noise distributions, including bounded, Gaussian and sub-Gaussian noise.

\begin{theorem}\label{thm1}
Under Assumptions \ref{ass1}-\ref{ass2} with $0<r\le1$, $\gamma \le \alpha_0 \le1$, assume $2r+\gamma>\alpha_0$, and let $\lambda=n^{-\frac{1}{2r+\gamma}}$.
Suppose the number of random features satisfies
\begin{equation}\label{eq:Mbudgetmain1}
M\gtrsim
\begin{cases}
n^{\frac{1}{2r+\gamma}}\log\frac{cn}{\delta}, & 0<r<\frac12,\\[2mm]
n^{\frac{1+\gamma(2r-1)}{2r+\gamma}}\log\frac{cn}{\delta}, & \frac12\le r\le1,\\[2mm]
\end{cases}
\end{equation}
for any $\delta\in(0,1)$, if the sample size $n\geq n_0$ is large enough, then with probability at least $1-\delta$ over the features and the samples,
\begin{align}\label{eq:thm1}
\bigl\|\widehat f_{\lambda,M}-f_\rho\bigr\|^2_{L^2(\rho_X)}
\le C_1n^{-\frac{2r}{2r+\gamma}}\log^2\frac{c_1}{\delta},
\end{align}
where $c, c_1, C_1$ are positive constants independent of $n$, $M$ and $\delta$, $n_0$ is a sufficiently large constant independent of $n$, $M$.
\end{theorem}

With the specified choice of the regularization parameter $\lambda=n^{-\frac{1}{2r+\gamma}}$ and the number of random features $M$ satisfying \eqref{eq:Mbudgetmain1}, Theorem \ref{thm1} guarantees that KRR-RF achieves the optimal learning rates of order $O(n^{-\frac{2r}{2r+\gamma}})$ with high probability. The derived bound aligns with those established in the literature \citep{rudi2017generalization,liu2021random,Liijcai2023,wang2026adaptive}. However, our results pertain to a broader range of source conditions satisfying $2r+\gamma> \alpha_0$, extending beyond the well-specified case $r \geq 1/2$ \citep{rudi2017generalization} and the misspecified case $2r+\gamma \geq 1$ \citep{liu2021random,Liijcai2023,wang2026adaptive}. This extension comes from our tight bounds on the leverage-type quantity, i.e., $\|C_{M,\lambda}^{-1/2}\phi_M(\bfx)\|_2^2 \leq 8 M_{\alpha} \lambda^{-\alpha}$ for any $\alpha>\alpha_0$ outside an exceptional set of measure at most $O(1/n)$. It allows us to obtain improved bounds of (i) the similarity between operators $C_{M,\lambda}$ and $\widehat{C}_{M,\lambda}$, leading to the condition on the sample size from $n \gtrsim \lambda^{-1}$ to $n \gtrsim \lambda^{-\alpha}$, and (ii) the noise error term, improving it roughly from ${\cal O}(n^{-1}\lambda^{-1/2}+n^{-1/2}\lambda^{-\gamma/2}))$ to ${\cal O}(n^{-1}\lambda^{-\alpha/2}+n^{-1/2}\lambda^{-\gamma/2}))$. In the benign case when $\alpha_0=\gamma$, the optimal learning rates can be achieved for the full source range $r\in (0,1]$. We also note that the required number of random features $M$ also recover the optimal scaling with respect to the sample size $n$ as established in previous works $\Omega(n^{\frac{1+\gamma(2r-1)}{2r+\gamma}})$ for $r\geq 1/2$ \citep{rudi2017generalization} and $\Omega(n^{\frac{1}{2r+\gamma}})$ for $r \leq 1/2$ and $2r+\gamma >1$ \citep{Liijcai2023,wang2026adaptive}. This demonstrates the efficiency of the random feature approximation in maintaining the optimal learning rates while potentially reducing computational complexity, even for the tough cases when $2r+\gamma \in (\alpha_0,1)$. 

\begin{remark}
The condition on $M$ in the feature budget \eqref{eq:Mbudgetmain1} suppresses an additional logarithmic factor $\log(cn)$ when $\alpha+\zeta=1$.  In our proof, the spectral estimate in Lemma~\ref{lem:residual} produces an extra $\log(1/\lambda)$ term, which propagates into Lemmas~\ref{lem:pointwise}, \ref{lem:exceptional}, and Proposition~\ref{prop:fluctuation} and adds a $\log(1/\lambda)$ factor to the feature budget. However, this additional logarithmic factor is independent of $\delta$ and is negligible compared to main terms. Thus we omit it 
from the statement for clarity in this theorem and subsequent theorems. The full feature budget including the logarithmic factor can be found in the proofs.
\end{remark}

In practice, random features are drawn directly from the
spectral measure $\pi$ in the integral representation
\eqref{in_re}. This uniform sampling scheme is easy to
implement, but it ignores the data distribution and therefore typically demands a large number of features to reach the desired approximation accuracy. In the regime $2r+\gamma<1$, the problem is more serious: our results show that uniform sampling requires $M\gtrsim n^{1/(2r+\gamma)}$, which is larger than $n$. This defeats the original purpose of random features, which is to reduce the computational cost of KRR by using $M\ll n$ features. Consequently, to obtain the optimal learning rates with fewer random features in the regime $2r+\gamma >\alpha_0$, we introduce the following compatibility condition, which is commonly used in the
literature \citep{rudi2015less,rudi2017generalization,
liu2020effective}.

\begin{assumption}[Compatibility condition]\label{ass7}
For any $\lambda>0$, define the maximum dimension of random features as
$ \mathcal{F}_{\infty}(\lambda)=\sup _{\bfw \in \Omega}\left\|(L_K+\lambda I)^{-1 / 2} \psi(\cdot, \bfw)\right\|_{L^2(\rho_X)}^2$. Assume there exist constants $\zeta \in[0,1]$ and $F>0$, such that $\mathcal{F}_{\infty}(\lambda) \leq F \lambda^{-\zeta}$. 
\end{assumption}

Assumption \ref{ass7} quantifies the compatibility between the
random feature representation and the data distribution through
the maximum dimension $\mathcal F_\infty(\lambda)$. It always holds for $\zeta=1$ and $F=\kappa^2$ under Assumption \ref{ass1}, which corresponds to the standard uniform sampling scheme of random features. We also define the favorable case when $\zeta=\gamma$, as it indicates a better alignment between the random feature representation and the data distribution. Theoretical example illustrating this scenario can be referred to in \cite{rudi2017generalization}.

\begin{remark}[Difference between $\alpha_0$ and $\zeta$]
\label{rem:zeta_alpha}
Both parameters impose polynomial decay on a leverage-type
quantity, but they concern different objects. The embedding
index $\alpha_0$ is intrinsic to the RKHS: it takes the
supremum over the input space through the eigenfunctions of
$L_K$, is independent of the feature construction. It
determines the statistical difficulty of the problem and thereby
controls the leverage-type quantity, which determines the regime boundary
$2r+\gamma>\alpha_0$ in our main theorems. In contrast, $\zeta$
is a property of the feature representation: it takes the
supremum over the feature parameter space $\Omega$, depends on
the sampling scheme (e.g., compatible sampling achieves the
optimal value $\zeta=\gamma$). It enters our results only
through the feature budget $M\gtrsim\lambda^{-\zeta}$. Finally, we can not identify the leverage-type quantity $\|C_{M,\lambda}^{-1/2}\phi_M(\bfx)\|_2^2$ with the
maximum dimension $\mathcal F_\infty(\lambda)=O(\lambda^{-\zeta})$, and derive the restriction $2r+\gamma\geq\zeta$ from it. The two quantities are not equal because the supremum and the $L^2$ norm are taken over different variables: $\|C_{M,\lambda}^{-1/2}\phi_M(\bfx)\|^2$ is a pointwise quantity in $\bfx\in X$, whereas $\mathcal F_\infty(\lambda)$
averages over $\bfx$ in $L^2_{\rho_X}$ and then takes the
supremum over $\bfw\in\Omega$. 
\end{remark}

\begin{theorem}\label{thm2}
Under Assumptions \ref{ass1}-\ref{ass7} with $0<r\le1$, $\gamma \le \alpha_0 \le1$, assume $2r+\gamma>\alpha_0$, and let $\lambda=n^{-\frac{1}{2r+\gamma}}$.
Suppose the number of random features satisfies
\begin{equation}\label{eq:Mbudgetmain2}
M\gtrsim
\begin{cases}
n^{\frac{\zeta}{2r+\gamma}}\log\frac{cn}{\delta}, & 0<r<\frac12,\\[2mm]
n^{\frac{\zeta+(1+\gamma-\zeta)(2r-1)}{2r+\gamma}}\log\frac{cn}{\delta}, & \frac12\le r\le1,\\[2mm]
\end{cases}
\end{equation}
for any $\delta\in(0,1)$, if the sample size $n\geq n_0$ is large enough, then with probability at least $1-\delta$ over the features and the samples,
\begin{align}\label{eq:thm2}
\bigl\|\widehat f_{\lambda,M}-f_\rho\bigr\|^2_{L^2(\rho_X)}
\le C_2n^{-\frac{2r}{2r+\gamma}}\log^2\frac{c_2}{\delta},
\end{align}
where $c, c_2, C_2$ are positive constants independent of $n$, $M$ and $\delta$, $n_0$ is a sufficiently large constant independent of $n$, $M$.
\end{theorem}

The learning rate established in Theorem \ref{thm2} is the same as those in Theorem \ref{thm1}, but the required number of random features is lower. Specifically, $M$ reduces from $\Omega(n^{\frac{1}{2r+\gamma}})$  to $\Omega(n^{\frac{\zeta}{2r+\gamma}})$ when $0<r<\frac12$ and from $\Omega(n^{\frac{1+\gamma(2r-1)}{2r+\gamma}})$ to $\Omega(n^{\frac{\zeta+(1+\gamma-\zeta)(2r-1)}{2r+\gamma}})$ when $\frac12\le r\le1$. These theoretical gains highlight the advantage of considering the compatibility parameter $\zeta$ in reducing the feature budget. We also note that condition \eqref{eq:Mbudgetmain2} is the same as that in \cite{Liijcai2023}and \cite{wang2026adaptive}, but our analysis provides a more refined source condition $2r+\gamma>\alpha_0$ rather than $2r+\gamma>1$.

\begin{figure}[t]
    \centering
    \includegraphics[width=\textwidth]{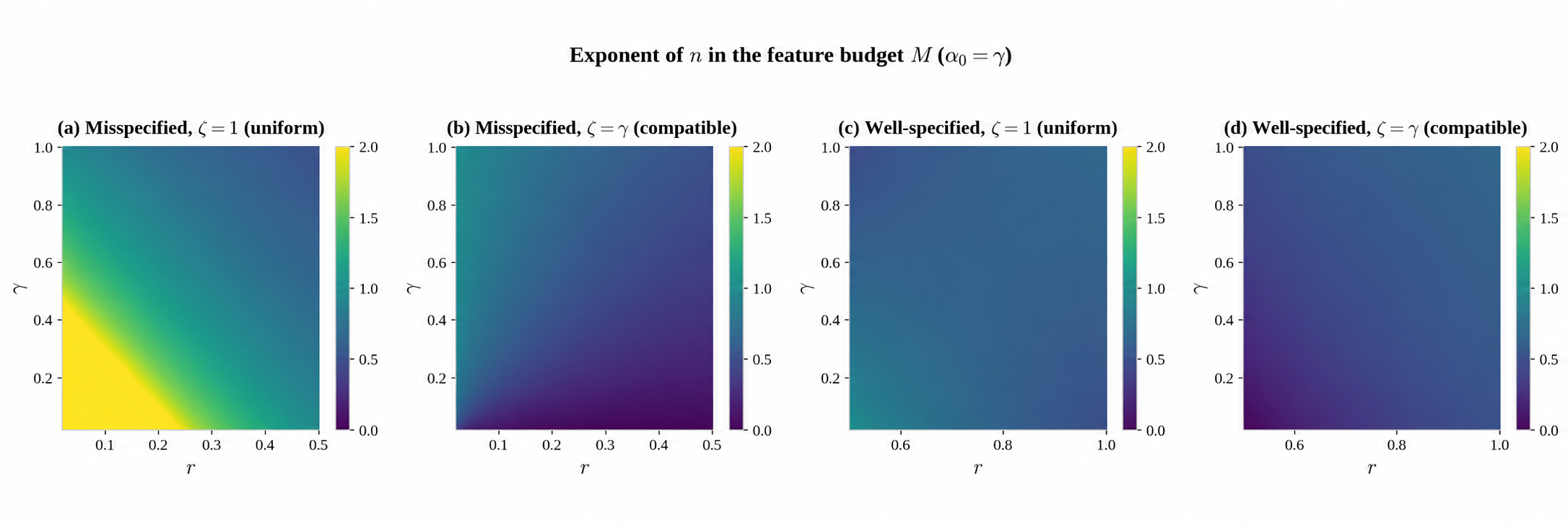}
    \caption{Exponent of $n$ in the random-feature budget $M$ in the benign RKHSs $\alpha_0=\gamma$.}
    \label{fig:feature_budget}
\end{figure}

From Remark \ref{rem:zeta_alpha}, there is no general ordering
between $\zeta$ and $\alpha_0$. Nevertheless, there exist benign cases in which two parameters coincide: for kernels whose eigenfunctions are
well behaved (e.g., uniformly bounded), the embedding index
attains its best value $\alpha_0=\gamma$, and with compatible
sampling the compatibility
parameter attains its optimal value $\zeta=\gamma$, so that
$\alpha_0=\zeta=\gamma$. In this case, our results show that
KRR-RF achieves the minimax optimal learning rate for all
$r\in(0,1]$. Figure \ref{fig:feature_budget} illustrates the exponent of $n$ in the random-feature budget $M$ for the benign RKHSs with $\alpha_0=\gamma$. Panels (a) and (b) correspond to the misspecified case with uniform and compatible sampling, respectively. Panels (c) and (d) correspond to the well-specified case with uniform and compatible sampling, respectively.

\begin{corollary}\label{cor1}
Under the conditions in Theorem \ref{thm2}, if we consider the benign RKHSs and compatible random features with $\alpha_0=\zeta=\gamma$, then for all $0<r\le1$, suppose the number of random features satisfies
\begin{equation}\label{eq:Mbudgetmain4}
M\gtrsim
\begin{cases}
n^{\frac{\gamma}{2r+\gamma}}\log\frac{cn}{\delta}, & 0<r<\frac12,\\[2mm]
n^{\frac{2r+\gamma-1}{2r+\gamma}}\log\frac{cn}{\delta}, & \frac12\le r\le1,\\[2mm]
\end{cases}
\end{equation}
for any $\delta\in(0,1)$, if the sample size $n\geq n_0$ is large enough, then with probability at least $1-\delta$ over the features and the samples,
\begin{align}\label{eq:cor1}
\bigl\|\widehat f_{\lambda,M}-f_\rho\bigr\|^2_{L^2(\rho_X)}
\le C_3n^{-\frac{2r}{2r+\gamma}}\log^2\frac{c_3}{\delta},
\end{align}
where $c, c_3, C_3$ are positive constants independent of $n$, $M$ and $\delta$, $n_0$ is a sufficiently large constant independent of $n$, $M$.
\end{corollary}

For the compatible random features, the required number of random features $M$ is always sublinear in the sample size $n$, which ensures computational efficiency while achieving the minimax optimal learning rate for the whole source range of $0<r\le 1$.

\subsection{Learning Rates of KRR-RF in the Heavy-tailed Noise Settings}

We next establish the learning rates of KRR-RF under the heavy-tailed noise condition specified in Assumption \ref{ass6}.

\begin{assumption}[Heavy-tailed noise condition]\label{ass6}
For all $\bfx \in \mathcal{X}$, assume there exist some constants $\sigma, B>0$, and $p \in \mathbb{N}$ with $p\geq3$ such that 
$$
\mathbb{E}(\varepsilon| \bfx)=0, \quad \mathbb{E}(\varepsilon^{2}| \bfx)\leq \sigma^2, \quad \text{and} \quad \mathbb{E}(|\varepsilon|^{p}| \bfx)\leq B.
$$
\end{assumption}

Compared to Assumption \ref{ass2}, Assumption \ref{ass6} is a weak moment condition on the noise and only requires a higher conditional absolute moment $p\geq 3$ to be bounded. It allows the noise to have polynomially decaying tails, such as the Student-$t$ distribution with more than three degrees of freedom, or centered Pareto and Fr\'echet distributions with shape parameter larger than three. Under this
condition, classical exponential Bernstein concentration inequalities no
longer apply, and our analysis instead relies on a Hilbert-valued Fuk--Nagaev inequality (Lemma \ref{lem:fuknagaev}) to
control the noise term.

\begin{proposition}\label{prop1}
Under Assumptions \ref{ass1}, \ref{ass3}, \ref{ass5}, \ref{ass7}, \ref{ass6} with $0<r\le1$, $\gamma \le \alpha_0 <1$ and $\alpha>\alpha_0$, let $\lambda\in (0,\min\{1,\|L_K\|\})$, suppose the number of random features satisfies
\begin{equation}\label{eq:Mbudgetmain5}
M\gtrsim
\begin{cases}
\lambda^{-\zeta}\log\frac{cn}{\lambda\delta}, & 0<r<\frac12,\\[2mm]
\lambda^{-\{\zeta+(1+\gamma-\zeta)(2r-1)\}}\log\frac{cn}{\lambda\delta}, & \frac12\le r\le1,\\[2mm]
\end{cases}
\end{equation}
for any $\delta\in(0,1)$, and the sample size $n$ satisfies
\begin{equation}\label{eq:covbudget_1}
n\ge\frac{832}{3}M_{\alpha}\lambda^{-\alpha}\log\frac{20M}{\delta},
\end{equation}
then with probability at least $1-\delta$ over the features and the samples,
\begin{align*}
\bigl\|\widehat f_{\lambda,M}-f_\rho\bigr\|_{L^2(\rho_X)} \le CR\left(
\sqrt{\frac{M_\alpha\lambda^{2r-\alpha}}{n\delta}}+\lambda^r
\right)+ C\left(
\frac{B^{\frac1p}\bigl(M_\alpha\lambda^{-\alpha}\bigr)^{\frac12-\frac1p}{\cal N}(\lambda)^{\frac1p}}{\delta^{\frac1p}n^{1-\frac1p}}+
\sigma\sqrt{\frac{{\cal N}(\lambda)\log\frac{c}{\delta}}{n}}
\right),
\end{align*}  
where $c, C$ are positive constants independent of $n$, $M$ and $\delta$
\end{proposition}

Proposition \ref{prop1} provides a error decomposition for the estimator $\widehat f_{\lambda,M}$ under the weak moment condition on the noise. For the light-tailed noise, the bound can be further simplified using classical Bernstein concentration inequalities, i.e., 
\begin{align*}
\bigl\|\widehat f_{\lambda,M}-f_\rho\bigr\|_{L^2(\rho_X)} \le CR\left(
\sqrt{\frac{M_\alpha\lambda^{2r-\alpha}}{n\delta}}+\lambda^r
\right)+C\left(
\frac{B\sqrt{M_\alpha\lambda^{-\alpha}}\log\frac2\delta}{n}+
\sigma\sqrt{\frac{{\cal N}(\lambda)\log\frac2\delta}{n}}
\right).
\end{align*}  
The key difference is the additional term $O\big(\lambda^{\frac{(2-p)\alpha}{2p}}{\cal N}(\lambda)^{\frac1p}\delta^{-\frac{1}{p}}n^{-(1-\frac1p)}\big)$ in the bound. This term arises due to the heavy-tailed nature of the noise, which we use a recently developed Hilbert-valued Fuk--Nagaev
inequality \citep{dkh1973some,nagaev1979large,yurinsky2006sums,mollenhauer2025regularized} instead of classical Bernstein concentration inequalities. It yields polynomial (rather than logarithmic) dependence on
the confidence level $\delta$. Using Assumption \ref{ass4} that 
${\cal N}(\lambda)\lesssim\lambda^{-\gamma}$ and $\lambda=n^{-\frac{1}{2r+\gamma}}$, one verifies that
polynomial term will be dominated by the logarithmic term whenever $n$ satisfies the explicit requirement
\begin{equation}\label{eq:heavybudget1}
n\gtrsim \delta^{-\frac{2(2r+\gamma)}{(p-2)(2r+2\gamma-\alpha)}} \Rightarrow 
\lambda^{\frac{(2-p)\alpha}{2p}}{\cal N}(\lambda)^{\frac1p}\delta^{-\frac{1}{p}}n^{-(1-\frac1p)}\lesssim \lambda^r, \quad {\cal N}(\lambda)^{1/2}n^{-1/2} \lesssim \lambda^r.
\end{equation}
This leads to the conclusion that for sufficiently large $n$, the heavy-tailed noise term does not affect the overall convergence rate, and the estimator achieves the same rate as in the light-tailed case.

\begin{theorem}\label{thm3}
Under Assumptions \ref{ass1}-\ref{ass5}, \ref{ass7}, \ref{ass6} with $0<r\le1$, $\gamma \le \alpha_0 \le1$ and $\alpha>\alpha_0$, assume $2r+\gamma>\alpha_0$, and let $\lambda=n^{-\frac{1}{2r+\gamma}}$.
Suppose the number of random features satisfies \eqref{eq:Mbudgetmain2} any $\delta\in(0,1)$, and if the sample size satisfies 
\begin{equation}\label{eq:nrequirement}
n\ge  \max \{n_0,c\delta^{-\frac{2(2r+\gamma)}{(p-2)(2r+2\gamma-\alpha)}} \}
\end{equation}
then with probability at least $1-\delta$ over the features and the samples,
\begin{align}\label{eq:thm3}
\bigl\|\widehat f_{\lambda,M}-f_\rho\bigr\|^2_{L^2(\rho_X)}
\le C_4n^{-\frac{2r}{2r+\gamma}}\log^2\frac{c_4}{\delta},
\end{align}
where $c, c_4, C_4$ are positive constants independent of $n$, $M$ and $\delta$, $n_0$ is a sufficiently large constant independent of $n$, $M$ defined in Theorem \ref{thm2}.
\end{theorem}

It is worth emphasizing that under heavy-tailed noise, both
the learning rates and the implementation of the algorithm
remain the same as in the light-tailed case. Specifically, the estimator uses the same regularization parameter $\lambda=n^{-1/(2r+\gamma)}$
and the same random-feature budget $M$, neither of which
depends on the moment parameter $p$. The only difference lies
in the requirement of the sample size: a slightly larger $n$ is needed
to make the polynomial tail term produced by the Fuk--Nagaev
inequality negligible. Hence, robustness to heavy-tailed noise
is achieved at no cost in either the statistical rate or the
computational complexity, only the sample size requirement is affected. 

\begin{figure}[t]
    \centering
    \includegraphics[width=0.9\textwidth]{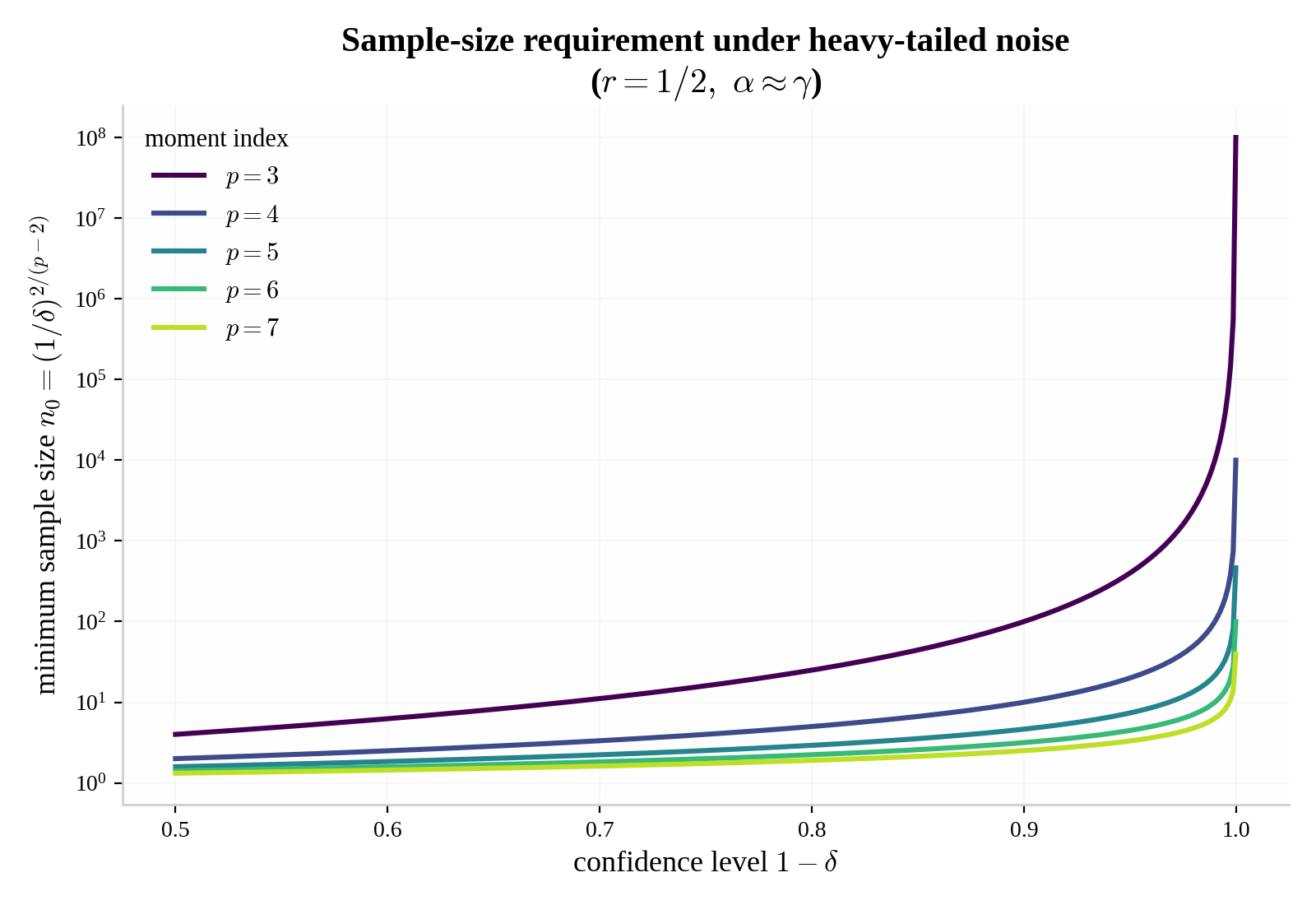}
    \caption{Minimum sample size required for the polynomial term to be dominated by the exponential term at a given confidence level $1-\delta$, for different values of $p$.}
    \label{fig:sample_size}
\end{figure}

For better illustration, Figure \ref{fig:sample_size} gives a simple
example showing the minimum sample size required for the
polynomial term to be dominated by the exponential term at a
given confidence level $1-\delta$, for different values of
$p$. We fix $\lambda=n^{-\frac{1}{2r+\gamma}}$, $r=1/2$, and
$\alpha \approx \gamma$, in which case the requirement
\eqref{eq:heavybudget} reduces to
$n\gtrsim (1/\delta)^{\frac{2}{p-2}}$. As shown in Figure
\ref{fig:sample_size}, the minimum sample size increases as $p$ decreases
or as $\delta$ decreases, which is consistent with the
intuition that heavier-tailed noise (smaller $p$) or higher
confidence (smaller $\delta$) requires more samples to achieve
the optimal learning rate.

The following two corollaries provide the results for the uniform and compatible random features under heavy-tailed noise, respectively.

\begin{corollary}\label{cor3}
Under the conditions in Theorem \ref{thm3}, if we consider the benign RKHSs and uniformly random features with $\alpha_0=\gamma$ and $\zeta=1$, then for all $0<r\le1$, suppose the number of random features satisfies
\eqref{eq:Mbudgetmain1} and the sample size satisfies \eqref{eq:nrequirement} , then with probability at least $1-\delta$ over the features and the samples,
\begin{align}\label{eq:cor3}
\bigl\|\widehat f_{\lambda,M}-f_\rho\bigr\|^2_{L^2(\rho_X)}
\le C_5n^{-\frac{2r}{2r+\gamma}}\log^2\frac{c_5}{\delta},
\end{align}
where $c, c_5, C_5$ are positive constants independent of $n$, $M$ and $\delta$.
\end{corollary}

\begin{corollary}\label{cor2}
Under the conditions in Theorem \ref{thm3}, if we consider the benign RKHSs and compatible random features with $\alpha_0=\zeta=\gamma$, then for all $0<r\le1$, suppose the number of random features satisfies
\eqref{eq:Mbudgetmain4} and the sample size satisfies \eqref{eq:nrequirement} , then with probability at least $1-\delta$ over the features and the samples,
\begin{align}\label{eq:cor2}
\bigl\|\widehat f_{\lambda,M}-f_\rho\bigr\|^2_{L^2(\rho_X)}
\le C_6n^{-\frac{2r}{2r+\gamma}}\log^2\frac{c_6}{\delta},
\end{align}
where $c, c_6, C_6$ are positive constants independent of $n$, $M$ and $\delta$.
\end{corollary}

\subsection{Minimax Lower Bound} 

To demonstrate the optimality of the derived rates for KRR-RF, we recall the  information-theoretic lower bound for KRR \citep{caponnetto2007optimal,zhang23mis}. Since the KRR-RF estimator is itself a
particular learning algorithm, any minimax lower bound
established over the class of all estimators automatically
applies to it. We impose the following standard two-sided eigenvalue decay condition.

\begin{assumption}
\label{ass_low}
There exist constants $c_{-},c_{+}>0$ such that the eigenvalues
$\{\mu_i\}_{i\ge1}$ of $L_K$ satisfy
$c_{-}j^{-1/\gamma}\le\mu_j\le c_{+}j^{-1/\gamma}$ for all
$j\ge1$ and $\gamma \in (0,1)$.
\end{assumption}

\begin{theorem}\label{thm:lower_bound}
Let $\rho_{X}$ be a probability distribution on $X$ such that the marginal $\rho_X$ satisfies Assumption \ref{ass_low}, and $\mathcal{P}$ be the class of all probability distributions on $X \times \mathbb{R}$ with marginal $\rho_X$ satisfying Assumption \ref{ass3} and Assumption \ref{ass2}. Then there exists a positive constant $C$ and a distribution $\rho \in \mathcal{P}$, 
\[
\inf_{\widehat f} \sup_{\rho \in \mathcal{P}} \mathbb{E} \bigl[\|\widehat f - f_\rho\|^2_{L^2(\rho_X)}\bigr] \ge C n^{-\frac{2r}{2r+\gamma}},
\]
where the infimum is taken over all possible estimators $\widehat f$ based on $n$ i.i.d. samples from $\rho$.
\end{theorem}

The proof of Theorem \ref{thm:lower_bound} follows from standard information-theoretic arguments for minimax lower bounds in KRR, as detailed in \cite{caponnetto2007optimal,zhang23mis}. Note that the proof use a Gaussian noise in the regression model, which 
satisfies both the light-tailed condition (Assumption \ref{ass2}) and the weak moment condition (Assumption \ref{ass6} with any $p\geq 3$), this
lower bound applies to the distribution classes considered in
both Theorem \ref{thm1} and Theorem \ref{thm2}. Consequently,
the rates $n^{-\frac{2r}{2r+\gamma}}$ established in both
theorems are minimax optimal up to logarithmic factors once $n$ is large enough. Our contribution thus lies in providing matching upper bounds under the misspecified setting when $2r+\gamma>\alpha_0$ and the heavy-tailed noise condition.

\section{Comparison with the Related Work}\label{sec:comparison}

We compare our results with existing work on kernel ridge regression (KRR) and random features (RF) methods, and robust kernel methods.

\subsection{Comparison with Kernel Ridge Regression (KRR)}

Kernel ridge regression (KRR) is a classical method for nonparametric regression in reproducing kernel Hilbert spaces (RKHS). In the context of learning theory \citep{vapnik1999nature,vapnik1999overview}, it has been extensively studied for its statistical properties and optimality under various assumptions. Early works \citep{cucker2002mathematical,devito2005model,wu2006learning,smale2007learning} established the capacity-independent learning rates for KRR, deriving bounds $O(n^{-\frac{2r}{2r+1}})$ solely from the source condition. Mendelson and Neeman \citep{mendelson2010regularization} further obtained capacity-independent minimax rates that cover targets outside the RKHS. \cite{caponnetto2007optimal} 
considered the polynomial decay of the eigenvalues of the RKHS and derived the capacity-dependent learning rates $O(n^{-\frac{2r}{2r+\gamma}})$ of KRR with source condition. However, these results primarily focus on the well-specified setting where the target function lies within the RKHS. 

Research on the misspecified settings, where
$f_\rho$ lies outside ${\cal H}$, has developed along two lines. The first line relies on the source and capacity conditions alone: optimal rates were established for KRR with specific kernels \citep{wang2022gaussian} and for more general spectral algorithms \citep{lin2018optimal,lin2020optimal}, but these analyses require $2r+\gamma\geq 1$, which excludes the harder part of the misspecified settings. The second line combines the integral operator approach with the embedding property of the RKHS, which controls the sup-norm of
functions in the interpolation spaces through an index
$\alpha_0 \in[\gamma,1]$. Initiated by \cite{steinwart2009optimal} and further developed by \cite{fischer2020sobolev}, this line established optimal rates of KRR under the boundedness of $f_\rho$, which implicitly requires $2r>\alpha_0$. Based on this approach, the analysis in \cite{zhang23mis} extended the optimality results to Sobolev-type RKHSs for the full range of $r\in(0,1]$. \cite{zhang2024optimality} further considered the general spectral algorithms beyond KRR and established their optimality under similar conditions. Our work differs from the above works in two aspects. First, they all study the exact kernel estimator, while we
study KRR with random features, for which the proof techniques are completely different. For example, one of our contribution is a tight upper bound of the leverage-type quantity, which is not involved in the analysis of the exact kernel estimator. Second, they assume the noise is either bounded or light-tailed, whereas our results remain valid under heavy-tailed noise with only a finite $p$-th moment.

Another line of research keeps the least squares loss and
studies its behavior when the noise possesses finite moments. \cite{mendelson2016upper} developed sharp
bounds for multiplier empirical processes, which provide the main technical tool for analyzing least squares without exponential concentration. Building on this,
\cite{han2019convergence} derived convergence rates for the nonparametric least squares estimator under a finite $p$-th moment condition, and \cite{mendelson2015learning,
kuchibhotla2022least,tong2025least} obtained similar results in related settings. However, the learning rates in these works depend explicitly on the moment parameter $p$ and are slower than the optimal rate under light-tailed noise. Recently, by leveraging a Hilbert-valued Fuk--Nagaev inequality, \cite{mollenhauer2025regularized} derived optimal
learning rates for regularized least squares over RKHS under heavy-tailed noise, which recovers the existing results for light-tailed noise.  Their analysis, however, is restricted to the exact kernel estimator in the well-specified case $r\geq 1/2$. Our work only extends the analysis to random features and the misspecified setting when $2r+\gamma >\alpha_0$. This is nontrivial due to the additional approximation error introduced by random features and the challenges posed by heavy-tailed noise.

\subsection{Comparison with Random Feature (RF) Methods}

Random features were first introduced by \cite{rahimi2007random,rahimi2008weighted} as a scalable approximation of kernel methods, where the kernel is
approximated by the inner product of randomized
finite-dimensional feature maps. Early works focused on the approximation quality and computational benefits of random features \citep{rahimi2007random,sriperumbudur2015optimal}. For the generalization performance, the seminal work by \cite{rudi2017generalization} established the capacity-dependent learning rates $O(n^{-\frac{2r}{2r+\gamma}})$ for KRR with random features (KRR-RF), showing that with a sufficient number of random features, the generalization performance can match that of the exact kernel method. The required
number of features was further reduced by sampling strategies that exploit the data distribution, such as leverage score sampling \citep{bach2017equivalence,avron2017faster}. Random
features have also been combined with distributed learning and general Lipschitz loss functions \citep{liu2021random,li2021towards,wang2024communication,wang2024optimal}. However, most of these results are restricted to the well-specified settings where $r\geq1/2$. Research on random features in the misspecified setting is relatively limited. \cite{liu2021random} studied KRR-RF in the misspecified setting and provided error analysis under certain conditions. \cite{Liijcai2023} further investigated this problem under distributed environments. Recently, \cite{wang2026adaptive} proposed an adaptive parameter selection method for KRR-RF in the misspecified setting. However, all of these works require $2r+\gamma> 1$, which excludes the harder part of the
misspecified regime. Thus none of the existing works fully address the misspecified regime when $2r+\gamma <1$.

The restriction $2r+\gamma\geq1$ in these works stems largely from the use of the worst-case uniform bound
$\|C_{M,\lambda}^{-1/2}\phi_M(\bfx)\|_2^2\leq\kappa^2/\lambda$. This bound corresponds to the trivial case $\alpha=1$ and ignores the embedding property of the RKHS.
Our work leverages a more refined analysis that takes into account the embedding property of the RKHS, allowing us to handle the misspecified settings when $2r+\gamma> \alpha_0$. For the benign RKHSs with $\alpha_0=\gamma$,  our results can cover the full source range $r\in(0,1]$. It is also worth noting the above works primarily focus on the light-tailed noise setting, whereas our analysis accommodates heavy-tailed noise with only a finite $p$-th moment.

Recently, \cite{wang2026generalization} studied the generalization performance of robust learning with random features under the $(1+\delta)$-th moment condition. Their idea is to truncate the response variable to control the influence of heavy-tailed noise. Our analysis differs from this approach for three reasons. First, they impose a $(1+\delta)$-th moment condition on the response for some $\delta>0$, which is different from our finite $p$-th moment assumption with $p\geq 3$. Second, their analysis relies on truncation techniques, whereas our approach directly handles heavy-tailed noise without truncation, leading to a more general and flexible framework. Third, the learning rates derived in their work depend explicitly on the moment parameter $\delta$, whereas our results achieve optimality independent of the moment parameter $p$. The proof techniques and theoretical tools used in our work are also fundamentally different from those in \cite{wang2026generalization}, since their work only need to control the truncation error compared to that in \cite{rudi2017generalization}.

\subsection{Comparison with Robust Kernel Methods}

Robust kernel methods have been extensively studied in the literature, particularly in the context of heavy-tailed noise and outlier-robust learning. These methods typically involve modifying the loss function such as the Huber loss, the Tukey's biweight loss or quantile loss to mitigate the influence of extreme observations. \cite{christmann2007consistency} studied the
relationship between the loss function and the tail behavior of the response. \cite{feng2022statistical} and \cite{liu2026kernel} analyzed kernel Huber regression and adaptive kernel Huber regression from a statistical learning viewpoint. \cite{feng2020learning} established generalization bounds for robust empirical risk minimization under $(1+\delta)$-th moment conditions. Similar works include \cite{tong2023ivanov} on Ivanov regularized Huber regression and \cite{tong2025modifiedhuber} on robust regression with modified Huber's loss. Quantile regression \citep{koenker1978regression,li2007quantile} estimates conditional quantiles instead of the conditional mean and is insensitive to heavy-tailed errors. Kernel quantile regression with random features was also studied in \citet{wang2024optimal,wang2024communication} without the light-tailed noise assumption. Nevertheless, quantile regression is not well-suited for conditional mean regression, which is the focus of this paper. There are also some works keep the squared loss but replace empirical risk minimization by a robust aggregation of the data. Representative examples include median-of-means estimators \cite{hsu2014heavy} and minimization of robust empirical  risk \cite{brownlees2015empirical}, which achieve sub-Gaussian-type deviations under weak moment assumptions. 

We note that robust procedures for general losses and robust aggregation schemes are important topics in 
their own right, but they are beyond the scope of this paper. First, the learning rates in the aforementioned works depend  explicitly on the moment parameter and deteriorate as the moment condition weakens. Moreover, no matching  lower bounds were established, so it is unclear whether these rates are optimal. Second, methods based on losses other than the squared loss, and quantile loss even no longer target the conditional mean. The resulting estimators also have no closed-form solution, which leads to high computational cost in the kernel setting. In contrast, our KRR-RF estimator retains the squared loss and hence admits an explicit solution with low computational cost. Theoretically, it also achieves the minimax optimal rate $n^{-\frac{2r}{2r+\gamma}}$ uniformly over the misspecified range $2r+\gamma>\alpha_0$, which is independent of the moment parameter $p$.

\section{Simulated Experiments}\label{sec:sim}

In this section, we conduct numerical experiments to evaluate the proposed KRR-RF method, including (i) validating the learning rates under both light-tailed and heavy-tailed noise, (ii) examining the effect of the number of random features $M$, and (iii) comparing with robust kernel methods.

\subsection{Simulation Setup}

Suppose that ${\cal X}=[0,1]$ and the marginal distribution of $\bfx$ is uniform on $[0,1]$. We consider the RKHS induced by the periodic Sobolev kernel 
$$
K_{\ell}(x,x')=\sum_{k\in\mathbb Z}
\mu_{k,\ell}e^{2\pi \mathrm{i}k(x-x')},
\qquad
\mu_{k,\ell}=
\left[1+(2\pi\ell k)^2\right]^{-2}.
$$
For experiments (i), we set $\ell=1$, for experiments (ii) and (iii), we set $\ell=1/2\pi$. For both choices of the kernel scale $\ell$, the eigenfunctions of the integral operator $L_K$ are the Fourier basis functions $e_k(x)=e^{2\pi \mathrm{i}kx}$, with corresponding eigenvalues
$$
\mu_{k,\ell}=\left[1+(2\pi\ell k)^2\right]^{-2}
\asymp (1+|k|)^{-4},\qquad k\in\mathbb Z.
$$
Consequently, the effective dimension satisfies 
$\mathcal N(\lambda)
=\sum_{k\in\mathbb Z}\frac{\mu_{k,\ell}}{\mu_{k,\ell}+\lambda}
\asymp \lambda^{-1/4}$ and capacity condition holds for $\gamma=1/4$. According to \cite{zhang2024optimality}, the periodic Sobolev RKHS is benign RKHS with $\alpha_0=\gamma=1/4$. The embedding condition holds for every $\alpha \in (1/4,1)$. We construct the true regression function as 
$$
f_{\rho,\ell}(x)= \frac{1}{c_0}
\sum_{k=1}^{\infty}
\frac{\mu_{k,\ell}^{r}\sqrt{2}
\cos(2\pi k(x-0.37))}
{\sqrt{k}\log(ek)}, \quad c_0^2=\sum_{k=1}^{\infty}\frac{1}{k\log^2(ek)}.
$$
By orthonormality of the Fourier basis, the source condition holds for $f_{\rho,\ell}=L_K^r g_{\rho,\ell}$ with $g_{\rho,\ell}= \frac{1}{c_0}
\sum_{k=1}^{\infty}\frac{\sqrt{2}
\cos(2\pi k(x-0.37))}
{\sqrt{k}\log(ek)}$ and $\|g_{\rho,\ell}\|^2_{_{L^2(\rho_X)}}=1$. Moreover, there holds that
$$
\|f_{\rho,\ell}\|^2_{_{L^2(\rho_X)}}= \frac{1}{c_0^2} \sum_{k=1}^{\infty} \frac{\mu_{k,\ell}^{2r}}{k \log^2(ek)} \leq 1,\quad \|f_{\rho,\ell}\|^2_{\mathcal H_{K_\ell}}= \frac{1}{c_0^2} \sum_{k=1}^{\infty} \frac{\mu_{k,\ell}^{2r-1}}{k \log^2(ek)} \asymp \frac{1}{c_0^2} \sum_{k=1}^{\infty} \frac{{k}^{3-8r}}{\log^2(ek)},
$$
where we use $\mu_{k,\ell} \asymp (1+|k|)^{-4} \in (0,1]$. Thus we can obtain that $f_{\rho,\ell} \in \mathcal H_{K_\ell}$ when $r\geq 1/2$ and  $f_{\rho,\ell} \notin \mathcal H_{K_\ell}$ when $r< 1/2$. 

We consider the following data generating process for our simulations:
$$
y=f_{\rho,\ell}(x)+\sigma\varepsilon,
$$
where three noise error are considered: (i) Gaussian noise $\varepsilon \sim N(0,1)$; (ii) Student's t noise $\varepsilon \sim t_4/\sqrt{2}$ with 4 degrees of freedom; (iii) Pareto noise $\varepsilon \sim \sqrt{48/5}(\text{Pareto}(5,1)-5/4)$. All three noise distributions have zero mean and unit variance. Here $\sigma$ is the noise level, where the first two experiments use $\sigma=1$. Case [i] is the standard light‑tailed distribution and satisfies assumption \ref{ass2}. Case [ii] and Case [iii] are heavy‑tailed distributions with finite variance and satisfies Assumption \ref{ass6} with $p=3$ and $p=4$.

\vspace{0.2cm}
\noindent \textbf{Random Features Construction.} We consider two random feature representations of the periodic Sobolev kernel: the uniform random features and the compatible random features. We first consider the real orthonormal Fourier basis 
$$
e_0(x)=1,\quad
e_{k,c}(x)=\sqrt2\cos(2\pi kx),\quad
e_{k,s}(x)=\sqrt2\sin(2\pi kx),
\quad k\ge1.
$$
The corresponding eigenvalues are 1 for $e_0(x)$ and $\mu_{k,\ell}$ for both $e_{k,c}(x)$ and $e_{k,s}(x)$ with $k\ge1$. Let $S_\ell=1+2\sum_{k=1}^\infty\mu_{k,\ell}$, we draw a random index $\omega$ from
$\Omega_{\mathrm U}=\{0\}\cup\{(k,c),(k,s):k\geq1\}$ with probabilities
$$
  \mathbb P(\omega=0)=\frac{1}{S_\ell},\qquad
  \mathbb P(\omega=(k,c))
  =\mathbb P(\omega=(k,s))=\frac{\mu_{k,\ell}}{S_\ell}.
$$
The corresponding feature is
\begin{align}\label{U_rf}
  \psi_{\mathrm U,\ell}(x,\omega)=\sqrt{S_\ell}\,e_\omega(x).
\end{align}
The representation is unbiased because
\begin{align}
  \mathbb E_\omega\bigl[
     \psi_{\mathrm U,\ell}(x,\omega)
     \psi_{\mathrm U,\ell}(x',\omega)\bigr]
  =e_0(x)e_0(x')+\sum_{k=1}^\infty\mu_{k,\ell}
    \bigl[e_{k,c}(x)e_{k,c}(x')
           +e_{k,s}(x)e_{k,s}(x')\bigr]\nonumber=K_\ell(x,x').
\end{align}
Since $\sup_{x,\omega}|\psi_{\mathrm U,\ell}(x,\omega)|^2\leq2S_\ell$,
this is a bounded random-feature representation. Moreover, its
compatibility quantity satisfies
$$
  \mathcal F_{\infty,\mathrm U}(\lambda)
  =\sup_{\omega\in\Omega_{\mathrm U}}
  \left\|(L_{K_\ell}+\lambda I)^{-1/2}
        \psi_{\mathrm U,\ell}(\cdot,\omega)
  \right\|_{L^2(\rho_X)}^2
  =\sup_{\omega\in\Omega_{\mathrm U}}
     \frac{S_\ell}{\mu_\omega+\lambda}
   =\frac{S_\ell}{\lambda},
$$
where $\mu_0=1$ and $\mu_{(k,c)}=\mu_{(k,s)}=\mu_{k,\ell}$.
Thus, Uniform RF satisfies Assumption \ref{ass1} and \ref{ass7} with $\zeta=1$.  For the compatible random features, we draw
$u\sim\operatorname{Unif}[0,1]$ and set
\begin{equation}\label{C_rf}
  \psi_{\mathrm C,\ell}(x,u)
  =\sum_{k\in\mathbb Z}\sqrt{\mu_{k,\ell}}
    e^{2\pi\mathrm{i}k(x-u)}=\frac{\cosh\!\left(\dfrac{1/2-d_{\mathbb T}(x,u)}{\ell}\right)}
         {2\ell\sinh\!\left(\dfrac{1}{2\ell}\right)},
\end{equation}
where $d_{\mathbb T}(x,u)=\min\{|x-u|,1-|x-u|\}$. The representation is also unbiased by orthonormality of the Fourier basis
$$
\mathbb E_u\bigl[
    \psi_{\mathrm C,\ell}(x,u)
    \psi_{\mathrm C,\ell}(x',u)\bigr]
  =\sum_{k\in\mathbb Z}\mu_{k,\ell}
     e^{2\pi\mathrm{i}k(x-x')}
  =K_\ell(x,x').
$$
Unlike the first representation, this construction has the favorable
compatibility quantity
$$
  \mathcal F_{\infty,\mathrm C}(\lambda)
  =\sup_{u\in[0,1]}
  \left\|(L_{K_\ell}+\lambda I)^{-1/2}
          \psi_{\mathrm C,\ell}(\cdot,u)
  \right\|_{L^2(\rho_X)}^2=\sum_{k\in\mathbb Z}
     \frac{\mu_{k,\ell}}{\mu_{k,\ell}+\lambda}
   =\mathcal N(\lambda)
   \asymp\lambda^{-1/4}.
$$
Thus, Compatible RF satisfies Assumption \ref{ass1} and \ref{ass7} with $\zeta=\gamma=1/4$.

\vspace{0.2cm}
\noindent \textbf{Error Metric and Numerical Evaluation.} 
Since the regression function is known in the simulation, we evaluate the squared population prediction error of 
an estimator $\widehat{f}$:
\[
\mathrm{Error} = \int_0^1(\widehat f(\bfx)-f_{\rho,\ell}(\bfx))^2\,dx
=\|\widehat f\|_{L^2(\rho_X)}^2
-2\langle\widehat f,f_{\rho,\ell}\rangle_{L^2(\rho_X)}
+\|f_{\rho,\ell}\|_{L^2(\rho_X)}^2 .
\]
Unlike the empirical test MSE computed from noisy responses, this criterion directly measures the estimation error and excludes the irreducible observation noise. For uniform RF, the prediction error is evaluated through the Fourier coefficients using Parseval's identity 
\citep{stein2003fourier}. For KRR and compatible RF, the cross terms are computed using the spectral representation of the target function, while the squared norms of the fitted predictors are evaluated by piecewise Gauss--Legendre quadrature \citep{golub1969calculation, davis1984methods}. The regression function $f_{\rho,\ell}$ is approximated by its first $J=131072$ Fourier coefficients and accounts for the remaining infinite series tail through numerical bounds.

\begin{figure}[t]
    \centering
    \includegraphics[width=\textwidth]{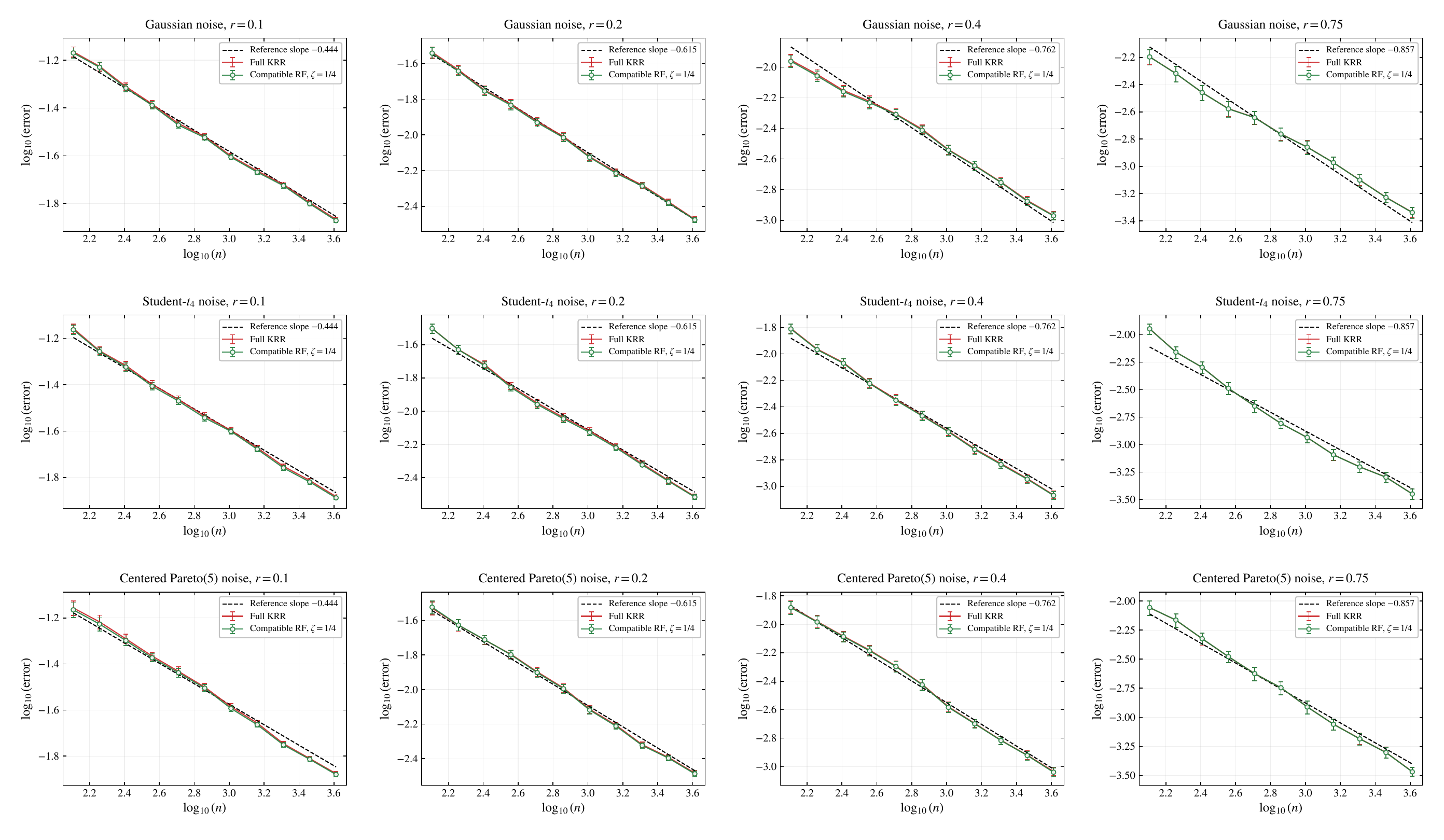}
    \caption{Learning rates under Gaussian (top row), standardized Student-$t_4$ (middle row), and centered 
standardized Pareto$(5)$ noise (bottom row). Columns correspond to $r\in\{0.1,0.2,0.4,0.75\}$. Each panel shows 
log errors against log sample sizes for full KRR and Compatible RF. The 
dashed line marks the theoretical slope $-2r/(2r+\gamma)$.}
    \label{fig1}
\end{figure}

\subsection{Validation of the Learning Rates}
We first validate the theoretical learning rates of KRR-RF under the light-tailed and heavy-tailed noise settings. We consider four different regression functions $f_{\rho,\ell}$ with varying smoothness levels $r \in \{0.1,0.2,0.4,0.75\}$, where the first three are less smooth and do not belong to the RKHS and the last one is sufficiently smooth and belongs to the RKHS. In particular, $r=0.1$ and $r=0.2$ satisfy $2r+\gamma<1$, a regime not covered by existing theory, and $r=0.1$ corresponds to an unbounded regression function. We vary the sample size $n\in\{128,181,256,362,512,724,1024,1448,2048,2896,4096\}$. For each $(r,n)$ setting, we compare the full KRR with KRR-RF methods using compatible random features. Following the theoretical results, we fix the number of random features by 
$$
 M_n(r)=\left\lceil
 n^{\frac{1/4+(2r-1)_+}{2r+1/4}}
 \log(en)\right\rceil.
$$
The regularization parameter is set according to the theoretical guidance as well, i.e., 
$$
\lambda_n(r)=Cn^{-\frac{1}{2r+1/4}},
$$
where $C$ is a constant chosen from $\{1/16,1/4,1,4,16\}$ by an independent validation set. Each experiment is repeated 100 times and the error is averaged over all repetitions. 

\begin{table}
 \centering
 \caption{Learning-rate slopes for compatible RF, fitted by least squares
 to the $\log_{10}$ mean errors at the final five sample sizes.}
 \label{tab:e1-compatible-slopes}
 \small
 \begin{tabular}{rrrrr}
  \toprule
  $r$ & Theory & Gaussian & Student-$t_4$ & Pareto$(5)$ \\
  \midrule
  $0.10$ & $-0.4444$ & $-0.4408$ & $-0.4748$ & $-0.4815$ \\
  $0.20$ & $-0.6154$ & $-0.5750$ & $-0.6509$ & $-0.6131$ \\
  $0.40$ & $-0.7619$ & $-0.7238$ & $-0.7869$ & $-0.7530$ \\
  $0.75$ & $-0.8571$ & $-0.8078$ & $-0.8188$ & $-0.9028$ \\
  \bottomrule
 \end{tabular}
\end{table}

Figure \ref{fig1} plots the logarithm of the error against logarithm of $n$ for different settings. The black dashed lines have the theoretical slopes
$-\frac{2r}{2r+\gamma}=-\frac{2r}{2r+1/4}$, namely $-0.444$, $-0.615$, $-0.762$, and $-0.857$ for increasing $r \in \{0.1,0.2,0.4,0.75\}$. Their slopes are fixed in advance, only their intercepts are fitted
to the eleven full-KRR error values. In all the panels, the log-error decreases almost linearly in $\log n$, and the slopes are close to the 
theoretical slope under all three noise distributions. For the less smooth functions with $r=0.1$ and $r=0.2$ where $2r+\gamma<1$, the empirical slopes still match the theoretical predictions well, demonstrating the validity of our theory even in this previously unexplored regime. We also note that the slopes under Student-$t_4$ and Pareto$(5)$ noise are almost the same as under Gaussian noise, which shows that heavy tails do not slow down the rate once the sample size is sufficiently large. We further report the fitted slopes of compatible RF by regressing the 
log errors on log sample sizes at the five largest sample sizes. There is no systematic difference between the three noise distributions, which 
is consistent with our theory that the rate does not depend on the tail behavior.

\begin{figure}[t]
    \centering
    \includegraphics[width=\textwidth]{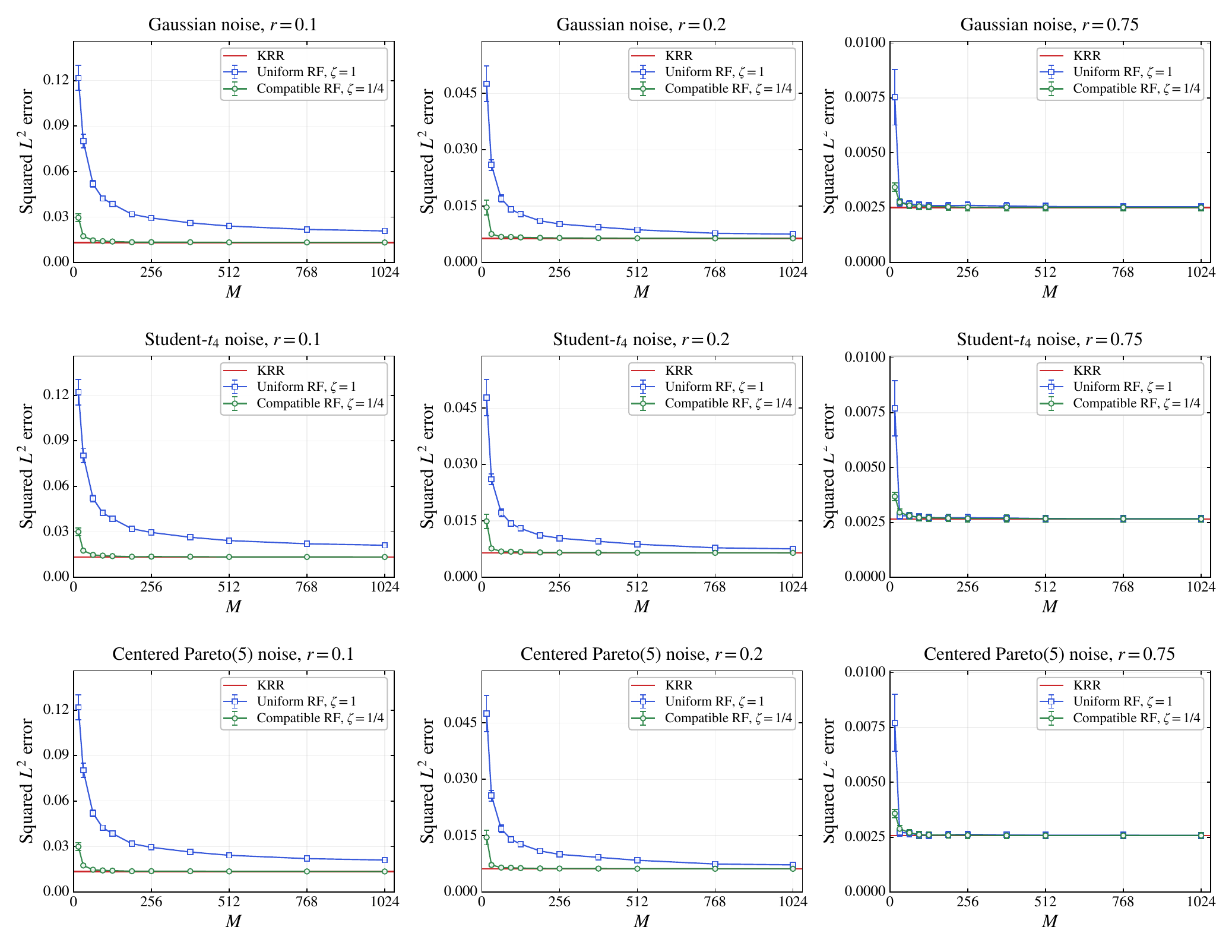}
    \caption{Effect of the number of random features $M$ under Gaussian (top row), standardized Student-$t_4$ (middle row), and centered 
standardized Pareto$(5)$ noise (bottom row). Columns correspond to $r\in\{0.1,0.2,0.75\}$. Each panel shows errors against $M$ for full KRR, uniform RF, and compatible RF.}
    \label{fig2}
\end{figure}

\subsection{Effect of the Number of Random Features}
We next investigate the effect of the number of random features on the learning performance of KRR-RF. We use the periodic Sobolev kernel with length scale $\ell=1/(2\pi)$, so that its eigenvalues are $\mu_{k,\ell}=(1+k^2)^{-2}$ for $k\in\mathbb Z$. We fix the training sample size at $n=2048$ and consider three source parameters $r\in\{0.1,0.2,0.75\}$. We compare full KRR, uniform RF ($\zeta=1$), and compatible RF
($\zeta=1/4$) by varying the number of random features $M\in\{16,32,64,96,128,192,256,384,512,768,1024\}$. For a given $r$, the regularization parameter is selected in an independent pilot experiment using full KRR under Gaussian noise. Specifically, we minimize the mean squared population prediction error over $\lambda\in\{10^{-5},10^{-4.5},\ldots,10^{0}\}$, using $20$ pilot repetitions per candidate. This yields $\lambda=10^{-4}$ for $r=0.1$, $\lambda=10^{-3}$ for $r=0.2$, and $\lambda=10^{-2}$ for $r=0.75$. For each $r$, the chosen $\lambda$ is fixed across all three noise
distributions, both random feature constructions, and all values of $M$.

Figure~\ref{fig2} shows how the prediction error changes with the number of random features $M$ under the three noise distributions, where the red line represents the full KRR. Compatible RF ($\zeta=1/4$) reaches the error level of full KRR with very few features in all nine panels. Its error is already close to the  full KRR line at $M\approx 32$ and stays flat afterwards, so increasing $M$ further brings almost no gain. However, uniform RF requires a much larger number of features to reach a comparable error level. The gap between the two sampling schemes shrinks as $r$ grows: for $r=0.75$ both methods converge to KRR within a few number of features and their curves almost coincide. These results are consistent with our theory: the feature budget required by KRR-RF is governed by the 
compatibility exponent $\zeta$, so a compatible sampling scheme ($\zeta=\gamma$) attains the optimal rate with far 
fewer features than uniform sampling ($\zeta=1$), and the saving is most pronounced in the misspecified settings. Finally, the same pattern holds under all three noise distributions, which demonstrates the robustness of the KRR-RF approach once the sample size is sufficiently large.

\subsection{Comparison with Robust Kernel Methods}
To further evaluate the competitive advantage of KRR-RF, specifically the trade-off between statistical efficiency 
and computational cost, we compare it with other four robust kernel methods with random features: 
\begin{itemize}
    \item \textbf{TKRR-RF}: Truncated KRR with random features \citep{wang2026generalization}. This method applies a truncation step to the heavy-tailed responses before performing KRR with random features. However, the truncation level needs to be carefully chosen to balance bias and robustness. We use the theoretically recommended truncation level from \citep{wang2026generalization} in our experiments.
    \item \textbf{KHR-RF}: Kernel Huber Regression \citep{feng2022statistical,liu2026kernel} with random features. This method employs the Huber loss \citep{huber2011robust} to achieve robustness against heavy-tailed noise, with a tuning parameter controlling the trade-off between efficiency and robustness. We use parameters $a=c\widehat s$, where $\widehat s$ is computed from three-fold out-of-fold KRR residuals on the training sample and equals their median absolute deviation divided by $0.675$. We select $c\in\{0.5,1,2,4,8,16,32,\infty\}$ from pilot experiments on the training sample and evaluate the performance on an independent validation set. The choice $c=\infty$ reduces KHR-RF to KRR-RF. We use the iteratively reweighted least squares algorithm for the convex Huber objective. 
    \item \textbf{KQR-RF}: Kernel Quantile Regression \citep{li2007quantile,wang2024optimal} with random features. This method focuses on estimating conditional quantiles of the response variable, providing robustness against heavy-tailed noise by targeting specific quantiles rather than the mean. We use the quantile level $\tau=0.5$ and ADMM algorithm  \citep{boyd2011distributed} with exact soft-thresholding of the median loss in our experiments.  
    \item \textbf{MoM-RF}: Median-of-Means KRR-RF \citep{hsu2014heavy}. This method divides the data into several blocks, computes the KRR-RF estimator within each block, and then takes the geometric median of these estimators to achieve robustness against heavy-tailed noise. The block size $B \in \{3,5,9\}$ is selected from pilot experiments on the training sample and evaluated on an independent validation set. We use  blockwise KRR-RF solutions followed by a modified Weiszfeld iteration \citep{weiszfeld2009point}.
\end{itemize}

\begin{figure}[t]
    \centering
    \includegraphics[width=\textwidth]{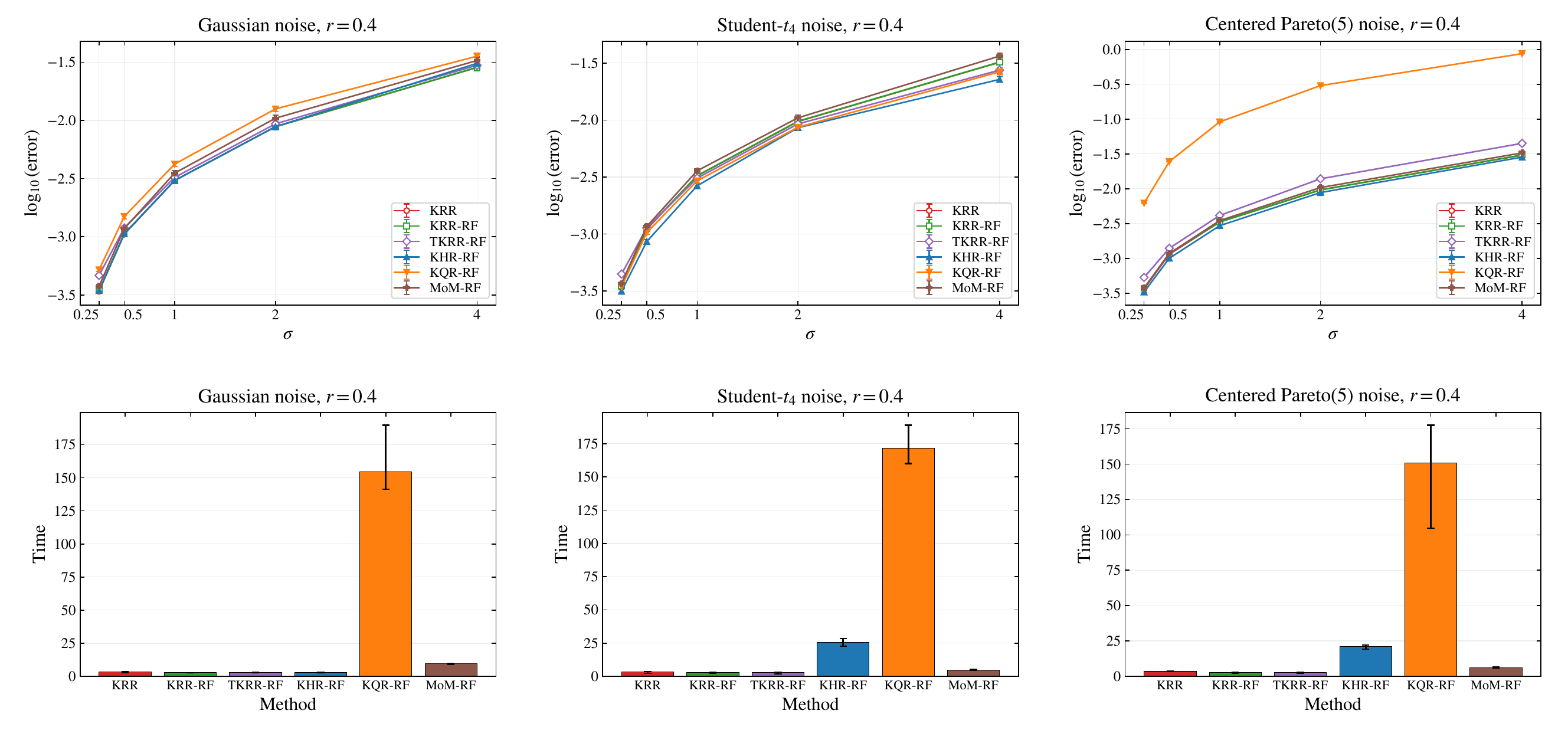}
    \caption{Comparison of KRR-RF with robust kernel methods under Gaussian (left column), standardized Student-$t_4$ (middle column), and centered standardized Pareto$(5)$ noise (right column). The top row shows log errors against the noise level $\sigma$, while the bottom row shows training times.}
    \label{fig3}
\end{figure}

The experimental setup follows the same protocol as in the previous subsection. We fix the regression function with source parameter $r=0.4$ and the training sample size at $n=2048$. We use the compatible sampling scheme with $\zeta=1/4$ for these RF-based methods, and fix the number of random features $M=512$. The regularization parameter for all methods is selected from $10^{-3}\cdot \{1/64,1/16,1/4,1,4,16,64\}$ via an independent validation set. We consider three noise distributions with varying levels $\sigma \in \{0.25,0.5,1,2,4\}$. Figure \ref{fig3} summarizes the prediction accuracy and computational cost of the six methods under the three noise distributions. The top row plots log errors against the noise level $\sigma$, while the bottom row reports the corresponding training time for each method.

Overall, the prediction errors increase as the noise level $\sigma$ becomes larger, as expected. Under Gaussian noise, KRR, KRR-RF, TKRR-RF, KHR-RF, and MoM-RF exhibit very similar prediction errors over the entire range of noise levels, while KQR-RF is slightly less accurate. Under heavy-tailed noise, KHR-RF achieves the lowest prediction error, especially at moderate and large noise levels. However, KRR-RF shows competitive performance compared with KHR-RF, and outperforms TKRR-RF and MoM-RF in most cases. In contrast, KQR-RF shows a substantially larger squared error as $\sigma$ increases, especially in the Pareto noise. This is because for Gaussian and Student-\(t_4\) noise, the conditional mean and median coincide due to symmetry. However, the centered Pareto(5) noise is asymmetric, so its conditional median differs from the conditional mean.

The lower panels of Figure \ref{fig3} demonstrate a clear computational advantage of random-feature methods. KRR-RF and TKRR-RF have the smallest training times among the six procedures, while KQR-RF is by far the most computationally expensive. KHR-RF and MoM-RF require substantially more time than KRR-RF because they involve iterative reweighting or repeated blockwise fitting and aggregation. Taken together, these experiments illustrate a trade-off between robustness
and computational efficiency. Our KRR-RF can achieve competitive prediction performance while maintaining a low computation cost. More importantly, as described in Section \ref{sec:comparison}, these robust methods are highly dependent on the moment parameters of the noise distribution in theory, whereas KRR-RF does not require such prior knowledge and remains effective across different noise settings.

\section{Real Data Application}\label{sec:real}
To further evaluate the practical performance of kernel ridge regression with random features, we conduct experiments on the cpusmall dataset from the LIBSVM Data Repository (\url{https://www.csie.ntu.edu.tw/~cjlin/libsvmtools/datasets/regression.html}). The cpusmall dataset contains $8192$ samples, each with $12$ numerical features describing CPU characteristics. For example, $Iread$ and $Iwrite$ measure the numbers of reads and writes per second between system memory and user memory. The target variable $usr$ is the percentage of time that the CPU runs in user mode, with larger values indicating a better performance. The details of all variables are summarized in Table~\ref{tab:cpusmall-vars}. 

\begin{figure}[t]
    \centering
    \includegraphics[width=\textwidth]{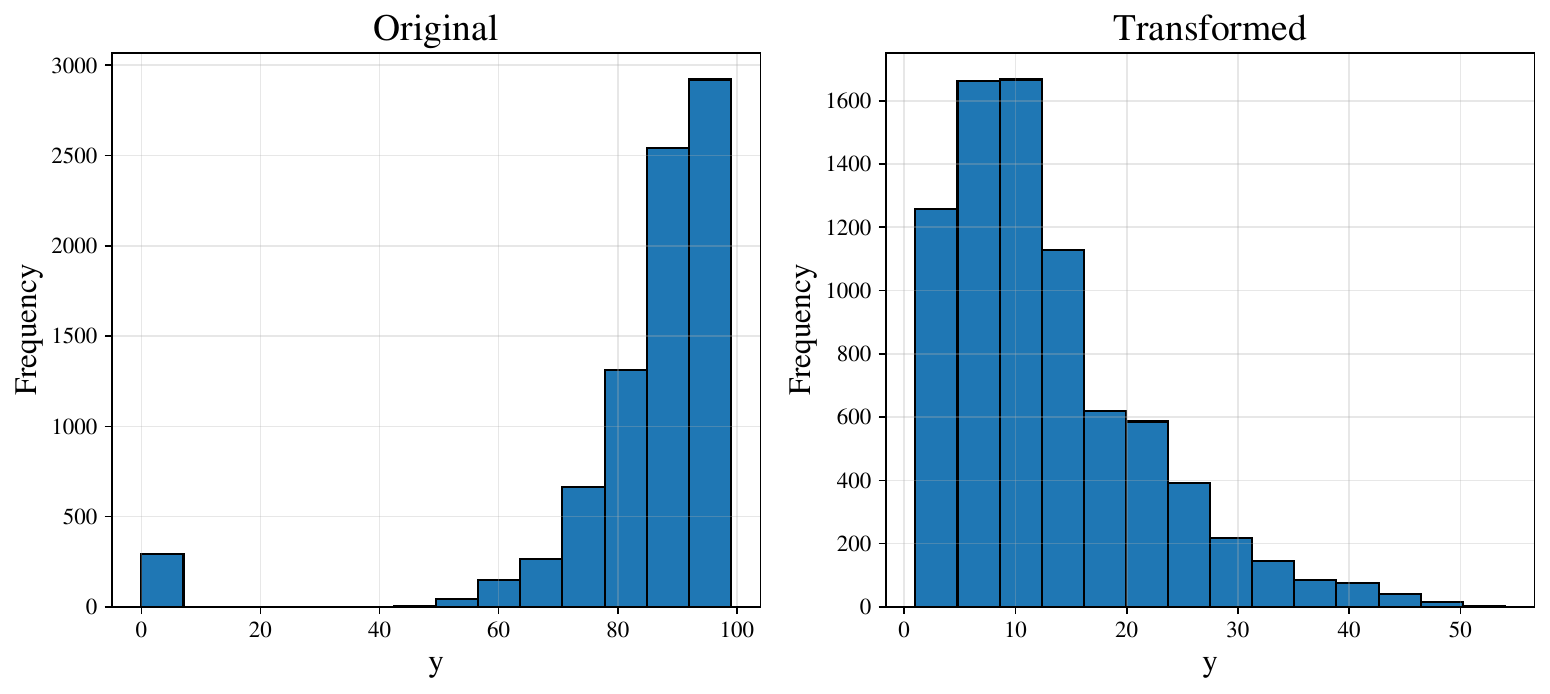}
    \caption{Histogram of the target variable $usr$ before (left) and after (right) transformation.}
    \label{fig:usr-hist}
\end{figure}

We plot the histogram of the target variable $usr$ in the left panel of Figure~\ref{fig:usr-hist}. Most values concentrate near the upper end of the range, with a small proportion of low values, so the distribution is moderately left-skewed. To construct a dataset suitable for studying heavy-tailed behavior, we first remove the samples with extremely 
low values, i.e., $y\leq 10$. This excludes 294 samples, approximately 3.59\% of the original dataset, and leaves 7,898 samples. We then transform the retained responses according to $y \mapsto 100 - y$. After this transformation, the low-end extreme values in the original data become large values, and the  distribution becomes right-skewed. Although the transformed target variable is bounded, but what matters for finite-sample behavior is the size of the moments rather than their existence. In practice, a bounded but strongly right-skewed distribution behaves like a heavy-tailed one when sample size is large. The histogram of the transformed target variable is shown in the right panel of Figure~\ref{fig:usr-hist}. We use these 7,898 samples and their transformed responses in the subsequent regression experiments. 

\begin{table}
\centering
\caption{Details of variables in the cpusmall dataset.}
\label{tab:cpusmall-vars}
\begin{threeparttable}
\small
\setlength{\tabcolsep}{3pt} 
\renewcommand{\arraystretch}{0.65} 
\begin{tabular}{lll}
\toprule
Variables  & Name & Description  \\
\midrule
$x_1$    & Iread    & Reads (transfers per second) between system memory and user memory. \\
$x_2$    & lwrite   & Writes (transfers per second) between system memory and user memory. \\
$x_3$    & scall    & Number of system calls of all types per second. \\
$x_4$    & sread    & Number of system read calls per second. \\
$x_5$    & swrite   & Number of system write calls per second. \\
$x_6$    & fork     & Number of system fork calls per second. \\
$x_7$    & exec     & Number of system exec calls per second. \\
$x_8$    & rchar    & Number of characters transferred per second by system read calls. \\
$x_9$    & wchar    & Number of characters transferred per second by system write calls. \\
$x_{10}$ & runqsz   & Process run queue size. \\
$x_{11}$ & freemem  & Number of memory pages available to user processes. \\
$x_{12}$ & freeswap & Number of disk blocks available for page swapping. \\
\midrule
$y$      & usr      & Portion of time (\%) that CPUs run in user mode. \\
\bottomrule
\end{tabular}
\end{threeparttable}
\end{table}

\begin{figure}[t]
    \centering
    \includegraphics[width=\textwidth]{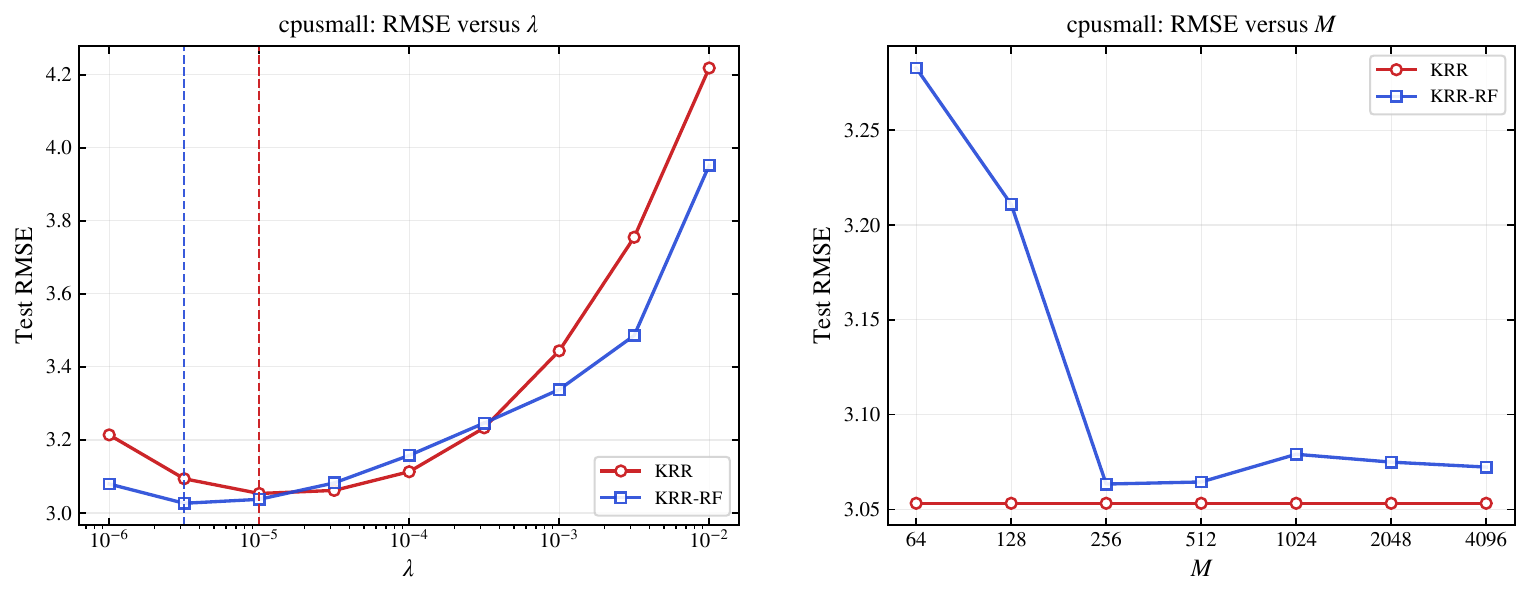}
    \caption{Sensitivity of KRR and KRR-RF to the regularization parameter $\lambda$ and the number of random features $M$.}
    \label{fig:krr-sensitivity}
\end{figure}

We first investigate the sensitivity of KRR and KRR-RF to the regularization parameter $\lambda$ and the number of random features $M$. 
We randomly select 3000 samples for training, 1000 samples for validation, and the remaining 3898 samples are used for testing. All input variables are standardized using the training data. We employ a Gaussian RBF kernel with bandwidth selected using the validation data. Specifically, let $d{med}$ denote the median pairwise Euclidean distance among the first 800 standardized training observations. We select the bandwidth from $(0.5d_{med}, d_{med}, 2d_{med})$ by minimizing the validation RMSE. The corresponding random features are given in Example \ref{ex1}. Predictive performance is evaluated using the root mean squared error (RMSE) on the test data. In the first experiment, we fix $M=1024$ for KRR-RF and vary $\lambda$ from $10^{-6}$ to $10^{-2}$. The left panel of Figure \ref{fig:krr-sensitivity} presents the corresponding test RMSE. Both methods exhibit sensitivity to the regularization parameter. The test error initially decreases and then increases as $\lambda$ grows. This behavior reflects the trade-off between insufficient and excessive regularization. The best parameters are $\lambda=10^{-5}$ for KRR and approximately $\lambda=3.16\times10^{-6}$ for KRR-RF.  In the second experiment, we vary $M$ over $\{64,128,256,512,1024,2048,4096\}$, fix bandwidths $d_{med}$ for KRR and KRR-RF, and select $\lambda$ using the validation set for each $M$. Since standard KRR does not depend on $M$, its test RMSE provides a horizontal reference line. As $M$ increases, the test RMSE of KRR-RF decreases and eventually stabilizes near the test RMSE of standard KRR, indicating that a sufficient number of random features can approximate the full kernel effectively.

\begin{figure}
    \centering
    \includegraphics[width=\textwidth]{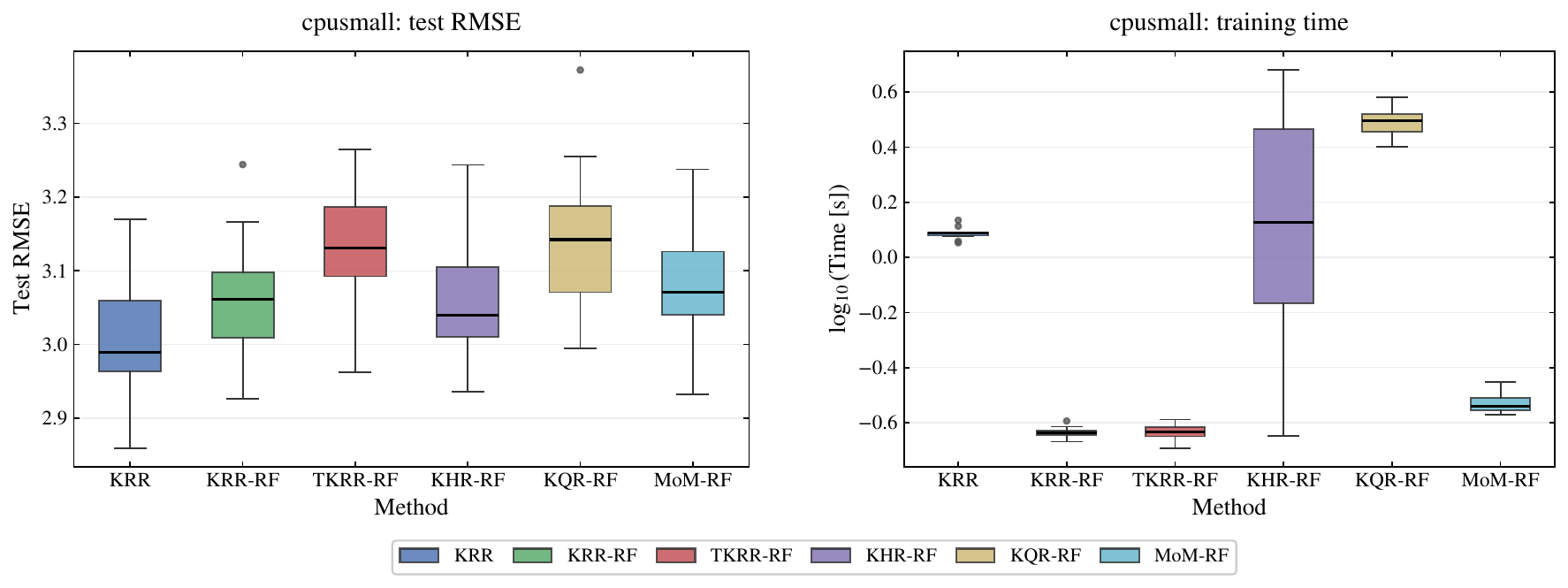}
    \caption{Predictive performance and time comparison of KRR, KRR-RF, and other robust methods on the CPU dataset.}
    \label{fig:krr-comparison}
\end{figure}

We also compare the predictive performance of KRR and KRR-RF with different other four robust methods considered in the simulation study: TKRR-RF, KHR-RF, KQR-RF, MoM-RF. All methods use a Gaussian RBF kernel, and employ random Fourier features with $M=512$. The regularization parameter $\lambda$ is selected using the validation set for each method. For TKRR-RF, the training responses are truncated at their empirical $97.5$-th percentile before fitting the model. The validation and test responses remain unchanged. For KHR-RF, KQR-RF, and MoM-RF, we follow the same algorithms as described in the simulation study. Figure \ref{fig:krr-comparison} presents the test RMSE and training time across 20 repeated data splits. The left panel shows that KRR achieves the lowest average RMSE. The performance of KRR-RF, KHR-RF, MoM-RF is comparable, while TKRR-RF and KQR-RF exhibit slightly higher test errors. However, as shown in the right panel, KRR-RF generally require less training time compared to standard KRR and other robust methods, highlighting the computational advantage of using random features and robust techniques. This is consistent with our conclusion in the simulation study.

At last, to check the noise characteristics of the cpusmall dataset, we analyze out-of-fold residuals obtained from KRR-RF estimator. We perform five-fold cross-fitting with five repeated random partitions. In each fold, KRR-RF is fitted using four subsets and evaluated on the held-out subset. Thus, every observation receives a prediction from a model that excludes that observation during training. We define the out-of-fold residuals as 
$$
\hat{e}_i=y_i-\widehat f^{(-k(i))}({\bfx}_i).
$$
We center the residuals by their median and standardize them using the MAD-based scale estimator. We then examine their empirical tail behavior using normal Q-Q plots, Hill's tail-index estimator, and tail exceedance probabilities.

The median out-of-fold RMSE over the five repetitions is $3.165$, and the median MAD-based scale is $2.368$. 
The standardized residuals are clearly heavier-tailed than Gaussian: $4.23\%$ of them exceed three robust scale 
units, while the Gaussian value is only $0.27\%$, about $15.7$ times smaller. 
The normal Q--Q plot shows a long upper tail, and the excess kurtosis is about $4.22$. 
The Hill estimates decrease from about $4.00$ to $3.08$ as the tail fraction increases from $2\%$ to $10\%$, so 
the estimated tail index depends on the threshold.
We further split the observations into four groups by the CPU run-queue size $\mathrm{runqsz}$, and compute the MAD-based  scale and the three-MAD exceedance probability within each group. The residual scale grows from $1.67$ in the lowest group to $2.88$ in the highest group, and the exceedance probabilities range from $2.56\%$ to $6.83\%$, all well above the Gaussian benchmark $0.27\%$. These results indicate that the noise in the cpusmall dataset is heteroscedastic and have much heavier tails than a Gaussian distribution.

\begin{figure}[t]
  \centering
  \includegraphics[width=\textwidth]{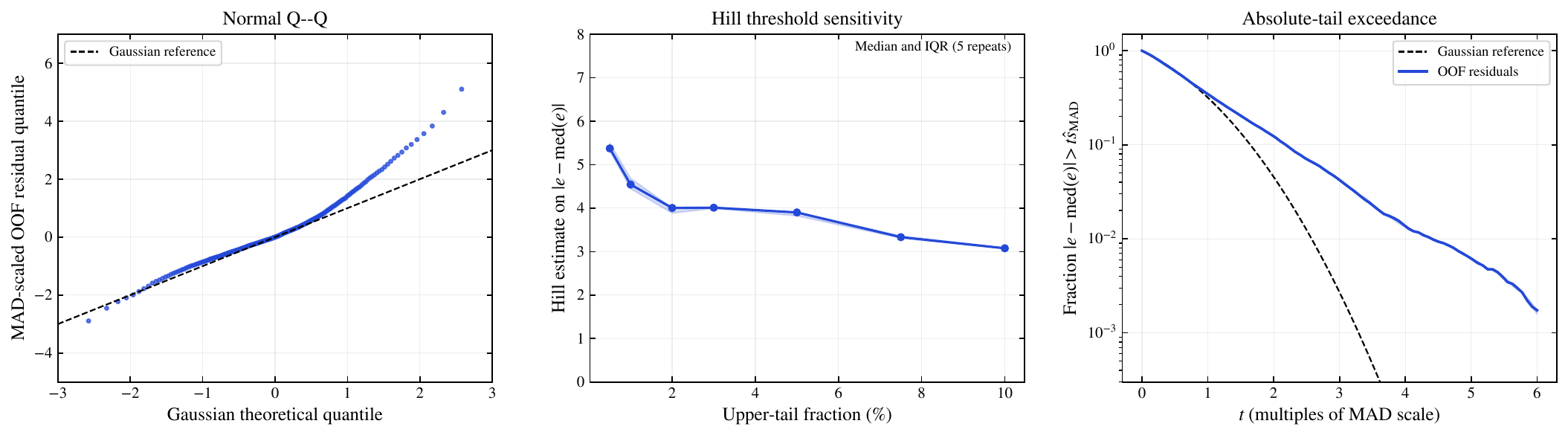}
  \caption{Normal Q-Q plot, Hill's tail-index estimator, and tail exceedance probabilities of the standardized out-of-fold residuals for the cpusmall dataset.}
\end{figure}

\newpage

\appendix

\section{Random Features Operators and Error Decomposition}

\noindent \textbf{Notation.} We denote $\|\cdot\|_2$, $\|\cdot\|$ and $\|\cdot\|_{HS}$ as the Euclidean norm, the operatorial norm and Hilbert-Schmidt norm. Let $\mathcal L$ be a Hilbert space, we use $\langle \cdot, \cdot \rangle_{\mathcal L}$ to denote its inner product, $\|\cdot\|_{\mathcal L}$ to denote its norm, and $\text{Tr}(\cdot)$ to denote the trace of an operator. Let $Q$ be a bounded self-adjoint linear operator on a separable Hilbert space $\mathcal L$, we denote $Q_\lambda$ as $Q_\lambda = Q + \lambda I$, where $\lambda>0$ and $I$ is the identity operator on $\mathcal L$. We denote $c, c_0, c_1,\ldots, C, C_0, C_1,\ldots$ as a universal constant whose value may differ from line to line.

\begin{definition}\label{def1}
For all $g \in L^{2}(\rho_{X})$, $\boldsymbol{\beta} \in \mathbb{R}^{M}$, $\alpha \in \mathbb{R}^{n}$ and $i \in \{1, \ldots, M\}$, we have
\begin{itemize}
\item $S_{M}: \mathbb{R}^{M} \to L^{2}(\rho_{X})$, \quad $(S_{M}\boldsymbol{\beta})(\cdot) = \phi_{M}(\cdot)^{\top}\boldsymbol{\beta}$,
\item $S_{M}^{*}: L^{2}(\rho_{X}) \to \mathbb{R}^{M}$, \quad $(S_{M}^{*}g)_{i} = \frac{1}{\sqrt{M}} \int_{X} \psi_{\omega_{i}}(\bfx) g(\bfx) d\rho_{X}(\bfx)$,
\item $\overline{S}_{M}^{*}: L^{2}(\rho_{X}) \to \mathbb{R}^{M}$, \quad $(\overline{S}_{M}^{*}g)_{i} = \frac{1}{n} \sum_{j=1}^{n} g(\bfx_j) \phi_{M}(\bfx_j)_i$,
\item $L_{M}: L^{2}(\rho_{X}) \to L^{2}(\rho_{X})$, \quad $(L_{M}g)(\cdot) = \int_{X} K_{M}(\cdot, \bfz) g(\bfz) d\rho_{X}(\bfz)$,
\item $C_{M}: \mathbb{R}^{M} \to \mathbb{R}^{M}$, \quad $C_{M} = \int_{X} \phi_{M}(\bfx) \phi_{M}(\bfx)^{\top} d\rho_{X}(\bfx)$,
\item $\widehat{C}_{M}: \mathbb{R}^{M} \to \mathbb{R}^{M}$, \quad $\widehat{C}_{M} = \frac{1}{n} \sum_{i=1}^{n} \phi_{M}(\bfx_i) \phi_{M}(\bfx_i)^{\top}$.
\end{itemize}
\end{definition} 

From \cite{rudi2017generalization} and Assumption \ref{ass1}, the linear operator $L$ is trace class and the linear operators $L_{M}, C_{M}, S_{M}, \widehat{C}_{M}, \widehat{S}_{M}$ are finite dimensional. It is easy to verify that $L_{M} = S_{M} S_{M}^{*}$, $C_{M} = S_{M}^{*} S_{M}$ and $\widehat{C}_{M} = \widehat{S}_{M}^{*} \widehat{S}_{M}= \overline{S}_{M}^{*} {S}_{M}$. And $L_{M}, C_{M}, \widehat{C}_{M}$ are self-adjoint and positive operators, with spectrum in $[0, \kappa^{2}]$. From these operators, we can write the estimator as
$$
\widehat{f}_{\lambda,M} = S_M\widehat{C}_{M,\lambda}^{-1}\widehat{S}_M^*\widehat{\mathbf{y}}.
$$
If
$
\widehat{\mathbf g}_n
=
\frac1{\sqrt{n}}
(g(\bfx_1),\ldots,g(\bfx_{n}))^\top,
$
then
$
\overline S_{M}^*g
=
\widehat{S}_{M}^*\widehat{\mathbf g}_n.
$

The following Proposition provides an error decomposition for the estimator $\widehat{f}_{\lambda,M}$ in terms of noise error, the population RF error, and the empirical fluctuation of the population RF error.

\begin{proposition}[Error decomposition]\label{lem:decomp}
Let the feature realization $\omega_1,\dots,\omega_M$ be fixed, then we can decompose the error as 
\begin{equation}\label{eq:decompbound}
\begin{aligned}
&\left\|\widehat{f}_{\lambda,M}-f_\rho\right\|_{L^2(\rho_X)} \leq \underbrace{\left\|\widehat{C}_{M,\lambda}^{-1/2}C_{M,\lambda}^{1/2}\right\|^2 \left\|C_{M,\lambda}^{-1/2}\widehat{S}_{M}^*\widehat{\boldsymbol{\varepsilon}}\right\|_2}_{\text{Noise error}}+  \underbrace{\left\|f_\rho-L_ML_{M,\lambda}^{-1}f_\rho\right\|_{L^2(\rho_X)}}_{\text{Population RF error}} \\
&+ \underbrace{\left\|\widehat{C}_{M,\lambda}^{-1/2}C_{M,\lambda}^{1/2}\right\|^2\left\|C_{M,\lambda}^{-1/2}\left(\overline S_{M}^*\bigl(f_\rho-L_ML_{M,\lambda}^{-1}f_\rho\bigr)- S_M^*\bigl(f_\rho-L_ML_{M,\lambda}^{-1}f_\rho\bigr) \right)\right\|_2}_{\text{Empirical RF fluctuation error}},
\end{aligned}
\end{equation}
where $\widehat{\boldsymbol{\varepsilon}}=\frac{1}{\sqrt{n}}(\varepsilon_1,\ldots,\varepsilon_n)^\top.$
\end{proposition}

\begin{proof}
Since $y_i=f_\rho(\bfx_i)+\varepsilon_i$,
$\widehat{\mathbf y}=\frac1{\sqrt n}(y_1,\dots,y_n)^\top$, and $\widehat{\boldsymbol{\varepsilon}}=\frac{1}{\sqrt{n}}(\varepsilon_1,\ldots,\varepsilon_n)^\top$, we have
$$
\widehat f_{\lambda,M}-f_\rho
=S_M\widehat C_{M,\lambda}^{-1}\widehat S_M^*\widehat{\mathbf y}-f_\rho
=\bigl(S_M\widehat C_{M,\lambda}^{-1}\overline S_M^*f_\rho-f_\rho\bigr)
+S_M\widehat C_{M,\lambda}^{-1}\widehat S_M^*\widehat{\boldsymbol{\varepsilon}},
$$
where we used $\widehat S_M^*\widehat{\mathbf y}
=\overline S_M^*f_\rho+\widehat S_M^*\widehat{\boldsymbol{\varepsilon}}$.
By the triangle inequality,
\begin{equation}\label{eq:proofnorm}
\|\widehat f_{\lambda,M}-f_\rho\|_{L^2(\rho_X)}
\le\bigl\|S_M\widehat C_{M,\lambda}^{-1}\overline S_M^*f_\rho-f_\rho\bigr\|_{L^2(\rho_X)}
+\bigl\|S_M\widehat C_{M,\lambda}^{-1}\widehat S_M^*\widehat{\boldsymbol{\varepsilon}}\bigr\|_{L^2(\rho_X)}.
\end{equation}
We bound the two terms separately.

\noindent \textbf{Noise error.} Since $\widehat S_M^*\widehat{\boldsymbol{\varepsilon}}\in\mathbb{R}^M$ and
$\widehat S_M^*\widehat{\boldsymbol{\varepsilon}}
=C_{M,\lambda}^{1/2}C_{M,\lambda}^{-1/2}\widehat S_M^*\widehat{\boldsymbol{\varepsilon}}$,
\begin{equation}\label{eq:proofnoise}
\begin{aligned}
\bigl\|S_M\widehat C_{M,\lambda}^{-1}\widehat S_M^*\widehat{\boldsymbol{\varepsilon}}\bigr\|_{L^2(\rho_X)}
&=\bigl\|S_M\widehat C_{M,\lambda}^{-1} C_{M,\lambda}^{1/2}
   C_{M,\lambda}^{-1/2}\widehat S_M^*\widehat{\boldsymbol{\varepsilon}}\bigr\|_{L^2(\rho_X)}\\
&\le\bigl\|S_M\widehat C_{M,\lambda}^{-1}C_{M,\lambda}^{1/2}\bigr\|
   \bigl\|C_{M,\lambda}^{-1/2}\widehat S_M^*\widehat{\boldsymbol{\varepsilon}}\bigr\|_2.
\end{aligned}
\end{equation}

\noindent \textbf{Population RF error and its empirical fluctuation.}
Recall that $Z^*f(ZZ^*)=f(Z^*Z)Z^*$ for any continuous spectral function $f$ and any compact operator $Z$. By Definition \ref{def1}, we have 
\begin{equation}\label{eq:proofpop}
L_{M} L_{M, \lambda}^{-1}=S_{M} S_{M}^{*}\left(S_{M} S_{M}^{*}+\lambda I\right)^{-1}=S_{M}\left(S_{M}^{*} S_{M}+\lambda I\right)^{-1} S_{M}^{*}=S_{M} C_{M, \lambda}^{-1} S_{M}^{*},
\end{equation}
and
\begin{equation}\label{eq:proofslack}
\begin{aligned}
S_M^*(f_\rho-L_ML_{M,\lambda}^{-1}f_\rho)& = S_M^*f_\rho- S_M^*S_MS_M^*L_{M,\lambda}^{-1}f_\rho =S_M^*f_\rho-C_MS_M^*L_{M,\lambda}^{-1}f_\rho\\
&= S_M^*f_\rho-C_M C_{M,\lambda}^{-1}S_M^*f_\rho
=\lambda C_{M,\lambda}^{-1}S_M^*f_\rho,
\end{aligned}
\end{equation}
where the last step uses $C_{M,\lambda}-C_{M}=\lambda I$.
Now we can write
\begin{equation}\label{eq:proofrecomb}
\begin{aligned}
&S_M\widehat C_{M,\lambda}^{-1}\overline S_M^*f_\rho-f_\rho
=S_M\widehat C_{M,\lambda}^{-1}
 \bigl(\overline S_M^*f_\rho-\widehat C_{M,\lambda}C_{M,\lambda}^{-1}S_M^*f_\rho\bigr)
 +\bigl(S_MC_{M,\lambda}^{-1}S_M^*f_\rho-f_\rho\bigr)\\
=&S_M\widehat C_{M,\lambda}^{-1}
 \bigl(\overline S_M^*f_\rho-\widehat C_MC_{M,\lambda}^{-1}S_M^*f_\rho
      -\lambda C_{M,\lambda}^{-1}S_M^*f_\rho\bigr)
 +\bigl(L_ML_{M,\lambda}^{-1}f_\rho-f_\rho\bigr)\\
 =&S_M\widehat C_{M,\lambda}^{-1}
 \bigl(\overline S_M^*f_\rho-\widehat C_MC_{M,\lambda}^{-1}S_M^*f_\rho
      -S_M^*(f_\rho-L_ML_{M,\lambda}^{-1}f_\rho)\bigr)
 +\bigl(L_ML_{M,\lambda}^{-1}f_\rho-f_\rho\bigr)\\
 = &S_M\widehat C_{M,\lambda}^{-1}
 \bigl(\overline S_{M}^*\bigl(f_\rho-L_ML_{M,\lambda}^{-1}f_\rho\bigr)-S_M^*(f_\rho-L_ML_{M,\lambda}^{-1}f_\rho)\bigr)
 +\bigl(L_ML_{M,\lambda}^{-1}f_\rho-f_\rho\bigr)
\end{aligned}
\end{equation}
where the second equality uses $\widehat C_{M,\lambda}=\widehat C_M+\lambda I$ and \eqref{eq:proofpop}, the third equality uses \eqref{eq:proofslack}. The last equality uses $\widehat C_MC_{M,\lambda}^{-1}S_M^*f_\rho=\overline S_M^*L_ML_{M,\lambda}^{-1}f_\rho$, since every $w\in\mathbb{R}^M$, there holds that
\begin{align*}
\bigl\langle w,\widehat C_MC_{M,\lambda}^{-1}S_M^*f_\rho\bigr\rangle_2
&=\bigl\langle \widehat C_Mw,C_{M,\lambda}^{-1}S_M^*f_\rho\bigr\rangle_2
 =\bigl\langle \overline S_M^*S_Mw,C_{M,\lambda}^{-1}S_M^*f_\rho\bigr\rangle_2\\
&=\bigl\langle S_Mw,S_MC_{M,\lambda}^{-1}S_M^*f_\rho\bigr\rangle_{L^2(\rho_X)}
 =\bigl\langle S_Mw,L_ML_{M,\lambda}^{-1}f_\rho\bigr\rangle_{L^2(\rho_X)}\\
&=\bigl\langle w,\overline S_M^*(L_ML_{M,\lambda}^{-1}f_\rho)\bigr\rangle_2,
\end{align*}
where we use $\widehat C_M=\overline S_M^*S_M$, and \eqref{eq:proofpop}.

Similar with \eqref{eq:proofnoise}, we can obtain that 
\begin{equation}\label{eq:empirical}
\begin{aligned}
&\bigl\|S_M\widehat C_{M,\lambda}^{-1}\overline S_M^*f_\rho-f_\rho\bigr\|_{L^2(\rho_X)} \leq \bigl\|f_\rho-L_ML_{M,\lambda}^{-1}f_\rho\bigr\|_{L^2(\rho_X)} \\
& \qquad +\left\|S_M\widehat{C}_{M,\lambda}^{-1}C_{M,\lambda}^{1/2}\right\|\left\|C_{M,\lambda}^{-1/2}\left(\overline S_{M}^*\bigl(f_\rho-L_ML_{M,\lambda}^{-1}f_\rho\bigr)- S_M^*\bigl(f_\rho-L_ML_{M,\lambda}^{-1}f_\rho\bigr) \right)\right\|_2.
\end{aligned}
\end{equation}

Finally, we show that $\left\|S_M\widehat{C}_{M,\lambda}^{-1}C_{M,\lambda}^{1/2}\right\| \leq \left\|\widehat{C}_{M,\lambda}^{-1/2}C_{M,\lambda}^{1/2}\right\|^2$,  
\begin{align}\label{eq:1}
\left\|S_M\widehat{C}_{M,\lambda}^{-1}C_{M,\lambda}^{1/2}\right\| \leq \left\|S_M{C}_{M,\lambda}^{-1/2}\right\| \left\|{C}_{M,\lambda}^{1/2}\widehat{C}_{M,\lambda}^{-1/2}\right\| \left\|\widehat{C}_{M,\lambda}^{-1/2}C_{M,\lambda}^{1/2}\right\| \leq \left\|\widehat{C}_{M,\lambda}^{-1/2}C_{M,\lambda}^{1/2}\right\|^2,
\end{align}
where we use $\left\|S_M{C}_{M,\lambda}^{-1/2}\right\| = \|{C}_{M,\lambda}^{-1/2}S_M^*S_M{C}_{M,\lambda}^{-1/2}\|^{1/2} =\|C_MC_{M,\lambda}^{-1}\|^{1/2} \leq 1 $.

Combining \eqref{eq:proofnorm}, \eqref{eq:proofnoise},
\eqref{eq:empirical}, and \eqref{eq:1} gives \eqref{eq:decompbound}. Thus we complete the proof.
\end{proof}

\section{Population RF Error}

\begin{lemma}\label{lem:fmrf}
Under Assumptions \ref{ass1} and \ref{ass3}, the following holds for any $\lambda > 0$ and $M \in \mathbb{N}$:
\begin{enumerate}
\item[(i)] (Misspecified case) if $0<r< \tfrac12$, then
\[
\left\|f_\rho-L_ML_{M,\lambda}^{-1}f_\rho\right\|_{L^2(\rho_X)}
\leq R\lambda^r\|L_{M,\lambda}^{-1/2}L_{K,\lambda}^{1/2}\|^{2r};
\]
\item[(ii)] (Well-specified case) if $\tfrac12 \le r\leq 1$, then
\begin{align*}
&\left\|f_\rho-L_ML_{M,\lambda}^{-1}f_\rho\right\|_{L^2(\rho_X)}\\
\leq & R\lambda^r+ R\sqrt{\lambda}  \|L_{M,\lambda}^{-1/2}L_{K,\lambda}^{1/2}\| \|L_{K,\lambda}^{-1/2}(L_K-L_M)\|^{2r-1} \|L_{K,\lambda}^{-1/2}(L_K-L_M)L_{K,\lambda}^{-1/2}\|^{2-2r}.
\end{align*}
\end{enumerate}
\end{lemma}

\begin{proof}
For the misspecified case when $0<r\leq \frac12$, from Assumption \ref{ass3}, $f_\rho = L_K^r h_\rho$ for some $h_\rho \in L^2(\rho_X)$, and $I-L_ML_{M,\lambda}^{-1}=\lambda L_{M,\lambda}^{-1}$, then we have
$$
f_\rho-L_ML_{M,\lambda}^{-1}f_\rho
=\lambda L_{M,\lambda}^{-1}L_K^r h_\rho
=\lambda L_{M,\lambda}^{-(1-r)}\bigl(L_{M,\lambda}^{-r}L_{K,\lambda}^{r}\bigr)\bigl(L_{K,\lambda}^{-r}L_{K}^{r}\bigr)h_\rho. 
$$
It then follows that 
\begin{align*}
\|f_\rho-L_ML_{M,\lambda}^{-1}f_\rho\|_{L^2(\rho_X)}&\leq \lambda \|L_{M,\lambda}^{-(1-r)}\| \|L_{M,\lambda}^{-r}L_{K,\lambda}^r\| \|L_{K,\lambda}^{-r}L_{K}^{r}\| \|h_\rho\|_{L^2(\rho_X)}\\
&\leq \lambda \cdot \lambda^{r-1} \cdot \|L_{M,\lambda}^{-1/2}L_{K,\lambda}^{1/2}\|^{2r} \|L_{K,\lambda}^{-1/2}L_K^{1/2}\|^{2r} \cdot R
\leq R\lambda^r\|L_{M,\lambda}^{-1/2}L_{K,\lambda}^{1/2}\|^{2r}.
\end{align*}
where we use $\|L_{M,\lambda}^{-(1-r)}\|\leq \lambda^{r-1}$ for $r \leq 1$ and the Cordes inequality $\|A^{\theta}B^{\theta}\|\leq\|AB\|^{\theta}$ for positive bounded operators $A,B$ and $0\leq\theta\leq1$, applied with $A=L_{M,\lambda}^{-1/2}$, $B=L_{K,\lambda}^{1/2}$ and $\theta=2r <1$ and $A=L_{K,\lambda}^{-1}$, $B=L_{K}^{1}$ and $\theta=r\leq1$, and $\|L_{K,\lambda}^{-1}L_K\|\leq 1$.

For the well-specified case when $r\geq \frac12$, note that 
$$
f_\rho-L_ML_{M,\lambda}^{-1}f_\rho
=\bigl(f_\rho-LL_{K,\lambda}^{-1}f_\rho\bigr)
+\bigl(LL_{K,\lambda}^{-1}-L_ML_{M,\lambda}^{-1}\bigr)f_\rho,
$$
we use \citet[Lemmas~4 and 5]{rudi2017generalization} to bound the two terms separately. Consequently, by the triangle inequality, we obtain that 
\begin{align*}
&\|f_\rho-L_ML_{M,\lambda}^{-1}f_\rho\|_{L^2(\rho_X)}
\leq \|f_\rho-LL_{K,\lambda}^{-1}f_\rho\|_{L^2(\rho_X)} + \|(LL_{K,\lambda}^{-1}-L_ML_{M,\lambda}^{-1})f_\rho\|_{L^2(\rho_X)} \\
\leq & R\lambda^r+ R\sqrt{\lambda}  \|L_{M,\lambda}^{-1/2}L_{K,\lambda}^{1/2}\| \|L_{K,\lambda}^{-1/2}(L_K-L_M)\|^{2r-1} \|L_{K,\lambda}^{-1/2}(L_K-L_M)L_{K,\lambda}^{-1/2}\|^{2-2r}.
\end{align*}

Combining the bounds for the well-specified and misspecified cases completes the proof.
\end{proof}

We introduce the effective dimension induced by kernel $K_M$. For any $\lambda>0$, we define ${\cal N}_M(\lambda)$ as
$$
{\cal N}_M(\lambda)=\text{Tr}(L_{M,\lambda}^{-1}L_M)=\text{Tr}(C_{M,\lambda}^{-1}C_M),
$$
where the second equality uses the cyclic property of the trace and \eqref{eq:proofpop},
\begin{align*}
\text{Tr}(L_{M,\lambda}^{-1}L_M)=\text{Tr}(L_{M,\lambda}^{-1}S_MS_M^*)=\text{Tr}(S_M^*L_{M,\lambda}^{-1}S_M)=\text{Tr}(C_{M,\lambda}^{-1}S_M^*S_M)=\text{Tr}(C_{M,\lambda}^{-1}C_M).
\end{align*}

\begin{lemma}\label{lem:feat}
Under Assumption~\ref{ass1}, if $0<\lambda\leq\|L_K\|$ and
$0<\eta<1/4$, we set $t_\eta=\log\frac{4\kappa^{2}}{\lambda\eta}$,
and suppose the number of random features satisfies
\begin{equation}\label{eq:Mbudget}
M \geq c_0\bigl(1+\mathcal F_\infty(\lambda)\bigr)t_\eta,
\end{equation}
with $c_0\geq 36$. Then, with probability at least $1-4\eta$ over the
random feature realization $\omega_1,\dots,\omega_M$, the following hold
simultaneously:
\begin{enumerate}
\item[(a)] $\bigl\|L_{K,\lambda}^{-1/2}(L_K-L_M)L_{K,\lambda}^{-1/2}\bigr\| \leq\min\bigl\{\tfrac12, C\sqrt{\frac{\mathcal F_\infty(\lambda)t_\eta}{M}}\bigr\}$, hence $\tfrac12 L_{K,\lambda}\preceq L_{M,\lambda}\preceq\tfrac32L_{K,\lambda}$ and
$\bigl\|L_{M,\lambda}^{-1/2}L_{K,\lambda}^{1/2}\bigr\|\leq\sqrt2$;
\item[(b)]
$\bigl\|L_{K,\lambda}^{-1/2}(L_K-L_M)\bigr\|
\leq C\kappa\Bigl(
\frac{\sqrt{\mathcal F_\infty(\lambda)}t_\eta}{M}
+\sqrt{\frac{\mathcal {\cal N}(\lambda)t_\eta}{M}}\Bigr)$;
\item[(c)] $\mathcal N_M(\lambda)\leq 4\mathcal {\cal N}(\lambda)$.
\end{enumerate}
\end{lemma}

\begin{proof}
We bound the three events separately and then take the union bound.

\noindent \textbf{Event for (a).}
Apply \citet[Proposition~6]{rudi2017generalization} with
$Q=L_K$, $U_i=\psi_{\omega_i}$, $n=M$ and failure level $\eta$,
with probability at least $1-2\eta$,
\begin{equation}\label{eq:proofrel}
\bigl\|L_{K,\lambda}^{-1/2}(L_K-L_M)L_{K,\lambda}^{-1/2}\bigr\|
\le
\frac{2\beta\bigl(1+\mathcal F_\infty(\lambda)\bigr)}{3M}
+\sqrt{\frac{2\beta\mathcal F_\infty(\lambda)}{M}},
\qquad
\beta=\log\frac{4\mathrm{Tr}(L_K)}{\lambda\eta}.
\end{equation}
Since $\mathrm{Tr}(L_K)\le\kappa^{2}$ and
$\lambda\le\|L_K\|\le\kappa^{2}$, we have
$\beta\le\log\frac{4\kappa^{2}}{\lambda\eta}=t_\eta$, and in
particular $t_\eta\ge\log\frac{4\kappa^{2}}{\|L_K\|\eta}\ge
\log\frac{4}{\eta}\ge\log 16>2$ for $\eta< 1/4$.
Under the budget \eqref{eq:Mbudget} with $c_0\ge36$,
\begin{align*}
\sqrt{\frac{2\beta\mathcal F_\infty(\lambda)}{M}}
\le \sqrt{\frac{2t_{\eta}\mathcal F_\infty(\lambda)}{M}} &\le
\sqrt{\frac{2\mathcal F_\infty(\lambda)t_\eta}
{36(1+\mathcal F_\infty(\lambda))t_\eta}}
\le\sqrt{\frac1{18}},\\
\frac{2\beta(1+\mathcal F_\infty(\lambda))}{3M}
&\le
\frac{2(1+\mathcal F_\infty(\lambda))t_\eta}
{3\cdot36(1+\mathcal F_\infty(\lambda))t_\eta}
=\frac1{54}.
\end{align*}
Moreover, note that
$1+\mathcal F_\infty(\lambda) \le 3 \mathcal F_\infty(\lambda)$,
where we use $\mathcal F_\infty(\lambda)\ge\mathcal {\cal N}(\lambda)\ge 1/2$, since $\lambda\le\|L_K\|$ implies
$\mathcal {\cal N}(\lambda)\ge\sigma_1/(\sigma_1+\lambda)\ge 1/2$ with
$\sigma_1=\|L_K\|$. Hence we have 
\[
\frac{2\beta(1+\mathcal F_\infty(\lambda))}{3M} \le \frac{2\mathcal F_\infty(\lambda)t_\eta}{M}. 
\le\frac{1}{3}\sqrt{\frac{\mathcal F_\infty(\lambda)t_\eta}{M}},
\]
where the last inequality uses the fact that $\frac{\mathcal F_\infty(\lambda)t_\eta}{M} \le \frac{1}{36}$ under the budget \eqref{eq:Mbudget} with $c_0\ge36$.
Combining the above we obtain that $\bigl\|L_{K,\lambda}^{-1/2}(L_K-L_M)L_{K,\lambda}^{-1/2}\bigr\| \leq\min\bigl\{\tfrac12, c_1\sqrt{\frac{\mathcal F_\infty(\lambda)t_\eta}{M}}\bigr\}$.

To derive the order bounds in (a), note that 
$$
L_M=L_{K,\lambda}^{1/2}(I-L_{K,\lambda}^{-1/2}(L_K-L_M)L_{K,\lambda}^{-1/2})L_{K,\lambda}^{1/2}-\lambda I
\succeq L_K-\tfrac12L_{K,\lambda},
$$ 
where we use $\bigl\|L_{K,\lambda}^{-1/2}(L_K-L_M)L_{K,\lambda}^{-1/2}\bigr\|\le \frac12$. Hence
$L_{M,\lambda}\succeq\tfrac12L_{K,\lambda}$, and the upper bound is
analogous. Finally, taking inverses reverses the order,
$L_{M,\lambda}^{-1}\preceq2L_{K,\lambda}^{-1}$, so
\[
\bigl\|L_{M,\lambda}^{-1/2}L_{K,\lambda}^{1/2}\bigr\|^{2}
=\bigl\|L_{K,\lambda}^{1/2}L_{M,\lambda}^{-1}L_{K,\lambda}^{1/2}\bigr\|
\le 2.
\]
Thus we have $\bigl\|L_{M,\lambda}^{-1/2}L_{K,\lambda}^{1/2}\bigr\| \le \sqrt{2}$.

\noindent \textbf{Event for (b).}
Apply \citet[Proposition~5]{rudi2017generalization} with
$U_i=L_{K,\lambda}^{-1/2}\psi_{\omega_i}$, $z_i=\psi_{\omega_i}$
(so that $\|z_i\|_\infty\le\kappa$),
$Q=L_K$, $T=L_K$, $T_M=L_M$ and failure level $\eta$, with
probability at least $1-\eta$,
\[
\bigl\|L_{K,\lambda}^{-1/2}(L_K-L_M)\bigr\|_{HS}
\le
\frac{4\sqrt{\mathcal F_\infty(\lambda)}\kappa\log\frac2\eta}{M}
+\sqrt{\frac{4\kappa^{2}\mathcal {\cal N}(\lambda)\log\frac2\eta}{M}}.
\]
Since $\|\cdot\|\le\|\cdot\|_{HS}$ and
$\log\frac2\eta<\log\frac{4\kappa^{2}}{\lambda\eta}=t_\eta$
with $\lambda\le\kappa^{2}$, both terms are bounded by
$4\kappa\Bigl(
\frac{\sqrt{\mathcal F_\infty(\lambda)}t_\eta}{M}
+\sqrt{\frac{\mathcal {\cal N}(\lambda)t_\eta}{M}}\Bigr)$, which gives (b).

\noindent \textbf{Event for (c).}
This is \citet[Proposition~10]{rudi2017generalization}, which states
that under the budget \eqref{eq:Mbudget} (their condition
$M\ge(4+18\mathcal F_\infty(\lambda))\log\frac{12\kappa^{2}}{\lambda\eta}$
is implied by \eqref{eq:Mbudget} with $c_0\ge36$ and
$t_\eta\ge\log 16> \log 3$, since $12\le 16\le 4\kappa^{2}/(\lambda\eta)$),
with probability at least $1-\eta$,
\[
\bigl|\mathcal N_M(\lambda)-\mathcal {\cal N}(\lambda)\bigr|
\le 1.55\mathcal {\cal N}(\lambda),
\]
hence $\mathcal N_M(\lambda)\le 2.55\mathcal {\cal N}(\lambda)\le4\mathcal {\cal N}(\lambda)$.

Taking the union of the three events, all
statements hold simultaneously with probability at least $1-4\eta$. Thus we conclude the proof.
\end{proof}

\begin{proposition}\label{prop:mergedbudget}
Under Assumptions \ref{ass1}, \ref{ass3}, \ref{ass4} and \ref{ass7}, if $0<\lambda\leq\|L_K\|$ and $0<\eta<1/4$, we set $t_\eta=\log\frac{4\kappa^{2}}{\lambda\eta}$.
\begin{itemize}
\item[(i)] (Misspecified case) If $0<r<\frac12$ and the number of random features satisfies
\begin{equation}\label{eq:budgetlow}
M\ge C\lambda^{-\zeta}t_\eta,
\end{equation}
then with probability at least $1-4\eta$ over the
random feature realization $\omega_1,\dots,\omega_M$,
\[
\bigl\|f_\rho-L_ML_{M,\lambda}^{-1}f_\rho\bigr\|_{L^2(\rho_X)}
\le2^rR\lambda^r.
\]
\item[(ii)] (Well-specified case) If $\frac12\le r\le1$ and the number of random features satisfies 
\begin{equation}\label{eq:interpbudget}
M\ge C\lambda^{-\{\zeta+(1+\gamma-\zeta)(2r-1)\}}t_\eta,
\end{equation}
then with probability at least $1-4\eta$ over the
random feature realization $\omega_1,\dots,\omega_M$,
\[
\bigl\|f_\rho-L_ML_{M,\lambda}^{-1}f_\rho\bigr\|_{L^2(\rho_X)}
\le CR\lambda^r.
\]
\end{itemize}
\end{proposition}

\begin{proof}
Throughout the proof we work on the event of
Lemma~\ref{lem:feat}, which has probability at least
$1-4\eta$ over the random feature realization $\omega_1,\dots,\omega_M$.
On this event, the bounds (a), (b), and (c) from Lemma~\ref{lem:feat} hold.

For the misspecified case when $0<r< \frac12$, from Lemmas \ref{lem:fmrf} and \ref{lem:feat}, when the number of random features $M$ satisfies $M \geq C\lambda^{-\zeta}t_\eta \geq c_0\bigl(1+\mathcal F_\infty(\lambda)\bigr)t_\eta$ from Assumption \ref{ass7} and $C$ is large enough, we have 
\begin{align*}
\left\|f_\rho-L_ML_{M,\lambda}^{-1}f_\rho\right\|_{L^2(\rho_X)}
\leq R\lambda^r\|L_{M,\lambda}^{-1/2}L_{K,\lambda}^{1/2}\|^{2r} \leq 2^r R\lambda^r.
\end{align*}

For the well-specified case when $r \geq \frac12$, from Lemmas \ref{lem:fmrf} and \ref{lem:feat}, when $M \geq C\lambda^{-\zeta}t_\eta \geq c_0\bigl(1+\mathcal F_\infty(\lambda)\bigr)t_\eta$ from Assumption \ref{ass7}, we have 
\begin{equation}\label{eq:well-specified-bound}
\begin{aligned}
&\left\|f_\rho-L_ML_{M,\lambda}^{-1}f_\rho\right\|_{L^2(\rho_X)}\\
\leq & R\lambda^r+ R\sqrt{\lambda}  \|L_{M,\lambda}^{-1/2}L_{K,\lambda}^{1/2}\| \|L_{K,\lambda}^{-1/2}(L_K-L_M)\|^{2r-1} \|L_{K,\lambda}^{-1/2}(L_K-L_M)L_{K,\lambda}^{-1/2}\|^{2-2r}\\
\leq & R\lambda^r+ C R\sqrt{2\lambda} \kappa^{2r-1}\left(
\frac{\sqrt{\mathcal F_\infty(\lambda)}t_\eta}{M}
+\sqrt{\frac{\mathcal {\cal N}(\lambda)t_\eta}{M}}\right)^{2r-1}\sqrt{\frac{\mathcal F_\infty(\lambda)t_\eta}{M}}^{2-2r}\\
\leq & R\lambda^r+ C R\sqrt{2\lambda} \kappa^{2r-1}\left(\sqrt{\mathcal F_\infty(\lambda)}\left(\frac{t_\eta}{M}\right)^{r} +\sqrt{\frac{{\cal N}(\lambda)^{2r-1}\mathcal F_\infty(\lambda)^{2-2r}t_\eta}{M}}
\right)
\end{aligned}
\end{equation}
where we use the inequality $(x+y)^{2r-1}\le x^{2r-1}+y^{2r-1}$ for $x,y\ge0$ to split the bound into two branches when $0\le 2r-1 \le 1$. 

On the one hand, when the number of random features $M$ satisfies 
\begin{align}\label{eq:N-branch-M-condition}
M \geq C\lambda^{-\{\zeta+(1+\gamma-\zeta)(2r-1)\}}t_\eta \geq C {\cal N}(\lambda)^{2r-1}F_\infty(\lambda)^{2-2r}\lambda^{1-2r}t_\eta,
\end{align}
where the last inequality follows from the bounds ${\cal N}(\lambda)\le C\lambda^{-\gamma}$ and $F_\infty(\lambda)\le F\lambda^{-\zeta}$ from Assumptions \ref{ass4} and \ref{ass7}. Thus we have 
\begin{align}\label{eq:N-branch-bound}
C R\sqrt{2\lambda} \kappa^{2r-1} 
\sqrt{\frac{{\cal N}(\lambda)^{2r-1}\mathcal F_\infty(\lambda)^{2-2r}t_\eta}{M}} \leq C R \lambda^r.
\end{align}
On the other hand, when the number of random features $M$ satisfies 
\begin{align}\label{eq:N-branch-M-condition_1}
M \geq C\lambda^{-\frac{2r-1+\zeta}{2r}} t_\eta \geq C F_\infty(\lambda)^{1/2r}\lambda^{\frac{1-2r}{2r}} t_\eta, 
\end{align}
where the last inequality follows from the bound $F_\infty(\lambda)\le F\lambda^{-\zeta}
$ from Assumptions \ref{ass7}. Thus we have 
\begin{align}\label{eq:N-branch-bound_1}
C R\sqrt{2\lambda} \kappa^{2r-1} \sqrt{\mathcal F_\infty(\lambda)}\left(\frac{t_\eta}{M}\right)^{r} \leq C R \lambda^r.
\end{align}
Note that $\zeta+(1+\gamma-\zeta)(2r-1)\ge \frac{2r-1+\zeta}{2r}$ for $\zeta \leq 1$, $\gamma\geq 0$, and $r\geq 1/2$, so \eqref{eq:N-branch-M-condition} is the more stringent condition compared to \eqref{eq:N-branch-M-condition_1}. For $\zeta \le 1$ and $r\geq 1/2$, $\zeta+(1+\gamma-\zeta)(2r-1)\ge\zeta$ also implies that \eqref{eq:N-branch-M-condition} is the more stringent condition compared to \eqref{eq:budgetlow}.

Plugging \eqref{eq:N-branch-bound} and \eqref{eq:N-branch-bound_1} into \eqref{eq:well-specified-bound}, we obtain the desired result
$$
\bigl\|f_\rho-L_ML_{M,\lambda}^{-1}f_\rho\bigr\|_{L^2(\rho_X)}
\le CR\lambda^r.
$$ 

Combining the bounds for the well-specified and misspecified cases completes the proof.
\end{proof}

\section{Empirical RF Fluctuation Error}\label{app:empirical_rf_fluctuation_error}

\begin{lemma}[Variational leverage identity]\label{lem:varlev}
For a fixed random feature realization $\omega_1,\dots,\omega_M$ and
$\rho_X$-almost every $\bfx\in\mathcal X$,
\begin{equation}\label{eq:varlev}
\bigl\|C_{M,\lambda}^{-1/2}\phi_M(\bfx)\bigr\|_2^{2}
=\inf_{h\in L^{2}(\rho_{X})}
\left\{\|h\|_{L^{2}(\rho_{X})}^{2}
+\frac{1}{\lambda}\bigl\|\phi_M(\bfx)-S_M^{*}h\bigr\|_2^{2}\right\}.
\end{equation}
The minimum is attained at
$h_0=L_{M,\lambda}^{-1}S_M\phi_M(\bfx)$.
\end{lemma}

\begin{proof}
Fix $\bfx\in\mathcal X$ and set
$h_0=L_{M,\lambda}^{-1}S_M\phi_M(\bfx)\in L^{2}(\rho_{X})$, which is
well defined since $S_M\phi_M(\bfx)$ belongs to the range of $L_M$
and $L_{M,\lambda}^{-1}$ is bounded.
We show that $h_0$ attains the infimum and that the optimal value equals
$\bigl\|C_{M,\lambda}^{-1/2}\phi_M(\bfx)\bigr\|_2^{2}$.

Recall that $Z^*f(ZZ^*)=f(Z^*Z)Z^*$ for any continuous spectral function $f$ and any compact operator $Z$. Let $f(\cdot)=(\cdot+\lambda I)^{-1}$, then we have $S_M^{*}L_{M,\lambda}^{-1}=C_{M,\lambda}^{-1}S_M^{*}$ and $L_{M,\lambda}^{-1}S_M=S_M C_{M,\lambda}^{-1}$. It follows that 
\begin{equation}\label{eq:varlevres}
\begin{aligned}
\phi_M(\bfx)-S_M^{*}h_0 &= \phi_M(\bfx)-S_M^{*}L_{M,\lambda}^{-1}S_M\phi_M(\bfx) =  \phi_M(\bfx)-C_{M,\lambda}^{-1}S_M^{*}S_M\phi_M(\bfx)\\
&=\phi_M(\bfx)-C_{M,\lambda}^{-1}C_M\phi_M(\bfx)
=\lambda C_{M,\lambda}^{-1}\phi_M(\bfx),
\end{aligned}
\end{equation}

Let $h\in L^{2}(\rho_{X})$ be arbitrary and write $h=h_0+\Delta$ with
$\Delta\in L^{2}(\rho_{X})$. By \eqref{eq:varlevres},
\[
\phi_M(\bfx)-S_M^{*}h
=\lambda C_{M,\lambda}^{-1}\phi_M(\bfx)-S_M^{*}\Delta,
\]
and therefore
\begin{equation}\label{eq:varlevexpand}
\begin{aligned}
&\|h\|_{L^{2}(\rho_{X})}^{2}
+\frac{1}{\lambda}\bigl\|\phi_M(\bfx)-S_M^{*}h\bigr\|^{2}_2 =\|h_0\|_{L^{2}(\rho_{X})}^{2}
+2\langle h_0,\Delta\rangle_{L^{2}(\rho_{X})}
+\|\Delta\|_{L^{2}(\rho_{X})}^{2} \notag\\
&\quad+\frac{1}{\lambda}\left(
\lambda^{2}\bigl\|C_{M,\lambda}^{-1}\phi_M(\bfx)\bigr\|_2^{2}
-2\lambda\bigl\langle C_{M,\lambda}^{-1}\phi_M(\bfx),S_M^{*}\Delta\bigr\rangle_2
+\bigl\|S_M^{*}\Delta\bigr\|^{2}_2\right)\\
&=\|h_0\|_{L^{2}(\rho_{X})}^{2}
+\lambda\bigl\|C_{M,\lambda}^{-1}\phi_M(\bfx)\bigr\|_2^{2}
+2\Bigl\langle h_0-S_MC_{M,\lambda}^{-1}\phi_M(\bfx),
\Delta\Bigr\rangle_{L^{2}(\rho_{X})}
+\|\Delta\|_{L^{2}(\rho_{X})}^{2}
+\frac{1}{\lambda}\bigl\|S_M^{*}\Delta\bigr\|_2^{2}\\
&=\|h_0\|_{L^{2}(\rho_{X})}^{2}
+\lambda\bigl\|C_{M,\lambda}^{-1}\phi_M(\bfx)\bigr\|_2^{2}+\|\Delta\|_{L^{2}(\rho_{X})}^{2}
+\frac{1}{\lambda}\bigl\|S_M^{*}\Delta\bigr\|_2^{2}
\end{aligned}
\end{equation}
The last two terms are nonnegative and vanish for $\Delta=0$, so the
infimum is attained at $h=h_0$.

Setting $\Delta=0$ in \eqref{eq:varlevexpand} and using
$S_M^{*}S_M=C_M$, we have
\begin{align*}
\|h_0\|_{L^{2}(\rho_{X})}^{2}+
\lambda\bigl\|C_{M,\lambda}^{-1}\phi_M(\bfx)\bigr\|_2^{2}
&=\bigl\langle C_{M,\lambda}^{-1}\phi_M(\bfx),
S_M^{*}S_MC_{M,\lambda}^{-1}\phi_M(\bfx)\bigr\rangle_{2}
+\lambda\bigl\langle C_{M,\lambda}^{-1}\phi_M(\bfx),
C_{M,\lambda}^{-1}\phi_M(\bfx)\bigr\rangle_2\\
&=\bigl\langle C_{M,\lambda}^{-1}\phi_M(\bfx),
(C_M+\lambda I)C_{M,\lambda}^{-1}\phi_M(\bfx)\bigr\rangle_{2}\\
&=\bigl\langle C_{M,\lambda}^{-1}\phi_M(\bfx),\phi_M(\bfx)\bigr\rangle_{2}
=\bigl\|C_{M,\lambda}^{-1/2}\phi_M(\bfx)\bigr\|_2^{2},
\end{align*}
which completes the proof.
\end{proof}

\begin{lemma}\label{lem:comparison}
Under Assumption \ref{ass5}, if $0<\lambda\le 1$ and $\alpha_0 < \alpha \le 1$, then for a fixed random feature realization $\omega_1,\dots,\omega_M$ and
$\rho_X$-almost every $\bfx\in\mathcal X$, 
\begin{equation}\label{eq:comparison}
\begin{aligned}
&\bigl\|L_{K,\lambda}^{-1}K_{\bfx}\bigr\|_{L^2(\rho_X)}^{2}
+\frac{1}{\lambda}\mathbb{E}_{\omega}
\Bigl|\psi_\omega(\bfx)
-\bigl\langle L_{K,\lambda}^{-1}K_{\bfx},\psi_\omega\bigr\rangle_{L^2(\rho_X)}\Bigr|^{2}\\
=&\frac{K(\bfx,\bfx)
-\bigl\langle L_{K,\lambda}^{-1}K_{\bfx},K_{\bfx}\bigr\rangle_{L^2(\rho_X)}}{\lambda}
=\sum_{j}\frac{\mu_j}{\mu_j+\lambda}e_j(\bfx)^{2} \le M_{\alpha}\lambda^{-\alpha}.
\end{aligned}
\end{equation}
\end{lemma}

\begin{proof}
Expanding the second expected term in the left side of \eqref{eq:comparison}, we have
\begin{equation}\label{eq:residexpand}
\begin{aligned}
&\frac{1}{\lambda}\mathbb{E}_{\omega}\Bigl|\psi_\omega(\bfx)
-\bigl\langle L_{K,\lambda}^{-1}K_{\bfx},\psi_\omega\bigr\rangle_{L^2(\rho_X)}\Bigr|^{2}\\
=&\frac{1}{\lambda}\mathbb{E}_{\omega}\psi_\omega(\bfx)^{2}
-\frac{2}{\lambda}\mathbb{E}_{\omega}\Bigl[\psi_\omega(\bfx)
 \bigl\langle L_{K,\lambda}^{-1}K_{\bfx},\psi_\omega\bigr\rangle_{L^2(\rho_X)}\Bigr]
+\frac{1}{\lambda}\mathbb{E}_{\omega}\Bigl|\bigl\langle L_{K,\lambda}^{-1}K_{\bfx},\psi_\omega\bigr\rangle_{L^2(\rho_X)}\Bigr|^{2}.
\end{aligned}
\end{equation}
We evaluate the three terms on the right side of \eqref{eq:residexpand} separately. For the first term, the integral representation \eqref{in_re} gives
\begin{equation}\label{eq:term1}
\mathbb{E}_{\omega}\psi_\omega(\bfx)^{2}=K(\bfx,\bfx).
\end{equation}
For the second term, linearity of the inner product and
\eqref{in_re} give
\begin{equation}\label{eq:term2}
\mathbb{E}_{\omega}\Bigl[\psi_\omega(\bfx)
 \bigl\langle L_{K,\lambda}^{-1}K_{\bfx},\psi_\omega\bigr\rangle_{L^2(\rho_X)}\Bigr]
=\bigl\langle L_{K,\lambda}^{-1}K_{\bfx},
 \mathbb{E}_{\omega}[\psi_\omega(\bfx)\psi_\omega]\bigr\rangle_{L^2(\rho_X)}
=\bigl\langle L_{K,\lambda}^{-1}K_{\bfx},K_{\bfx}\bigr\rangle_{L^2(\rho_X)},
\end{equation}
For the third term, we also have
\begin{equation}\label{eq:term3}
\begin{aligned}
\mathbb{E}_{\omega}\Bigl|\bigl\langle L_{K,\lambda}^{-1}K_{\bfx},\psi_\omega\bigr\rangle_{L^2(\rho_X)}\Bigr|^{2}
&=\bigl\langle L_{K,\lambda}^{-1}K_{\bfx},
 \mathbb{E}_{\omega}\bigl[\psi_\omega\otimes\psi_\omega\bigr]
 L_{K,\lambda}^{-1}K_{\bfx}\bigr\rangle_{L^2(\rho_X)}\\
&=\bigl\langle L_{K,\lambda}^{-1}K_{\bfx},L_KL_{K,\lambda}^{-1}K_{\bfx}\bigr\rangle_{L^2(\rho_X)},
\end{aligned}
\end{equation}
Substituting \eqref{eq:term1}, \eqref{eq:term2},
and \eqref{eq:term3} into \eqref{eq:residexpand}, we obtain
\begin{equation}\label{eq:residsum}
\begin{aligned}
\frac{1}{\lambda}\mathbb{E}_{\omega}\Bigl|\psi_\omega(\bfx)
-\bigl\langle L_{K,\lambda}^{-1}K_{\bfx},\psi_\omega\bigr\rangle_{L^2(\rho_X)}\Bigr|^{2} &= \frac{1}{\lambda}K(\bfx,\bfx)
-\frac{2}{\lambda}\bigl\langle L_{K,\lambda}^{-1}K_{\bfx},K_{\bfx}\bigr\rangle_{L^2(\rho_X)}\\
&+\frac{1}{\lambda}\bigl\langle L_{K,\lambda}^{-1}K_{\bfx},LL_{K,\lambda}^{-1}K_{\bfx}\bigr\rangle_{L^2(\rho_X)}.
\end{aligned}
\end{equation}
Note that 
\begin{equation}\label{eq:complete}
\begin{aligned}
&\bigl\langle L_{K,\lambda}^{-1}K_{\bfx},L_KL_{K,\lambda}^{-1}K_{\bfx}\bigr\rangle_{L^2(\rho_X)}
+\lambda\bigl\|L_{K,\lambda}^{-1}K_{\bfx}\bigr\|_{L^2(\rho_X)}^{2}\\
=&\bigl\langle L_{K,\lambda}^{-1}K_{\bfx},L_KL_{K,\lambda}^{-1}K_{\bfx}\bigr\rangle_{L^2(\rho_X)}
+\lambda\bigl\langle L_{K,\lambda}^{-1}K_{\bfx},L_{K,\lambda}^{-1}K_{\bfx}\bigr\rangle_{L^2(\rho_X)}\\
=&\bigl\langle L_{K,\lambda}^{-1}K_{\bfx},(L_K+\lambda I)L_{K,\lambda}^{-1}K_{\bfx}\bigr\rangle_{L^2(\rho_X)}\\
=&\bigl\langle L_{K,\lambda}^{-1}K_{\bfx},L_{K,\lambda} L_{K,\lambda}^{-1}K_{\bfx}\bigr\rangle_{L^2(\rho_X)}
=\bigl\langle L_{K,\lambda}^{-1}K_{\bfx},K_{\bfx}\bigr\rangle_{L^2(\rho_X)}.
\end{aligned}
\end{equation}
Combining \eqref{eq:residsum} and \eqref{eq:complete}, we have 
\begin{equation*}
\bigl\|L_{K,\lambda}^{-1}K_{\bfx}\bigr\|_{L^2(\rho_X)}^{2}
+\frac{1}{\lambda}\mathbb{E}_{\omega}
\Bigl|\psi_\omega(\bfx)
-\bigl\langle L_{K,\lambda}^{-1}K_{\bfx},\psi_\omega\bigr\rangle_{L^2(\rho_X)}\Bigr|^{2}=\frac{K(\bfx,\bfx)
-\bigl\langle L_{K,\lambda}^{-1}K_{\bfx},K_{\bfx}\bigr\rangle_{L^2(\rho_X)}}{\lambda},
\end{equation*}
which proves the first equality in \eqref{eq:comparison}.

For the spectral equality, the Mercer expansion
$K_{\bfx}=\sum_j\mu_j e_j(\bfx)e_j$ converges in $L^2(\rho_X)$ for
$\rho_X$-almost every $\bfx$, and $L_{K,\lambda}^{-1}$ acts diagonally,
$L_{K,\lambda}^{-1}e_j=(\mu_j+\lambda)^{-1}e_j$. Hence, by orthonormality
of $\{e_j\}_{j\ge 1}$,
\[
\bigl\langle L_{K,\lambda}^{-1}K_{\bfx},K_{\bfx}\bigr\rangle_{L^2(\rho_X)}
=\sum_j\frac{\mu_j^2}{\mu_j+\lambda}e_j(\bfx)^2,
\qquad
K(\bfx,\bfx)=\sum_j\mu_je_j(\bfx)^2,
\]
and subtracting the two gives
$\lambda^{-1}\sum_j\bigl(\mu_j-\mu_j^2/(\mu_j+\lambda)\bigr)e_j(\bfx)^2
=\sum_j\frac{\mu_j}{\mu_j+\lambda}e_j(\bfx)^2$. The last inequality follows from Lemma \ref{lem:embed}. Thus we complete the proof.

\end{proof}

Since both terms on the left side of \eqref{eq:comparison} are nonnegative. Hence under Assumption \ref{ass5}, we have
\begin{equation}\label{eq:compbounds}
\bigl\|L_{K,\lambda}^{-1}K_{\bfx}\bigr\|_{L^2(\rho_X)}^{2}\le M_{\alpha}\lambda^{-\alpha},
\qquad
\mathbb{E}_{\omega}\Bigl|\psi_\omega(\bfx)
-\bigl\langle L_{K,\lambda}^{-1}K_{\bfx},\psi_\omega\bigr\rangle_{L^2(\rho_X)}\Bigr|^{2}
\le M_{\alpha}\lambda^{1-\alpha}.
\end{equation}
Under Assumption \ref{ass1} that $\|\psi_\omega\|_{L^2(\rho_X)}\le\kappa$, the triangle inequality gives
\begin{equation}\label{eq:envelope}
\Bigl|\psi_\omega(\bfx)
-\bigl\langle L_{K,\lambda}^{-1}K_{\bfx},\psi_\omega\bigr\rangle_{L^2(\rho_X)}\Bigr|^{2}
\le\kappa^{2}\bigl(1+\sqrt{M_{\alpha}\lambda^{-\alpha}}\bigr)^{2}\le4\kappa^{2}M_{\alpha}\lambda^{-\alpha},
\end{equation}
where the last inequality follows from the fact that $1+\sqrt{M_{\alpha}\lambda^{-\alpha}}\le 2\sqrt{M_{\alpha}\lambda^{-\alpha}}$ for $M_{\alpha}\lambda^{-\alpha}\ge 1$ with $M_\alpha\ge 1$ and $0<\lambda\le 1$.

Note that \eqref{eq:envelope} already gives a uniform bound for the
scalar residual under uniformly sampled features.
With the compatibility condition of Assumption~\ref{ass7}, we now
refine this bound.

\begin{lemma}\label{lem:residual}
Under Assumptions \ref{ass1}, \ref{ass5} and
\ref{ass7}, let $0<\lambda\le\min\{1,\|L_K\|\}$, then for any $\alpha_0 < \alpha \le 1$, a fixed random feature realization $\omega_1,\dots,\omega_M$ and
$\rho_X$-almost every $\bfx\in\mathcal X$, 
\begin{equation}\label{eq:residualbound}
\bigl|\psi_\omega(\bfx)
-\langle L_{K,\lambda}^{-1}K_{\bfx},\psi_\omega\rangle_{L^2(\rho_X)}\bigr|^2
\le \begin{cases}
CM_\alpha\lambda^{1-\zeta-\alpha}, & \zeta+\alpha\neq 1,\\[6pt]
\displaystyle
CM_\alpha\left[1+\log\frac{1+\|L_K\|}{\lambda}\right], & \zeta+\alpha=1.
\end{cases} 
\end{equation}
\end{lemma}

\begin{proof}
For a fixed $\omega$, we expand
$\psi_\omega=\sum_j a_je_j$ in $L^2(\rho_X)$, where
$(\mu_j,e_j)$ are the nonzero eigenpairs of $L_K$ and
$a_j=\langle\psi_\omega,e_j\rangle_{L^2(\rho_X)}$. By self-adjointness of $L_K$ and
$I-L_KL_{K,\lambda}^{-1}=\lambda L_{K,\lambda}^{-1}$,
\begin{equation}\label{eq:residualpointwise}
\begin{aligned}
\psi_\omega(\bfx)-\langle L_{K,\lambda}^{-1}K_{\bfx},\psi_\omega\rangle_{L^2(\rho_X)}&=\psi_\omega(\bfx)-\langle K_{\bfx},L_{K,\lambda}^{-1}\psi_\omega\rangle_{L^2(\rho_X)}
\\
&=\psi_\omega(\bfx)-\bigl(L_KL_{K,\lambda}^{-1}\psi_\omega\bigr)(\bfx)=\bigl(\lambda L_{K,\lambda}^{-1}\psi_\omega\bigr)(\bfx),
\end{aligned}
\end{equation}
so the scalar we must bound is the pointwise value of the $L^2(\rho_X)$
function $\lambda L_{K,\lambda}^{-1}\psi_\omega$.

We first control the coefficients $a_j$.
For $0<t\le1$, Assumption~\ref{ass7} gives
$$
\sum_{\mu_j\le t}|a_j|^2
\le2t\sum_j\frac{|a_j|^2}{\mu_j+t}
=2t\bigl\|L_{K,t}^{-1/2}\psi_\omega\bigr\|_{L^2(\rho_X)}^2
\le2Ft^{1-\zeta}.
$$
For $t>1$, Assumption~\ref{ass1} gives
$$
\sum_{\mu_j\le t}|a_j|^2\le \sum_{j}|a_j|^2 = \|\psi_\omega\|_{L^2(\rho_X)}^2 \le\kappa^2\le\kappa^2t^{1-\zeta},
$$
where the last inequality uses $\zeta\le1$. Combining the two cases $0<t\le1$ and $t>1$ gives the following bound:
\begin{equation}\label{eq:shellbound}
\sum_{\mu_j\le t}|a_j|^2 \le(2F+\kappa^2)t^{1-\zeta}.
\end{equation}

We split into three cases according to the sign of $1-\zeta-\alpha$.

\noindent \textbf{Case 1: $\zeta+\alpha<1$.}
By \eqref{eq:residualpointwise} and the spectral decomposition in $L^2(\rho_X)$, we have
\begin{equation}\label{eq:residualspectral}
\bigl\|L_K^{-\alpha/2}
\bigl(\psi_\omega-L_KL_{K,\lambda}^{-1}\psi_\omega\bigr)\bigr\|_{L^2(\rho_X)}^2
=\sum_j\frac{\lambda^2\mu_j^{-\alpha}}{(\mu_j+\lambda)^2}|a_j|^2.
\end{equation}
We bound the sum in two parts. For $\mu_j\le\lambda$, we write $(0,\lambda]$ as the disjoint union of the intervals $(2^{-k-1}\lambda,2^{-k}\lambda]$, $k\ge0$, and
bound each part separately. Take $t=2^{-k}\lambda$ in \eqref{eq:shellbound}.
Since $\zeta+\alpha<1$, the sum over $k\ge0$ is
\begin{equation}\label{eq:shellbound1}
\begin{aligned}
\sum_{\mu_j\le\lambda}
\frac{\lambda^2\mu_j^{-\alpha}}{(\mu_j+\lambda)^2}|a_j|^2
&=\sum_{k\ge0}
\sum_{2^{-k-1}\lambda<\mu_j\le2^{-k}\lambda}
\frac{\lambda^2\mu_j^{-\alpha}}{(\mu_j+\lambda)^2}|a_j|^2\\
&\le\sum_{k\ge0}(2^{-k-1}\lambda)^{-\alpha}
\sum_{2^{-k-1}\lambda<\mu_j\le2^{-k}\lambda}|a_j|^2\\
&\le (2F+\kappa^2)\sum_{k\ge0}
(2^{-k-1}\lambda)^{-\alpha}(2^{-k}\lambda)^{1-\zeta}\\
&=(2F+\kappa^2)\lambda^{1-\zeta-\alpha}
2^{\alpha}\sum_{k\ge0}2^{-k(1-\zeta-\alpha)}\\
&\le \frac{2^{\alpha}(2F+\kappa^2)}{1-2^{-(1-\zeta-\alpha)}} \lambda^{1-\zeta-\alpha}.
\end{aligned}
\end{equation}
Here the first inequality uses
$\lambda^2\mu_j^{-\alpha}/(\mu_j+\lambda)^2\le\mu_j^{-\alpha}$ and $\mu_j>2^{-k-1}\lambda$ for the shell, the second inequality uses \eqref{eq:shellbound} with
$t=2^{-k}\lambda$, and the last inequality uses $\sum_{k\ge0}2^{-k(1-\zeta-\alpha)}$ converges to $1/(1-2^{-(1-\zeta-\alpha)})$ since $1-\zeta-\alpha>0$.

For $\mu_j>\lambda$, we split the sum over the scales
$2^k\lambda<\mu_j\le2^{k+1}\lambda$, $k\ge0$.
Since $\mu_j\le\|L_K\|$, only finitely many scales are nonempty,
and the empty scales contribute zero.
Take $t=2^{k+1}\lambda$ in \eqref{eq:shellbound},
the sum over $k\ge0$ is
\begin{equation}\label{eq:shellbound2}
\begin{aligned}
\sum_{\mu_j>\lambda}
\frac{\lambda^2\mu_j^{-\alpha}}{(\mu_j+\lambda)^2}|a_j|^2
&=\sum_{k\ge0}
\sum_{2^k\lambda<\mu_j\le2^{k+1}\lambda}
\frac{\lambda^2\mu_j^{-\alpha}}{(\mu_j+\lambda)^2}|a_j|^2\\
&\le\sum_{k\ge0}\lambda^2(2^k\lambda)^{-2-\alpha}
\sum_{2^k\lambda<\mu_j\le2^{k+1}\lambda}|a_j|^2\\
&\le(2F+\kappa^2)\sum_{k\ge0}
\lambda^2(2^k\lambda)^{-2-\alpha}(2^{k+1}\lambda)^{1-\zeta}\\
&=(2F+\kappa^2)\lambda^{1-\zeta-\alpha}
2^{1-\zeta}\sum_{k\ge0}2^{-k(1+\zeta+\alpha)}\\
&\le\frac{2^{1-\zeta}(2F+\kappa^2)}{1-2^{-(1+\zeta+\alpha)}}
\lambda^{1-\zeta-\alpha}.
\end{aligned}
\end{equation}
Here the first inequality uses
$\lambda^2\mu_j^{-\alpha}/(\mu_j+\lambda)^2
\le\lambda^2\mu_j^{-2-\alpha}$
because $\mu_j+\lambda>\mu_j$, and $\mu_j>2^k\lambda$ on each scale,
the second inequality uses \eqref{eq:shellbound} with
$t=2^{k+1}\lambda$,
and the last inequality uses
$\sum_{k\ge0}2^{-k(1+\zeta+\alpha)}=1/(1-2^{-(1+\zeta+\alpha)})$,
which converges for every $\zeta,\alpha>0$.

Combining \eqref{eq:shellbound}, \eqref{eq:shellbound1}, and \eqref{eq:shellbound2}, we have
\[
\bigl\|L_K^{-\alpha/2}
\bigl(\psi_\omega-L_KL_{K,\lambda}^{-1}\psi_\omega\bigr)\bigr\|_{L^2(\rho_X)}^2
\le C \lambda^{1-\zeta-\alpha},
\]
where $C$ is a constant depending on $F$, $\kappa$, $\zeta$, and $\alpha$.

Applying the embedding condition of Assumption~\ref{ass5},
\[
\bigl|\psi_\omega(\bfx)-\langle L_{K,\lambda}^{-1}K_{\bfx},\psi_\omega\rangle_{L^2(\rho_X)}\bigr|^2
\le M_\alpha
\bigl\|L_K^{-\alpha/2}
\bigl(\psi_\omega-L_KL_{K,\lambda}^{-1}\psi_\omega\bigr)\bigr\|_{L^2(\rho_X)}^2
\le C M_\alpha\lambda^{1-\zeta-\alpha}.
\]

\noindent\textbf{Case 2: $\zeta+\alpha>1$.} By \eqref{eq:residualpointwise} and the triangle inequality,
\begin{equation}\label{eq:residualpointwisecase2}
\bigl|\psi_\omega(\bfx)-\langle L_{K,\lambda}^{-1}K_{\bfx},\psi_\omega\rangle_{L^2(\rho_X)}\bigr|^2
=\bigl|\psi_\omega(\bfx)-L_KL_{K,\lambda}^{-1}\psi_\omega(\bfx)\bigr|^2
\le2\kappa^2+2\bigl|L_KL_{K,\lambda}^{-1}\psi_\omega(\bfx)\bigr|^2,
\end{equation}
where the inequality uses $|\psi_\omega(\bfx)|\le\kappa$ from
Assumption~\ref{ass1}. It remains to bound the second part pointwise, and its spectral decomposition is
\begin{equation}\label{eq:regspectral}
\bigl\|L_K^{-\alpha/2}L_KL_{K,\lambda}^{-1}\psi_\omega\bigr\|_{L^2(\rho_X)}^2
=\sum_j\frac{\mu_j^{2-\alpha}}{(\mu_j+\lambda)^2}|a_j|^2.
\end{equation}

For $\mu_j\le\lambda$, the multiplier satisfies
$
\frac{\mu_j^{2-\alpha}}{(\mu_j+\lambda)^2}
\le\frac{\mu_j^{2-\alpha}}{\lambda^2}
\le\lambda^{-\alpha}.
$
By \eqref{eq:shellbound} with $t=\lambda$,
\begin{equation}\label{eq:regspectral1}
\sum_{\mu_j\le\lambda}
\frac{\mu_j^{2-\alpha}}{(\mu_j+\lambda)^2}|a_j|^2
\le\lambda^{-\alpha}\sum_{\mu_j\le\lambda}|a_j|^2
\le(2F+\kappa^2)\lambda^{1-\zeta-\alpha}.
\end{equation}

For $\mu_j>\lambda$, similarly with Case 1, we split the sum over the scales
$2^k\lambda<\mu_j\le2^{k+1}\lambda$, $k\ge0$, and take $t=2^{k+1}\lambda$ in \eqref{eq:shellbound}. The sum over $k\ge0$ is
\begin{equation}\label{eq:regspectral2}
\begin{aligned}
\sum_{\mu_j>\lambda}
\frac{\mu_j^{2-\alpha}}{(\mu_j+\lambda)^2}|a_j|^2
&=\sum_{k\ge0}
\sum_{2^k\lambda<\mu_j\le2^{k+1}\lambda}
\frac{\mu_j^{2-\alpha}}{(\mu_j+\lambda)^2}|a_j|^2\\
&\le\sum_{k\ge0}(2^k\lambda)^{-\alpha}
\sum_{2^k\lambda<\mu_j\le2^{k+1}\lambda}|a_j|^2\\
&\le(2F+\kappa^2)\sum_{k\ge0}
(2^k\lambda)^{-\alpha}(2^{k+1}\lambda)^{1-\zeta}\\
&=(2F+\kappa^2)\lambda^{1-\zeta-\alpha}
2^{1-\zeta}\sum_{k\ge0}2^{-k(\zeta+\alpha-1)}\\
&=\frac{2^{1-\zeta}(2F+\kappa^2)}{1-2^{-(\zeta+\alpha-1)}}
\lambda^{1-\zeta-\alpha}.
\end{aligned}
\end{equation}
Here the first inequality uses $\frac{\mu_j^{2-\alpha}}{(\mu_j+\lambda)^2}
\le\frac{\mu_j^{2-\alpha}}{\mu_j^2}
=\mu_j^{-\alpha}
\le(2^k\lambda)^{-\alpha}$, because $\mu_j+\lambda>\mu_j$ and $\mu_j>2^k\lambda$ on each scale, the second inequality uses \eqref{eq:shellbound} with
$t=2^{k+1}\lambda$, 
and the last equality uses
$\sum_{k\ge0}2^{-k(\zeta+\alpha-1)}=1/(1-2^{-(\zeta+\alpha-1)})$,
which converges because $\zeta+\alpha-1>0$.

Combining \eqref{eq:residualpointwisecase2}-\eqref{eq:regspectral2}, and the embedding condition of Assumption~\ref{ass5}, we have 
$$
\bigl|\psi_\omega(\bfx)-\langle L_{K,\lambda}^{-1}K_{\bfx},\psi_\omega\rangle_{L^2(\rho_X)}\bigr|^2 \le 2\kappa^2+C M_\alpha\lambda^{1-\zeta-\alpha}.
$$
Since $1-\zeta-\alpha<0$ and $\lambda\le1$, we have
$\lambda^{1-\zeta-\alpha}\ge1$, so the constant $2\kappa^2$ can be absorbed into \(CM_\alpha\lambda^{1-\zeta-\alpha}\) after enlarging \(C\). 
This proves $\bigl|\psi_\omega(\bfx)-\langle L_{K,\lambda}^{-1}K_{\bfx},\psi_\omega\rangle_{L^2(\rho_X)}\bigr|^2 \le C M_\alpha\lambda^{1-\zeta-\alpha}$.

\noindent\textbf{Case 3: $\zeta+\alpha=1$.} Repeat the argument of Case 2 with $\zeta+\alpha=1$.
For $\mu_j\le\lambda$, the same bound as in Case 2 gives
\begin{equation}\label{eq:regspectral_0}
\begin{aligned}
\sum_{\mu_j\le\lambda}
\frac{\mu_j^{2-\alpha}}{(\mu_j+\lambda)^2}|a_j|^2
\le(2F+\kappa^2)\lambda^{1-\zeta-\alpha}
=2F+\kappa^2.
\end{aligned}
\end{equation}

For $\mu_j>\lambda$, similarly with Case 1, we split the sum over the scales
$2^k\lambda<\mu_j\le2^{k+1}\lambda$, $k\ge0$, and take $t=2^{k+1}\lambda$ in \eqref{eq:shellbound}. Since $\mu_j\le\|L_K\|$, the number of nonempty scales is at most
$1+\lfloor\log_2(\|L_K\|/\lambda)\rfloor$, so we have 
\begin{equation}\label{eq:regspectral_1}
\begin{aligned}
\sum_{\mu_j>\lambda}
\frac{\mu_j^{2-\alpha}}{(\mu_j+\lambda)^2}|a_j|^2
&=\sum_{k=0}^{\lfloor\log_2(\|L_K\|/\lambda)\rfloor}
\sum_{2^k\lambda<\mu_j\le2^{k+1}\lambda}
\frac{\mu_j^{2-\alpha}}{(\mu_j+\lambda)^2}|a_j|^2\\
&\le\sum_{k=0}^{\lfloor\log_2(\|L_K\|/\lambda)\rfloor}(2^k\lambda)^{-\alpha}
\sum_{2^k\lambda<\mu_j\le2^{k+1}\lambda}|a_j|^2\\
&\le(2F+\kappa^2)\sum_{k=0}^{\lfloor\log_2(\|L_K\|/\lambda)\rfloor}
(2^k\lambda)^{-\alpha}(2^{k+1}\lambda)^{1-\zeta}\\
&=(2F+\kappa^2)
2^{1-\zeta}\Bigl[1+\lfloor\log_2(\|L_K\|/\lambda)\rfloor\Bigr],
\end{aligned}
\end{equation}
where the last equality uses $\zeta+\alpha=1$.

Similarly with Case 2, combining \eqref{eq:residualpointwisecase2}, \eqref{eq:regspectral}, \eqref{eq:regspectral_0}, \eqref{eq:regspectral_1}, and the embedding condition of Assumption~\ref{ass5}, we have 
$$
\bigl|\psi_\omega(\bfx)-\langle L_{K,\lambda}^{-1}K_{\bfx},\psi_\omega\rangle_{L^2(\rho_X)}\bigr|^2 \le CM_\alpha\left[1+\log\frac{1+\|L_K\|}{\lambda}\right].
$$

Combining the results of all three cases completes the proof.

\end{proof}

\begin{lemma}\label{lem:pointwise}
Under Assumptions \ref{ass1}, \ref{ass5} and \ref{ass7}, let $0<\lambda\le\min\{1,\|L_K\|\}$, then for any $\alpha_0 < \alpha \le 1$, every $t\ge4$ and $\rho_X$-almost every $\bfx\in\mathcal X$,
\begin{equation}\label{eq:pointwise}
\mathbb{P}_{\Omega}\Bigl\{
\bigl\|C_{M,\lambda}^{-1/2}\phi_M(\bfx)\bigr\|_2^{2}
>tM_{\alpha}\lambda^{-\alpha}\Bigr\}
\le
\begin{cases}
\exp(-cM\lambda^\zeta t), & \zeta+\alpha\ne 1,\\[6pt]
\displaystyle
\exp\Bigl(-\frac{cM\lambda^\zeta t}
{1+\log((1+\|L_K\|)/\lambda)}\Bigr), & \zeta+\alpha=1,
\end{cases}
\end{equation}
\end{lemma}

\begin{proof}
For $\rho_X$-almost every $\bfx\in\mathcal X$, we define i.i.d. random variables
\[
U_i=\Bigl|\psi_{\omega_i}(\bfx)
-\bigl\langle L_{K,\lambda}^{-1}K_{\bfx},\psi_{\omega_i}\bigr\rangle_{L^{2}(\rho_X)}\Bigr|^{2},
\quad i=1,\dots,M.
\]
and $b_\lambda = \begin{cases}
CM_\alpha\lambda^{1-\zeta-\alpha}, & \zeta+\alpha\neq 1,\\
CM_\alpha\left[1+\log\frac{1+\|L_K\|}{\lambda}\right], & \zeta+\alpha=1.
\end{cases}$ From Lemma \ref{lem:residual},  \eqref{eq:compbounds} and \eqref{eq:envelope}, we have 
\begin{equation}\label{eq:pointmoments}
0\le U_i\le b_\lambda,
\quad
\mathbb{E}U_i\le M_{\alpha}\lambda^{1-\alpha},
\quad
\operatorname{Var}(U_i)\le b_\lambda M_{\alpha}\lambda^{1-\alpha},
\end{equation}
where the last inequality follows from the fact that $\operatorname{Var}(U_i)\le\mathbb{E}U_i^{2}\le \sup U_i \cdot\mathbb{E}U_i$.

By Lemma~\ref{lem:varlev}, the definition of the infimum gives, for the
particular choice $h=L_{K,\lambda}^{-1}K_{\bfx}$,
\begin{equation}\label{eq:pointvarbound}
\bigl\|C_{M,\lambda}^{-1/2}\phi_M(\bfx)\bigr\|_2^{2}
\le\bigl\|L_{K,\lambda}^{-1}K_{\bfx}\bigr\|_{L^{2}(\rho_X)}^{2}
+\frac{1}{\lambda}\cdot\frac{1}{M}\sum_{i=1}^{M}U_i,
\end{equation}
where
$\frac{1}{M}\sum_{i=1}^{M}U_i
=\frac{1}{M}\sum_{i=1}^{M}
\bigl|\psi_{\omega_i}(\bfx)
-\bigl\langle L_{K,\lambda}^{-1}K_{\bfx},\psi_{\omega_i}\bigr\rangle_{L^{2}(\rho_X)}\bigr|^{2}$
is exactly the second term of the variational objective evaluated at this
$h$. On the event
$\bigl\{\bigl\|C_{M,\lambda}^{-1/2}\phi_M(\bfx)\bigr\|_2^{2}
>tM_{\alpha}\lambda^{-\alpha}\bigr\}$,
rearranging \eqref{eq:pointvarbound} yields
$$
\frac{1}{M}\sum_{i=1}^{M}U_i
>\lambda\left(tM_{\alpha}\lambda^{-\alpha}
-\bigl\|L_{K,\lambda}^{-1}K_{\bfx}\bigr\|_{L^{2}(\rho_X)}^{2}\right)>(t-1)M_{\alpha}\lambda^{1-\alpha}.
$$
where the second inequality uses $\bigl\|L_{K,\lambda}^{-1}K_{\bfx}\bigr\|_{L^{2}(\rho_X)}^{2} \le M_{\alpha}\lambda^{-\alpha}$ from \eqref{eq:compbounds}. Note that $\mathbb{E}U_i\le M_{\alpha}\lambda^{1-\alpha}$ from \eqref{eq:pointmoments}, we have 
\begin{equation}\label{eq:pointdev}
  \frac{1}{M}\sum_{i=1}^{M}(U_i-\mathbb{E}U_i)
>(t-1)M_{\alpha}\lambda^{1-\alpha}-M_{\alpha}\lambda^{1-\alpha}
=(t-2)M_{\alpha}\lambda^{1-\alpha}.
\end{equation}

Now we apply the scalar Bernstein inequality in Lemma \ref{lem:sbern}
with $u=(t-2)M_{\alpha}\lambda^{1-\alpha}$, $b=b_\lambda$, and
$v=b_\lambda M_{\alpha}\lambda^{1-\alpha}$,
\begin{align*}
&\mathbb{P}_{\Omega}\left\{\frac{1}{M}\sum_{i=1}^{M}(U_i-\mathbb{E}U_i) > (t-2)M_{\alpha}\lambda^{1-\alpha}\right\}
\le \exp\left(-\frac{3M(t-2)^2M_{\alpha}\lambda^{1-\alpha}}{2b_\lambda(t+1)}\right)\\
& \qquad \le \exp\left(-\frac{M\lambda^{1-\alpha}M_{\alpha}t}{8b_\lambda}\right) \le \begin{cases}
\exp\left(-cM\lambda^\zeta t\right), & \zeta+\alpha\ne1,\\[6pt]
\displaystyle \exp\left(-\frac{cM\lambda^\zeta t}{1+\log((1+\|L_K\|)/\lambda)}\right),
& \zeta+\alpha=1,
\end{cases}
\end{align*}
where the first inequality holds for $t\ge4$ since
$(t-2)^{2}/(t+1)\ge t/12$, the last inequality uses $c=1/(8C)$ with $C$ being the constant in the definition of $b_\lambda$. Since the event $\bigl\{\bigl\|C_{M,\lambda}^{-1/2}\phi_M(\bfx)\bigr\|_2^{2} > tM_{\alpha}\lambda^{-\alpha}\bigr\}$ implies the event $\left\{\frac{1}{M}\sum_{i=1}^{M}(U_i-\mathbb{E}U_i) > (t-2)M_{\alpha}\lambda^{1-\alpha}\right\}$, we have
\[
\mathbb{P}_{\Omega}\left\{\bigl\|C_{M,\lambda}^{-1/2}\phi_M(\bfx)\bigr\|_2^{2} > tM_{\alpha}\lambda^{-\alpha}\right\} \le \begin{cases}
\exp\left(-cM\lambda^\zeta t\right), & \zeta+\alpha\ne1,\\[6pt]
\displaystyle \exp\left(-\frac{cM\lambda^\zeta t}{1+\log((1+\|L_K\|)/\lambda)}\right),
& \zeta+\alpha=1.
\end{cases}
\]
Thus we complete the proof.
\end{proof}

For a fixed random feature realization $\omega_1,\dots,\omega_M$, define two sets 
\begin{equation}\label{eq:omegas}
\mathcal{A}_1=\bigl\{\bfx:\bigl\|C_{M,\lambda}^{-1/2}\phi_M(\bfx)\bigr\|_2^{2}
\le8M_{\alpha}\lambda^{-\alpha}\bigr\},
\qquad
\mathcal{A}_2=\mathcal X\setminus\mathcal{A}_1 .
\end{equation}

\begin{lemma}\label{lem:exceptional}
Under Assumptions \ref{ass1}, \ref{ass5} and \ref{ass7}, let $0<\lambda\le\min\{1,\|L_K\|\}$ and $0<\eta<1/4$,  for any $\alpha_0 < \alpha \le 1$, if the number of random features $M$ satisfies
\begin{equation}\label{eq:exceptionalbudget}
M\ge C_0\lambda^{-\zeta}
\begin{cases}
\displaystyle\log\frac{c(n+M_\alpha\lambda^{-\alpha})}{\eta^2},
& \zeta+\alpha\ne1,\\[10pt]
\displaystyle
\Bigl[1+\log\frac{1+\|L_K\|}{\lambda}\Bigr]
\log\frac{c(n+M_\alpha\lambda^{-\alpha})}{\eta^2},
& \zeta+\alpha=1, 
\end{cases}
\end{equation}
then with probability at least $1-2\eta$ over the random features,
\begin{equation}\label{eq:exceptional}
\rho_X(\mathcal{A}_2)\le\frac{\eta}{n},
\qquad
\int_{\mathcal{A}_2}\bigl\|C_{M,\lambda}^{-1/2}\phi_M(\bfx)\bigr\|_2^{2}
d\rho_X(\bfx)\le\frac18 .
\end{equation}
\end{lemma}

\begin{proof}
First, the map
$(\omega,\bfx)\mapsto\bigl\|C_{M,\lambda}^{-1/2}\phi_M(\bfx)\bigr\|_2^{2}$
is jointly measurable, since each coordinate $\psi_{\omega_i}(\bfx)$ is
measurable in $(\omega,\bfx)$ by assumption, and
$C_{M,\lambda}^{-1/2}$ depends measurably on $\omega$ only. Hence
Tonelli's theorem applies to every nonnegative function of this map.

For the mass of $\mathcal{A}_2$, Tonelli's theorem and \eqref{eq:pointwise}
with $t=8$  give
\begin{equation}\label{eq:expmass}
\begin{aligned}
\mathbb{E}_{\Omega}\bigl[\rho_X(\mathcal{A}_2)\bigr]&=\mathbb{E}_{\Omega}\left[\int_{\mathcal X}
 \mathbf{1}\bigl\{\bigl\|C_{M,\lambda}^{-1/2}\phi_M(\bfx)\bigr\|_2^{2}>8M_{\alpha}\lambda^{-\alpha}\bigr\}d\rho_X(\bfx)\right]\\
&=\int_{\mathcal X}\mathbb{P}_{\Omega}\Bigl\{
 \bigl\|C_{M,\lambda}^{-1/2}\phi_M(\bfx)\bigr\|_2^{2}
 >8M_{\alpha}\lambda^{-\alpha}\Bigr\}d\rho_X(\bfx) \\
&\le\int_{\mathcal X}
 \exp\left(-\frac{M\lambda^{1-\alpha}M_{\alpha}}{b_\lambda}\right)d\rho_X(\bfx)
=\exp\left(-\frac{M\lambda^{1-\alpha}M_{\alpha}}{b_\lambda}\right),
\end{aligned}
\end{equation}
where $b_\lambda$ is defined in Lemma \ref{lem:pointwise}.

Recall the tail identity
$\mathbb{E}\bigl[W\mathbf{1}\{W>u\}\bigr]
=u\mathbb{P}(W>u)+\int_u^{\infty}\mathbb{P}(W>v)dv$,
applied to $W=\bigl\|C_{M,\lambda}^{-1/2}\phi_M(\bfx)\bigr\|_2^{2}$
with $u=8M_{\alpha}\lambda^{-\alpha}$, and integrated against $\rho_X$, together with Tonelli's theorem gives
\begin{equation}\label{eq:explev1}
\begin{aligned}
\mathbb{E}_{\Omega}\left[\int_{\mathcal{A}_2}
 \bigl\|C_{M,\lambda}^{-1/2}\phi_M(\bfx)\bigr\|_2^{2}d\rho_X(\bfx)\right] &=\int_{\mathcal X}\left(
 8M_{\alpha}\lambda^{-\alpha}
 \mathbb{P}_{\Omega}\Bigl\{\bigl\|C_{M,\lambda}^{-1/2}\phi_M(\bfx)\bigr\|_2^{2}
 >8M_{\alpha}\lambda^{-\alpha}\Bigr\}\right.\\
&\left.
 +\int_{8M_{\alpha}\lambda^{-\alpha}}^{\infty}
 \mathbb{P}_{\Omega}\Bigl\{
 \bigl\|C_{M,\lambda}^{-1/2}\phi_M(\bfx)\bigr\|_2^{2}>v\Bigr\}dv\right)
 d\rho_X(\bfx). 
\end{aligned}
\end{equation}
For the inner probabilities, fix $v\ge8M_{\alpha}\lambda^{-\alpha}$
and apply \eqref{eq:pointwise} at threshold
$t=v/(M_{\alpha}\lambda^{-\alpha})\ge8\ge4$:
\begin{equation}\label{eq:tailv}
\mathbb{P}_{\Omega}\Bigl\{
\bigl\|C_{M,\lambda}^{-1/2}\phi_M(\bfx)\bigr\|_2^{2}>v\Bigr\}
\le\exp\left(
-\frac{M\lambda v}{8b_{\lambda}}\right).
\end{equation}
Substituting \eqref{eq:tailv} and the bound from \eqref{eq:expmass}
into \eqref{eq:explev1},
\begin{equation}\label{eq:explev}
\begin{aligned}
&\mathbb{E}_{\Omega}\left[\int_{\mathcal{A}_2}
 \bigl\|C_{M,\lambda}^{-1/2}\phi_M(\bfx)\bigr\|_2^{2}d\rho_X(\bfx)\right]\\
\le &8M_{\alpha}\lambda^{-\alpha}
 \exp\left(-\frac{M\lambda^{1-\alpha}M_{\alpha}}{b_\lambda}\right)
 +\int_{8M_{\alpha}\lambda^{-\alpha}}^{\infty}
 \exp\left(-\frac{M\lambda v}{8b_{\lambda}}\right)dv
\\
=&8M_{\alpha}\lambda^{-\alpha}
 \exp\left(-\frac{M\lambda^{1-\alpha}M_{\alpha}}{b_\lambda}\right)
 +\frac{8b_{\lambda}}{M\lambda}
 \exp\left(-\frac{M\lambda^{1-\alpha}M_{\alpha}}{b_\lambda}\right)
\\
\le &9M_{\alpha}\lambda^{-\alpha}
 \exp\left(-\frac{M\lambda^{1-\alpha}M_{\alpha}}{b_\lambda}\right),
\end{aligned}
\end{equation}
where the last inequality uses the definition of $b_\lambda$ which gives
$b_\lambda\le C M_\alpha\lambda^{1-\zeta-\alpha}
[1+\log((1+\|L_K\|)/\lambda)]$ in both cases, so that
$8b_\lambda/(M\lambda)\le M_\alpha\lambda^{-\alpha}$ whenever
\begin{equation}\label{eq:budgetmid}
M\ge 8C\lambda^{-\zeta}\Bigl[1+\log\frac{1+\|L_K\|}{\lambda}\Bigr].
\end{equation}

Now we define two events
\[
G_1:=\Bigl\{\rho_X(\mathcal{A}_2)>\frac{\eta}{n}\Bigr\},
\qquad
G_2:=\Bigl\{\int_{\mathcal{A}_2}
\bigl\|C_{M,\lambda}^{-1/2}\phi_M(\bfx)\bigr\|^{2}d\rho_X(\bfx)
>\frac18\Bigr\}.
\]
and apply Markov's inequality to \eqref{eq:expmass} and
\eqref{eq:explev},
$$
\mathbb{P}_{\Omega}(G_1)
\le\frac{n}{\eta}
\exp\left(-\frac{M\lambda^{1-\alpha}M_{\alpha}}{b_\lambda}\right),
\qquad
\mathbb{P}_{\Omega}(G_2)
\le72M_{\alpha}\lambda^{-\alpha}
\exp\left(-\frac{M\lambda^{1-\alpha}M_{\alpha}}{b_\lambda}\right).
$$
Each of the above bounds is at most $\eta$ if
$$
\frac{M\lambda^{1-\alpha}M_\alpha}{b_\lambda}
\ge\max\left\{
\log\frac{n}{\eta^{2}},
\log\frac{72M_{\alpha}\lambda^{-\alpha}}{\eta}\right\}.
$$
Substituting the definition of $b_\lambda$, this is implied by
\begin{equation}
M\ge C_0\lambda^{-\zeta}
\begin{cases}
\displaystyle\log\frac{c(n+M_\alpha\lambda^{-\alpha})}{\eta^2},
& \zeta+\alpha\ne1,\\[10pt]
\displaystyle
\Bigl[1+\log\frac{1+\|L_K\|}{\lambda}\Bigr]
\log\frac{c(n+M_\alpha\lambda^{-\alpha})}{\eta^2},
& \zeta+\alpha=1, \notag
\end{cases}
\end{equation}
with $C_0$ large enough to cover \eqref{eq:budgetmid} and $c\ge72$. This is possible because $\log\frac{c(n+M_\alpha\lambda^{-\alpha})}{\eta^2}
\ge\alpha\log(1/\lambda)+\log(cM_\alpha/\eta^2)$,
so in the case $\zeta+\alpha\ne1$ the condition
$C_0\ge 8C/\alpha$ gives
$C_0\log\frac{c(n+M_\alpha\lambda^{-\alpha})}{\eta^2} \ge 8C[1+\log((1+\|L_K\|)/\lambda)]$
for all $0<\lambda\le1$, provided $c$ is chosen with
$\log(cM_\alpha/\eta^2)\ge 1+\log(1+\|L_K\|)$.
On the complement of $G_1\cup G_2$, whose probability is at least
$1-\mathbb{P}_{\Omega}(G_1)-\mathbb{P}_{\Omega}(G_2)\ge1-2\eta$, both
statements of \eqref{eq:exceptional} hold. Thus we complete the proof.

\end{proof}

\begin{lemma}\label{lem:covariance}
Under Assumptions \ref{ass1}, \ref{ass5} and \ref{ass7}, let $0<\lambda\le\min\{1,\|L_K\|\}$ and $0<\eta<1/4$.  For any $\alpha_0 < \alpha \le 1$, suppose that the number of random features $M$ satisfies \eqref{eq:exceptionalbudget}, and the sample size $n$ satisfies
\begin{equation}\label{eq:covbudget}
n\ge\frac{832}{3}M_{\alpha}\lambda^{-\alpha}\log\frac{2M}{\eta},
\end{equation}
then with probability at least $1-4\eta$ over the random features and the
samples,
\begin{equation}\label{eq:covwhitened}
\bigl\|\widehat{C}_{M,\lambda}^{-1/2}C_{M,\lambda}^{1/2}\bigr\|
\le \sqrt{\frac85},
\end{equation}
and consequently
\begin{equation}\label{eq:covfactor}
\bigl\|S_M\widehat{C}_{M,\lambda}^{-1}C_{M,\lambda}^{1/2}\bigr\|
\le\frac85,
\end{equation}
\end{lemma}

\begin{proof}
Work on the event of Lemma~\ref{lem:exceptional}, which has
probability at least $1-2\eta$ over the random features.
By the definition of $\mathcal{A}_1$ in \eqref{eq:omegas}, every
$\bfx\in\mathcal{A}_1$ satisfies
\begin{equation}\label{eq:goodfeature}
\bigl\|C_{M,\lambda}^{-1/2}\phi_M(\bfx)\bigr\|_2^{2}
\le8M_{\alpha}\lambda^{-\alpha},
\end{equation}
and Lemma~\ref{lem:exceptional} also gives
$\rho_X(\mathcal{A}_2)\le \frac{\eta}{n}$ and
$\int_{\mathcal{A}_2}\bigl\|C_{M,\lambda}^{-1/2}\phi_M(\bfx)\bigr\|_2^{2}
d\rho_X(\bfx)\le\frac18$.

For this fixed random features realization, we define random variables
\[
\bfz_i=C_{M,\lambda}^{-1/2}\phi_M(\bfx_i),
\qquad
Z_i=\bfz_i\bfz_i^\top\mathbf{1}\{\bfx_i\in\mathcal{A}_1\},
\qquad i=1,\dots,n.
\]
The matrices $Z_1,\dots,Z_n$ are i.i.d., since the samples
$\bfx_1,\dots,\bfx_n$ are i.i.d. By \eqref{eq:goodfeature}, on $\mathcal{A}_1$, it holds that
\begin{equation}\label{eq:Ybound}
0\preceq Z_i\preceq\|\bfz_i\|_2^{2}I
\preceq8M_{\alpha}\lambda^{-\alpha}I.
\end{equation}
Since $Z_i^{2}=\bfz_i(\bfz_i^\top\bfz_i)\bfz_i^\top\mathbf{1}^2\{\bfx_i\in\mathcal{A}_1\}=\|\bfz_i\|_2^{2}\mathbf{1}\{\bfx_i\in\mathcal{A}_1\}Z_i$,
taking expectations and using \eqref{eq:Ybound} gives
\begin{equation}\label{eq:Ysecond}
\mathbb{E}\bigl[Z_i^{2}\bigr]
\preceq8M_{\alpha}\lambda^{-\alpha}\mathbb{E}\bigl[Z_i\bigr] \preceq 8M_{\alpha}\lambda^{-\alpha}I,
\end{equation}
where the last inequality uses $\mathbb{E}\bigl[Z_i\bigr]\preceq  C_{M,\lambda}^{-1/2}\mathbb{E}\bigl[\phi_M(\bfx)\phi_M(\bfx)^\top\bigr] C_{M,\lambda}^{-1/2}
\preceq C_{M,\lambda}^{-1/2}C_MC_{M,\lambda}^{-1/2}
\preceq I$ since $C_M\preceq C_{M,\lambda}$. From \eqref{eq:Ybound} and \eqref{eq:Ysecond}, we can further get 
$$
\|Z_i-\mathbb E Z_i\| \le 8 M_{\alpha}\lambda^{-\alpha}
$$
since $Z_i-\mathbb E Z_i \preceq Z_i \preceq 8 M_{\alpha}\lambda^{-\alpha} I$ and $Z_i-\mathbb E Z_i \succeq -\mathbb E Z_i \succeq - 8 M_{\alpha}\lambda^{-\alpha}I$. And 
$$
\left\|\frac{1}{n}\sum_{i=1}^n\mathbb E\bigl[(Z_i-\mathbb E Z_i)^2\bigr]\right\|
\le \left\|\frac{1}{n}\sum_{i=1}^n\mathbb E\bigl[Z_i^2\bigr]\right\|
\le 8 M_{\alpha}\lambda^{-\alpha} 
$$
Now we apply the matrix Bernstein inequality in
Lemma~\ref{lem:mbern} to $Z_i-\mathbb{E}Z_i$ with  $B=8M_{\alpha}\lambda^{-\alpha}$ and $V=8M_{\alpha}\lambda^{-\alpha}$, and we take $U=1/4$, 
\[
\begin{aligned}
\mathbb{P}\Bigl\{
\Bigl\|\frac1n\sum_{i=1}^n\bigl(Z_i-\mathbb{E}Z_i\bigr)\Bigr\|
>\frac14
\Bigr\}
&\le 2M\exp\left(
-\frac{n/16}{(16+\frac43)M_{\alpha}\lambda^{-\alpha}}
\right)\\
&=2M\exp\left(
-\frac{3n}{832M_{\alpha}\lambda^{-\alpha}}
\right)
\le\eta,
\end{aligned}
\]
where the last inequality uses \eqref{eq:covbudget} with
$C=\frac{832}{3}$. Thus with probability at least $1-\eta$, we have
\begin{align}\label{eq:bernsteinresult}
\left\|\frac1n\sum_{i=1}^n\bigl(Z_i-\mathbb{E}Z_i\bigr)\right\|
\le \frac14.
\end{align}

On the event of Lemma~\ref{lem:exceptional}, since $\rho_X(\mathcal{A}_2)\le\eta/n$, we have
\[
\mathbb{P}\Bigl\{\bigcup_{i=1}^n\{\bfx_i\in\mathcal{A}_2\}\Bigr\}
\le n\rho_X(\mathcal{A}_2)\le\eta.
\]
Thus the complementary event, where none of the samples lie in $\mathcal{A}_2$, has probability at least $1-\eta$. On this event, we have $\mathbf{1}\{\bfx_i\in\mathcal{A}_1\}=1$ for all $i=1,\dots,n$, so 
\begin{align}\label{eq:identityonsamples}
\frac1n\sum_{i=1}^nZ_i
=\frac1n\sum_{i=1}^n\bfz_i\bfz_i^\top
=C_{M,\lambda}^{-1/2}\widehat{C}_MC_{M,\lambda}^{-1/2}.
\end{align}
On the other hand, since $\mathbf{1}\{\bfx\in\mathcal{A}_1\}
=1-\mathbf{1}\{\bfx\in\mathcal{A}_2\}$,
\begin{align}\label{eq:Zimean}
\mathbb{E}\bigl[Z_i\bigr]
=C_{M,\lambda}^{-1/2}C_MC_{M,\lambda}^{-1/2}
-\mathbb{E}\Bigl[
\bfz_i\bfz_i^\top\mathbf{1}\{\bfx_i\in\mathcal{A}_2\}
\Bigr],
\end{align}
where we used
$\mathbb{E}\bigl[\phi_M(\bfx)\phi_M(\bfx)^\top\bigr]=C_M$.
The subtracted part is a positive matrix, and its norm is at most
\begin{equation}\label{eq:removedpart}
\Bigl\|\mathbb{E}\Bigl[
\bfz_i\bfz_i^\top\mathbf{1}\{\bfx_i\in\mathcal{A}_2\}
\Bigr]\Bigr\|
\le\mathbb{E}\Bigl[\|\bfz_i\|_2^{2}\mathbf{1}\{\bfx_i\in\mathcal{A}_2\}\Bigr]
=\int_{\mathcal{A}_2}\bigl\|C_{M,\lambda}^{-1/2}\phi_M(\bfx)\bigr\|_2^{2}
d\rho_X(\bfx)
\le\frac18,
\end{equation}
where the last inequality follows from Lemma~\ref{lem:exceptional}.

Combining \eqref{eq:bernsteinresult}-\eqref{eq:removedpart}, we obtain by the triangle inequality that 
\[
\left\|C_{M,\lambda}^{-1/2}\bigl(\widehat{C}_M-C_M\bigr)C_{M,\lambda}^{-1/2}\right\|
\le \left\|\frac1n\sum_{i=1}^n\bigl(Z_i-\mathbb{E}Z_i\bigr)\right\|
+\Bigl\|\mathbb{E}\Bigl[
\bfz_i\bfz_i^\top\mathbf{1}\{\bfx_i\in\mathcal{A}_2\}
\Bigr]\Bigr\|
\le \frac14+\frac18=\frac38
\]
with probability at least $1-2\eta$ over the
samples for the fixed random features. By Lemma E.2 of \cite{blanchard2010optimal}, with probability at least $1-2\eta$ over the samples for the fixed random features, we have
\begin{align}\label{eq:order}
\|\widehat{C}_{M,\lambda}^{-1/2}C_{M,\lambda}^{1/2}\| \leq \sqrt{\frac85}.
\end{align}

For \eqref{eq:covfactor}, note that
\[
\bigl\|S_M\widehat{C}_{M,\lambda}^{-1/2}\bigr\|^{2}
=\bigl\|\widehat{C}_{M,\lambda}^{-1/2}S_M^{*}S_M
\widehat{C}_{M,\lambda}^{-1/2}\bigr\|
=\bigl\|\widehat{C}_{M,\lambda}^{-1/2}C_M
\widehat{C}_{M,\lambda}^{-1/2}\bigr\|
\le\bigl\|\widehat{C}_{M,\lambda}^{-1/2}C_{M,\lambda}
\widehat{C}_{M,\lambda}^{-1/2}\bigr\|
\le\frac85,
\]
where the first equality uses the relation
$\|T\|^{2}=\|T^{*}T\|$ with $T=S_M\widehat{C}_{M,\lambda}^{-1/2}$,
and the last step uses \eqref{eq:order}.
Thus we have
\[
\bigl\|S_M\widehat{C}_{M,\lambda}^{-1}C_{M,\lambda}^{1/2}\bigr\|
\le\bigl\|S_M\widehat{C}_{M,\lambda}^{-1/2}\bigr\|
\bigl\|\widehat{C}_{M,\lambda}^{-1/2}C_{M,\lambda}^{1/2}\bigr\|
\le\sqrt{\frac85}\cdot\sqrt{\frac85}=\frac85.
\]

Finally, we combine the probabilities.
The event of Lemma~\ref{lem:exceptional} has probability
at least $1-2\eta$ over the random features. Conditionally on this event, the bound \eqref{eq:bernsteinresult} and the identity \eqref{eq:identityonsamples} each hold with
probability at least $1-\eta$ over the samples, so they hold
simultaneously with probability at least $1-2\eta$.
The bounds \eqref{eq:order}, \eqref{eq:covwhitened}, and
\eqref{eq:covfactor} are deterministic consequences of
\eqref{eq:bernsteinresult} and \eqref{eq:identityonsamples}, so
they hold simultaneously with probability at least
$(1-2\eta)(1-2\eta)\ge1-4\eta$ over the features and the samples.
Thus we complete the proof.
\end{proof}

\begin{proposition}\label{prop:fluctuation}
Under Assumptions \ref{ass1}, \ref{ass3}, \ref{ass4}, \ref{ass5} and \ref{ass7}, if 
$0<\lambda\le\min\{1,\|L_K\|\}$ and for any $\alpha_0 < \alpha \le 1$, we assume
\begin{itemize}
\item[(i)] (Misspecified case) If $0<r<\frac12$ and the number of random features satisfies
\begin{equation}\label{eq:M1}
M\ge 
\begin{cases}
C\lambda^{-\zeta}\log\frac{cn}{\lambda\eta}, & \zeta+\alpha\ne1,\\[2mm]
\displaystyle C\lambda^{-\zeta}\log\frac{cn}{\lambda\eta} \left[1+\log\frac{1+\|L_K\|}{\lambda}\right], & \zeta+\alpha=1.
\end{cases}
\end{equation}
\item[(ii)] (Well-specified case) If $\frac12\le r\le1$ and the number of random features satisfies 
\begin{equation}\label{eq:M2}
  M\ge 
\begin{cases}
C\lambda^{-\{\zeta+(1+\gamma-\zeta)(2r-1)\}}\log\frac{cn}{\lambda\eta}, & \zeta+\alpha\ne1,\\[2mm]
\displaystyle C\lambda^{-\{\zeta+(1+\gamma-\zeta)(2r-1)\}}\log\frac{cn}{\lambda\eta} \left[1+\log\frac{1+\|L_K\|}{\lambda}\right], & \zeta+\alpha=1.
\end{cases}
\end{equation}
\end{itemize}
For both cases, we further assume the sample size $n$ satisfies \eqref{eq:covbudget}. Then for $\eta\in (0,\frac{1}{9})$, with probability at least $1-9\eta$ over the features and the samples,
\begin{equation}
\begin{aligned}
&\left\|\widehat{C}_{M,\lambda}^{-1/2}C_{M,\lambda}^{1/2}\right\|^2\left\|C_{M,\lambda}^{-1/2}\left(\overline S_{M}^*\bigl(f_\rho-L_ML_{M,\lambda}^{-1}f_\rho\bigr)- S_M^*\bigl(f_\rho-L_ML_{M,\lambda}^{-1}f_\rho\bigr) \right)\right\|_2 \\
\leq & CR\left(
\sqrt{\frac{M_\alpha\lambda^{2r-\alpha}}{n\eta}}+\lambda^r
\right).
\label{eq:fluctuationbound}
\end{aligned}
\end{equation}
\end{proposition}

\begin{proof}
We first focus on $\left\|C_{M,\lambda}^{-1/2}\left(\overline S_{M}^*\bigl(f_\rho-L_ML_{M,\lambda}^{-1}f_\rho\bigr)- S_M^*\bigl(f_\rho-L_ML_{M,\lambda}^{-1}f_\rho\bigr) \right)\right\|_2$. For this fixed random features realization, we define random variables
\[
\bfz_i=C_{M,\lambda}^{-1/2}\phi_M(\bfx_i),
\qquad \bfd_i=(f_\rho-L_ML_{M,\lambda}^{-1}f_\rho)(\bfx_i),
\qquad i=1,\dots,n.
\]
By Definition \ref{def1}, we have 
\[
\begin{aligned}
  &C_{M,\lambda}^{-1/2}\left(\overline S_{M}^*\bigl(f_\rho-L_ML_{M,\lambda}^{-1}f_\rho\bigr)- S_M^*\bigl(f_\rho-L_ML_{M,\lambda}^{-1}f_\rho\bigr) \right)
  = \frac1n\sum_{i=1}^n\left(\bfz_i\bfd_i-\mathbb E[\bfz_i\bfd_i]\right) \\
=& \frac1n\sum_{i=1}^n \Bigl(\bfz_i\bfd_i\mathbf 1\{\bfx_i\in\mathcal A_1\}-\mathbb E\bigl[\bfz\bfd\mathbf 1\{\bfx \in\mathcal A_1\}\bigr]\Bigr)+\frac1n\sum_{i=1}^n \Bigl(\bfz_i\bfd_i\mathbf 1\{\bfx_i\in\mathcal A_2\}-\mathbb E\bigl[\bfz\bfd\mathbf 1\{\bfx \in\mathcal A_2\}\bigr]\Bigr),
\end{aligned}
\]
where $\mathcal A_1$ and $\mathcal A_2$ are defined as in \eqref{eq:omegas}. We now proceed to bound the two terms separately. 

For the first term, we define 
\[
\chi_i=\frac1n\Bigl(\bfz_i\bfd_i\mathbf 1\{\bfx_i\in\mathcal A_1\}-\mathbb E\bigl[\bfz\bfd\mathbf 1\{\bfx \in\mathcal A_1\}\bigr]\Bigr),\quad i=1,\dots,n.
\]
Then $\chi_1,\dots,\chi_n$ are i.i.d.\ random vectors in $\mathbb R^M$ with $\mathbb E\chi_i=0$. Moreover, we have
\begin{align*}
\sum_{i=1}^n\mathbb E\|\chi_i\|_2^2
\le \frac1n\mathbb E\Bigl[\|\bfz\|_2^2\bfd^2\mathbf 1\{\bfx\in\mathcal A_1\}\Bigr]& \leq \frac{8M_{\alpha}\lambda^{-\alpha}}{n}\mathbb E\bigl[\bfd^2\mathbf 1\{\bfx\in\mathcal A_1\}\bigr] \leq \frac{8M_{\alpha}\lambda^{-\alpha}}{n}\mathbb E\bigl[\bfd^2\bigr]\\
&= \frac{8M_{\alpha}\lambda^{-\alpha}}{n}\|\bfd\|_{L^2(\rho_X)}^2
\end{align*}
where we use $\|\bfz_i\|_2^2\le 8M_{\alpha}\lambda^{-\alpha}$ on the set $\mathcal A_1$. Hence, by Chebyshev's inequality, for any $t>0$,
\[
\mathbb P\Bigl(\Bigl\|\sum_{i=1}^n\chi_i\Bigr\|_2>t\Bigr)\le \frac{1}{t^2} \mathbb E \Bigl[\Bigl\|\sum_{i=1}^n\chi_i\Bigr\|_2^2\Bigr]
= \frac{1}{t^2}\sum_{i=1}^n\mathbb E\|\chi_i\|_2^2
\le  \frac{8M_{\alpha}\lambda^{-\alpha}}{nt^2}\|\bfd\|_{L^2(\rho_X)}^2,
\]
where the second equality uses 
$\mathbb E\langle\chi_i,\chi_j\rangle=\langle\mathbb E\chi_i,\mathbb E\chi_j\rangle=0$ 
for $i\neq j$, which follows from the conditional independence and the 
centering $\mathbb E\chi_i=0$.
Setting the right-hand side equal to $\eta$, we obtain that with probability 
at least $1-\eta$,
\begin{equation}\label{eq:trunc}
\Bigl\|\sum_{i=1}^n\chi_i\Bigr\|_2\le \sqrt{\frac{8M_{\alpha}\lambda^{-\alpha}\|\bfd\|_{L^2(\rho_X)}^2}{n\eta}}.
\end{equation}

For the second term, on the event of Lemma~\ref{lem:exceptional}, since $\rho_X(\mathcal{A}_2)\le\eta/n$, we have
\[
\mathbb{P}\Bigl\{\bigcup_{i=1}^n\{\bfx_i\in\mathcal{A}_2\}\Bigr\}
\le n\rho_X(\mathcal{A}_2)\le\eta.
\]
Thus the complementary event, where none of the samples lie in $\mathcal{A}_2$, has probability at least $1-\eta$. On this event, we have $\mathbf{1}\{\bfx_i\in\mathcal{A}_1\}=1$ for all $i=1,\dots,n$, so 
\begin{align}\label{eq:sample_A2_zero}
\frac1n\sum_{i=1}^n \bfz_i\bfd_i\mathbf 1\{\bfx_i\in\mathcal A_2\} = 0
\end{align}
On the other hand, by Cauchy--Schwarz,
\begin{equation}\label{eq:expectation_A2_bound}
\begin{aligned}
\bigl\|\mathbb E\bigl[\bfz\bfd\mathbf 1\{\bfx \in\mathcal A_2\}\bigr]\bigr\|_2
\le\|\bfd\|_{L^2(\rho_X)}
\Bigl(\int_{\mathcal A_2}\|\bfz\|_2^2d\rho_X(\bfx)\Bigr)^{1/2}
\le\frac{\|\bfd\|_{L^2(\rho_X)}}{\sqrt8},
\end{aligned}
\end{equation}
where the last inequality uses the second conclusion of Lemma \ref{lem:exceptional}.

On the event of Lemma~\ref{lem:feat}, we have $\|d\|_{L^2(\rho_X)} \le CR \lambda^r$, so Plugging this into \eqref{eq:trunc} and \eqref{eq:expectation_A2_bound}, we get 
\begin{align*}
\left\|C_{M,\lambda}^{-1/2}\left(\overline S_{M}^*\bigl(f_\rho-L_ML_{M,\lambda}^{-1}f_\rho\bigr)- S_M^*\bigl(f_\rho-L_ML_{M,\lambda}^{-1}f_\rho\bigr) \right)\right\|_2 
\leq  CR\left(
\sqrt{\frac{M_\alpha\lambda^{2r-\alpha}}{n\eta}}+\lambda^r
\right).
\end{align*}
Combining the above inequality and Lemma \ref{lem:covariance} that $\bigl\|\widehat{C}_{M,\lambda}^{-1/2}C_{M,\lambda}^{1/2}\bigr\|
\le \sqrt{\frac85}$, we can obtain the desired bound for all scenarios. In the following, we verify the conditions on the number of random features, and the probability. 

We first check the conditions on the number of random features. We show that the conditions \eqref{eq:M1} for $0<r<\frac12$ and \eqref{eq:M2} for $\frac12\le r\le1$ covers all conditions on $M$ required in the used lemmas. Recall $t_\eta=\log\frac{4\kappa^2}{\lambda\eta}$ from Proposition \ref{prop:mergedbudget}. Since $cn\ge4\kappa^2$ for $c$ large enough, we have
\[
\log\frac{cn}{\lambda\eta}\ge\log\frac{4\kappa^2}{\lambda\eta}=t_\eta.
\]
Moreover, \eqref{eq:covbudget} gives $n\ge\frac{832}{3}M_\alpha\lambda^{-\alpha}$ and $\lambda\le1$, hence
$M_\alpha\lambda^{1-\alpha}=\lambda\cdot M_\alpha\lambda^{-\alpha}\le\frac{3n}{832}$, and with $\eta<1/9$,
\[
\log \frac{c'(n+M_\alpha\lambda^{-\alpha})}{\eta^2}
=\log \frac{c'(n\lambda+M_\alpha\lambda^{1-\alpha})}{\lambda\eta^2}
\le \log \frac{c'n\bigl(1+\frac{3}{832}\bigr)}{\lambda\eta^2}
\le 2\log \frac{cn}{\lambda\eta}
\]
for $c$ large enough, where $c'$ is the constant in \eqref{eq:exceptionalbudget}. Combining this with $\lambda^{-\zeta}$ in \eqref{eq:M1} and $\lambda^{-\{\zeta+(1+\gamma-\zeta)(2r-1)\}}\ge\lambda^{-\zeta}$ in \eqref{eq:M2}, the conditions \eqref{eq:M1} and \eqref{eq:M2} imply the condition \eqref{eq:exceptionalbudget} in both cases of Lemma \ref{lem:exceptional}.
The extra factor $\bigl[1+\log\frac{1+\|L_K\|}{\lambda}\bigr]$ when $\zeta+\alpha=1$ matches that of \eqref{eq:exceptionalbudget}. The same comparison, together with $\log\frac{cn}{\lambda\eta}\ge t_\eta$, shows that \eqref{eq:M1} and \eqref{eq:M2} imply the conditions of Proposition \ref{prop:mergedbudget}, i.e.\ \eqref{eq:budgetlow} for $0<r<\frac12$ and \eqref{eq:interpbudget} for $\frac12\le r\le1$. Since \eqref{eq:exceptionalbudget} is also the condition required by Lemma \ref{lem:covariance}, all conditions on $M$ in the used lemmas are satisfied by \eqref{eq:M1} and \eqref{eq:M2}.

Combining the probabilities, the bound holds with probability at least
$1-9\eta$ over the features and the samples: $4\eta$ from the
event of Proposition \ref{prop:mergedbudget}, $4\eta$ from Lemma
\ref{lem:covariance}, whose event already contains the event of Lemma
\ref{lem:exceptional} and the event that no sample lies in $\mathcal A_2$,
$\eta$ from the Chebyshev bound \eqref{eq:trunc}. Thus we complete the proof.

\end{proof}

\section{Noise Error}

\begin{proposition}\label{lem:noiseerror}
Under Assumptions \ref{ass1}, \ref{ass2}, \ref{ass5} and \ref{ass7}, if $0<\lambda\le1$ and for any $\alpha_0 < \alpha \le 1$, suppose that the number of random features $M$ satisfies \eqref{eq:exceptionalbudget} and \eqref{eq:Mbudget}, and the sample size $n$ satisfies \eqref{eq:covbudget}. Then for $\eta\in(0,1/9)$, with probability at least $1-9\eta$ over the features and the samples,
\begin{equation}
\begin{aligned}
\left\|\widehat{C}_{M,\lambda}^{-1/2}C_{M,\lambda}^{1/2}\right\|^2\left\|C_{M,\lambda}^{-1/2}\widehat{S}_{M}^*\widehat{\boldsymbol{\varepsilon}}\right\|_2 \leq  C\left(
\frac{B\sqrt{M_\alpha\lambda^{-\alpha}}\log\frac2\eta}{n}
+\sigma\sqrt{\frac{{\cal N}(\lambda)\log\frac2\eta}{n}}
\right).
\label{eq:noiseerror}
\end{aligned}
\end{equation}
\end{proposition}

\begin{proof}
We first focus on $\left\|C_{M,\lambda}^{-1/2}\widehat{S}_{M}^*\widehat{\boldsymbol{\varepsilon}}\right\|_2$. For this fixed random features realization, we define random variables
\[
\bfz_i=C_{M,\lambda}^{-1/2}\phi_M(\bfx_i),
\qquad \varepsilon_i=y_i-f_\rho(\bfx_i),
\qquad i=1,\dots,n.
\]
By Definition \ref{def1}, we have
\begin{equation}\label{eq:noise_decomposition}
\begin{aligned}
C_{M,\lambda}^{-1/2}\widehat{S}_{M}^*\widehat{\boldsymbol{\varepsilon}}
=\frac1n\sum_{i=1}^n\bfz_i\varepsilon_i =\frac1n\sum_{i=1}^n\bfz_i\varepsilon_i\mathbf 1\{\bfx_i\in\mathcal A_1\}
+\frac1n\sum_{i=1}^n\bfz_i\varepsilon_i\mathbf 1\{\bfx_i\in\mathcal A_2\},
\end{aligned}
\end{equation}
where $\mathcal A_1$ and $\mathcal A_2$ are defined as in \eqref{eq:omegas}. We now proceed to bound the two sums separately.

For the first sum, we define
\[
\xi_i=\varepsilon_i\bfz_i\mathbf 1\{\bfx_i\in\mathcal A_1\},
\quad i=1,\dots,n.
\]
Then $\xi_1,\dots,\xi_n$ are i.i.d.\ random vectors in $\mathbb R^M$ with $\mathbb E\xi_i=0$, since $\mathbb E[\varepsilon|\bfx]=0$ by Assumption \ref{ass2}. For every integer $q\ge2$, its $q$-th moment can be bounded as
\begin{align*}
\mathbb E\|\xi_i\|_2^q
&=\mathbb E\Bigl[|\varepsilon_i|^q\|\bfz_i\|_2^q\mathbf 1\{\bfx_i\in\mathcal A_1\}\Bigr]\le\bigl(8M_\alpha\lambda^{-\alpha}\bigr)^{q/2-1}\mathbb E\Bigl[\|\bfz_i\|_2^2\mathbb E\bigl[|\varepsilon_i|^q\big|\bfx_i\bigr]\Bigr]\\
&\le\bigl(8M_\alpha\lambda^{-\alpha}\bigr)^{q/2-1}\cdot\frac12q!\sigma^2B^{q-2}\mathbb E\|\bfz_i\|_2^2\le\frac12q!\bigl(B\sqrt{8M_\alpha\lambda^{-\alpha}}\bigr)^{q-2}\cdot 4\sigma^2{\cal N}(\lambda),
\end{align*}
where the first inequality follows from the fact that $\|\bfz_i\|_2^2\le 8M_{\alpha}\lambda^{-\alpha}$ on the set $\mathcal A_1$, the second inequality uses Assumption \ref{ass2}, and the last inequality uses the bound on $\mathbb E\|\bfz_i\|_2^2$ derived below
\begin{align*}
\mathbb E\|\bfz\|_2^2
&=\mathbb E\bigl[\phi_M(\bfx)^\top C_{M,\lambda}^{-1}\phi_M(\bfx)\bigr]=\mathbb E\Bigl[\text{Tr}\bigl(C_{M,\lambda}^{-1}\phi_M(\bfx)\phi_M(\bfx)^\top\bigr)\Bigr]\\
&=\text{Tr}\Bigl(C_{M,\lambda}^{-1}\mathbb E\bigl[\phi_M(\bfx)\phi_M(\bfx)^\top\bigr]\Bigr)=\text{Tr}\bigl(C_{M,\lambda}^{-1}C_M\bigr)
={\cal N}_M(\lambda)\le 4{\cal N}(\lambda),
\end{align*}
where the equalities use the cyclicity of the trace and Definition \ref{def1}, the last inequality is (c) of Lemma \ref{lem:feat}. Applying Lemma \ref{lem:bennett} to $\xi_1,\dots,\xi_n$ with $\widetilde B=B\sqrt{8M_\alpha\lambda^{-\alpha}}$ and $\widetilde\sigma^2=4\sigma^2{\cal N}(\lambda)$, we obtain with probability at least $1-\eta$,
\begin{equation}\label{eq:noisetrunc}
\Bigl\|\frac1n\sum_{i=1}^n\xi_i\Bigr\|_2
\le\frac{2B\sqrt{8M_\alpha\lambda^{-\alpha}}\log\frac2\eta}{n}
+\sigma\sqrt{\frac{8{\cal N}(\lambda)\log\frac2\eta}{n}}.
\end{equation}

For the second sum, on the event of Lemma \ref{lem:exceptional}, since $\rho_X(\mathcal{A}_2)\le\eta/n$, we have
\[
\mathbb{P}\Bigl\{\bigcup_{i=1}^n\{\bfx_i\in\mathcal{A}_2\}\Bigr\}
\le n\rho_X(\mathcal{A}_2)\le\eta.
\]
Thus the complementary event, where none of the samples lie in $\mathcal{A}_2$, has probability at least $1-\eta$. On this event, we have $\mathbf{1}\{\bfx_i\in\mathcal{A}_2\}=0$ for all $i=1,\dots,n$, so
\begin{align}\label{eq:noise_A2_zero}
\frac1n\sum_{i=1}^n \bfz_i\varepsilon_i\mathbf 1\{\bfx_i\in\mathcal A_2\} = 0.
\end{align}

Plugging \eqref{eq:noisetrunc} and \eqref{eq:noise_A2_zero} into the decomposition \eqref{eq:noise_decomposition}, and applying Lemma \ref{lem:covariance}, we finally get the desired bound
\begin{align*}
\left\|\widehat{C}_{M,\lambda}^{-1/2}C_{M,\lambda}^{1/2}\right\|^2\left\|C_{M,\lambda}^{-1/2}\widehat{S}_{M}^*\widehat{\boldsymbol{\varepsilon}}\right\|_2 
\leq  \frac{8}{5}\Bigl\|\frac1n\sum_{i=1}^n\xi_i\Bigr\|_2
\leq  C\left(
\frac{B\sqrt{M_\alpha\lambda^{-\alpha}}\log\frac2\eta}{n}
+\sigma\sqrt{\frac{{\cal N}(\lambda)\log\frac2\eta}{n}}
\right).
\end{align*}

The conditions on $M$ are exactly what the used lemmas require: \eqref{eq:exceptionalbudget} is the condition of Lemma \ref{lem:exceptional} and Lemma \ref{lem:covariance}, and \eqref{eq:Mbudget} is the condition of Lemma \ref{lem:feat}. Combining the probabilities, the bound holds with probability at least $1-9\eta$ over the features and the samples: $4\eta$ from the event of Lemma \ref{lem:feat}, $4\eta$ from Lemma \ref{lem:covariance}, whose event already contains the event of Lemma \ref{lem:exceptional} and the event that no sample lies in $\mathcal A_2$, and $\eta$ from Lemma \ref{lem:bennett} in \eqref{eq:noisetrunc}. Thus we complete the proof.
\end{proof}

\subsection{Noise Error under Heavy-tailed Conditions}

\begin{proposition}\label{lem:noiseerror_h}
Under Assumptions \ref{ass1}, \ref{ass5}, \ref{ass6} and \ref{ass7}, if $0<\lambda\le1$ and for any $\alpha_0 < \alpha \le 1$, suppose that the number of random features $M$ satisfies \eqref{eq:exceptionalbudget} and \eqref{eq:Mbudget}, and the sample size $n$ satisfies \eqref{eq:covbudget}. Then for $\eta\in(0,1/9)$, with probability at least $1-9\eta$ over the features and the samples,
\begin{equation}
\begin{aligned}
\left\|\widehat{C}_{M,\lambda}^{-1/2}C_{M,\lambda}^{1/2}\right\|^2\left\|C_{M,\lambda}^{-1/2}\widehat{S}_{M}^*\widehat{\boldsymbol{\varepsilon}}\right\|_2 \leq  C\left(
\frac{B^{\frac1p}\bigl(M_\alpha\lambda^{-\alpha}\bigr)^{\frac12-\frac1p}{\cal N}(\lambda)^{\frac1p}}{\eta^{\frac1p}n^{1-\frac1p}}
+\sigma\sqrt{\frac{{\cal N}(\lambda)\log\frac{c}{\eta}}{n}}
\right).
\label{eq:noiseerror_h}
\end{aligned}
\end{equation}
\end{proposition}

\begin{proof}
We first focus on $\left\|C_{M,\lambda}^{-1/2}\widehat{S}_{M}^*\widehat{\boldsymbol{\varepsilon}}\right\|_2$. For this fixed random features realization, we define random variables
\[
\bfz_i=C_{M,\lambda}^{-1/2}\phi_M(\bfx_i),
\qquad \varepsilon_i=y_i-f_\rho(\bfx_i),
\qquad i=1,\dots,n.
\]
By Definition \ref{def1}, we have
\begin{equation}\label{eq:noise_decomposition_h}
\begin{aligned}
C_{M,\lambda}^{-1/2}\widehat{S}_{M}^*\widehat{\boldsymbol{\varepsilon}}
=\frac1n\sum_{i=1}^n\bfz_i\varepsilon_i =\frac1n\sum_{i=1}^n\bfz_i\varepsilon_i\mathbf 1\{\bfx_i\in\mathcal A_1\}
+\frac1n\sum_{i=1}^n\bfz_i\varepsilon_i\mathbf 1\{\bfx_i\in\mathcal A_2\},
\end{aligned}
\end{equation}
where $\mathcal A_1$ and $\mathcal A_2$ are defined as in \eqref{eq:omegas}. We now proceed to bound the two sums separately.

For the first sum, we define
\[
\xi_i=\varepsilon_i\bfz_i\mathbf 1\{\bfx_i\in\mathcal A_1\},
\quad i=1,\dots,n.
\]
Then $\xi_1,\dots,\xi_n$ are i.i.d.\ random vectors in $\mathbb R^M$ with $\mathbb E\xi_i=0$, since $\mathbb E[\varepsilon|\bfx]=0$ by Assumption \ref{ass6}. We first bound the second moment and the $p$-th moment of $\xi_i$. The same computation as in Proposition \ref{lem:noiseerror} gives
\begin{align*}
\mathbb E\|\bfz\|_2^2
&=\mathbb E\bigl[\phi_M(\bfx)^\top C_{M,\lambda}^{-1}\phi_M(\bfx)\bigr]=\mathbb E\Bigl[\text{Tr}\bigl(C_{M,\lambda}^{-1}\phi_M(\bfx)\phi_M(\bfx)^\top\bigr)\Bigr]\\
&=\text{Tr}\Bigl(C_{M,\lambda}^{-1}\mathbb E\bigl[\phi_M(\bfx)\phi_M(\bfx)^\top\bigr]\Bigr)=\text{Tr}\bigl(C_{M,\lambda}^{-1}C_M\bigr)
={\cal N}_M(\lambda)\le 4{\cal N}(\lambda),
\end{align*}
where the equalities use the cyclicity of the trace and Definition \ref{def1}, and the last inequality is (c) of Lemma \ref{lem:feat}. Hence, by Assumption \ref{ass6},
\begin{align*}
\mathbb E\|\xi_i\|_2^2
&=\mathbb E\Bigl[\varepsilon^2\|\bfz\|_2^2\mathbf 1\{\bfx\in\mathcal A_1\}\Bigr]
\le\mathbb E\Bigl[\|\bfz\|_2^2\mathbb E\bigl[\varepsilon^2\big|\bfx\bigr]\Bigr]
\le\sigma^2\mathbb E\|\bfz\|_2^2
\le4\sigma^2{\cal N}(\lambda),
\end{align*}
and similarly, since $\|\bfz\|_2^{p-2}\mathbf 1\{\bfx\in\mathcal A_1\}\le\bigl(8M_\alpha\lambda^{-\alpha}\bigr)^{p/2-1}$,
\begin{align*}
\mathbb E\|\xi_i\|_2^p
&=\mathbb E\Bigl[|\varepsilon|^p\|\bfz\|_2^p\mathbf 1\{\bfx\in\mathcal A_1\}\Bigr]
\le\bigl(8M_\alpha\lambda^{-\alpha}\bigr)^{p/2-1}\mathbb E\Bigl[\|\bfz\|_2^2\mathbb E\bigl[|\varepsilon|^p\big|\bfx\bigr]\Bigr]\\
&\le B\bigl(8M_\alpha\lambda^{-\alpha}\bigr)^{p/2-1}\mathbb E\|\bfz\|_2^2
\le4B\bigl(8M_\alpha\lambda^{-\alpha}\bigr)^{p/2-1}{\cal N}(\lambda).
\end{align*}
Applying Lemma \ref{lem:fuknagaev} to $\xi_1,\dots,\xi_n$ with $\widetilde\sigma^2=4\sigma^2{\cal N}(\lambda)$ and $\widetilde B=4B\bigl(8M_\alpha\lambda^{-\alpha}\bigr)^{p/2-1}{\cal N}(\lambda)$, we obtain with probability at least $1-\eta$,
\begin{equation}\label{eq:noisetrunc_h}
\Bigl\|\frac1n\sum_{i=1}^n\xi_i\Bigr\|_2
\le\left(\frac{8c_pB\bigl(8M_\alpha\lambda^{-\alpha}\bigr)^{p/2-1}{\cal N}(\lambda)}{\eta n^{p-1}}\right)^{1/p}
+\sigma\sqrt{\frac{4{\cal N}(\lambda)\log(2c_p/\eta)}{c_p^{\prime}n}},
\end{equation}
where $c_p,c_p^{\prime}$ are the constants in Lemma \ref{lem:fuknagaev}.

For the second sum, on the event of Lemma \ref{lem:exceptional}, since $\rho_X(\mathcal{A}_2)\le\eta/n$, we have
\[
\mathbb{P}\Bigl\{\bigcup_{i=1}^n\{\bfx_i\in\mathcal{A}_2\}\Bigr\}
\le n\rho_X(\mathcal{A}_2)\le\eta.
\]
Thus the complementary event, where none of the samples lie in $\mathcal{A}_2$, has probability at least $1-\eta$. On this event, we have $\mathbf{1}\{\bfx_i\in\mathcal{A}_2\}=0$ for all $i=1,\dots,n$, so
\begin{align}\label{eq:noise_A2_zero_h}
\frac1n\sum_{i=1}^n \bfz_i\varepsilon_i\mathbf 1\{\bfx_i\in\mathcal A_2\} = 0.
\end{align}

Plugging \eqref{eq:noisetrunc_h} and \eqref{eq:noise_A2_zero_h} into the decomposition \eqref{eq:noise_decomposition_h}, and applying Lemma \ref{lem:covariance} that $\bigl\|\widehat C_{M,\lambda}^{-1/2}C_{M,\lambda}^{1/2}\bigr\|^2\le\frac85$, we obtain
\begin{align*}
&\left\|\widehat{C}_{M,\lambda}^{-1/2}C_{M,\lambda}^{1/2}\right\|^2\left\|C_{M,\lambda}^{-1/2}\widehat{S}_{M}^*\widehat{\boldsymbol{\varepsilon}}\right\|_2
\le C\left(
\frac{B^{\frac1p}\bigl(M_\alpha\lambda^{-\alpha}\bigr)^{\frac12-\frac1p}{\cal N}(\lambda)^{\frac1p}}{\eta^{\frac1p}n^{1-\frac1p}}
+\sigma\sqrt{\frac{{\cal N}(\lambda)\log\frac{c}{\eta}}{n}}
\right),
\end{align*}

The conditions on $M$ are exactly what the used lemmas require: \eqref{eq:exceptionalbudget} is the condition of Lemma \ref{lem:exceptional} and Lemma \ref{lem:covariance}, and \eqref{eq:Mbudget} is the condition of Lemma \ref{lem:feat}. Combining the probabilities, the bound holds with probability at least $1-9\eta$ over the features and the samples: $4\eta$ from the event of Lemma \ref{lem:feat}, $4\eta$ from Lemma \ref{lem:covariance}, whose event already contains the event of Lemma \ref{lem:exceptional} and the event that no sample lies in $\mathcal A_2$, and $\eta$ from Lemma \ref{lem:fuknagaev} in \eqref{eq:noisetrunc_h}. Thus we complete the proof.
\end{proof}

\section{Proof of the Main Results}

\subsection{Proof of Theorem \ref{thm2}}

\begin{proof}
Since $2r+\gamma>\alpha_0$, there exists some $\alpha>\alpha_0$ and $2r+\gamma>\alpha$, such that Propositions \ref{prop:mergedbudget}, \ref{prop:fluctuation}, and \ref{lem:noiseerror} can be applied with this choice of $\alpha$. Combining these, we obtain bounds on the three terms in the error decomposition \eqref{eq:decompbound}
\begin{align*}
\bigl\|\widehat f_{\lambda,M}-f_\rho\bigr\|_{L^2(\rho_X)} \le CR\left(
\sqrt{\frac{M_\alpha\lambda^{2r-\alpha}}{n\eta}}+\lambda^r
\right)+C\left(
\frac{B\sqrt{M_\alpha\lambda^{-\alpha}}\log\frac2\eta}{n}
+\sigma\sqrt{\frac{{\cal N}(\lambda)\log\frac2\eta}{n}}
\right)
\end{align*}  
with probability at least $1-10\eta=1-\delta \in (0,1)$ over the features and the samples: $4\eta$ from the event of Lemma \ref{lem:feat}, $4\eta$ from Lemma \ref{lem:covariance}, whose event already contains the event of Lemma \ref{lem:exceptional} and the event that no sample lies in $\mathcal A_2$, $\eta$ from the Chebyshev bound \eqref{eq:trunc} and $\eta$ from the Bernstein bound \eqref{eq:noisetrunc}. We next simplify the bounds using the specified choice of $\lambda$ and the conditions in the theorem.

When $n$ satisfies that $n\lambda^\alpha\ge \frac{M_\alpha}{\eta \log^2 \frac{2}{\eta}}$, we have
\[
CR \sqrt{\frac{M_\alpha\lambda^{2r-\alpha}}{n\eta}}
= CR\lambda^r\sqrt{\frac{M_\alpha}{n\lambda^\alpha\eta}}
\le CR\lambda^r\log\frac2\eta.
\]
When $2r+\gamma>\alpha$ and $\lambda=n^{-\frac{1}{2r+\gamma}}$, we have
\[
\frac{CB\sqrt{M_\alpha\lambda^{-\alpha}}\log\frac2\eta}{n}
\le CB\sqrt{M_\alpha}\lambda^{2r+\gamma-\frac\alpha2}\log\frac2\eta
\le CB\sqrt{M_\alpha}\lambda^r\log\frac2\eta,
\]
where the last inequality uses $\lambda^{r+\gamma-\alpha/2}<\lambda^{\gamma/2}<1$. By
Assumption \ref{ass4} that $\mathcal N(\lambda)\le Q^2\lambda^{-\gamma}$ and $\lambda=n^{-\frac{1}{2r+\gamma}}$, we have
\[
C\sigma\sqrt{\frac{\mathcal N(\lambda)\log\frac2\eta}{n}}
\le C\sigma Q\lambda^{-\gamma/2} \lambda^{r+\gamma/2}\sqrt{\log\frac2\eta}
\le C\sigma Q\lambda^{r} \log\frac2\eta,
\]
where the last inequality uses $\log\frac2\eta > 1$ for $\eta \in (0,1/10)$. 

Combining these three estimates yields
\[
\bigl\|\widehat f_{\lambda,M}-f_\rho\bigr\|_{L^2(\rho_X)}
\le (CR+CB\sqrt{M_\alpha}+C\sigma Q)\lambda^r\log\frac{2}{\eta} = C_1n^{-\frac{r}{2r+\gamma}}\log\frac{c_1}{\delta},
\]
where $C_1=CR+CB\sqrt{M_\alpha}+C\sigma Q$ and $c_1=20$. In the following, we verify the conditions on the sample size, the number of random features.

We first verify the conditions on the sample size $n$. With the choice 
$\lambda=n^{-\frac{1}{2r+\gamma}}$ and $\eta=\frac{\delta}{10}$, the conditions on $n$ 
required above are the following three:
\begin{itemize}
\item[(i)] $\lambda\le\min\{1,\|L_K\|\}$ in Proposition \ref{prop:fluctuation}, which is 
equivalent to 
\[
n\ge \max\bigl\{1,\|L_K\|^{-(2r+\gamma)}\bigr\}.
\]
\item[(ii)] Condition \eqref{eq:covbudget} in Lemma \ref{lem:covariance}, i.e., 
$n\ge\frac{832}{3}M_{\alpha}\lambda^{-\alpha}\log\frac{2M}{\eta}$. Note that there exist $C$ large enough such that $\log\frac{2M}{\eta}\le C\log\frac{cn}{\delta}$, so by 
$\lambda^{-\alpha}=n^{\frac{\alpha}{2r+\gamma}}$ it suffices that 
\[
n^{\frac{2r+\gamma-\alpha}{2r+\gamma}}\ge CM_\alpha\log\frac{cn}{\delta}.
\]
Since $\alpha<2r+\gamma$, the exponent $\frac{2r+\gamma-\alpha}{2r+\gamma}$ is positive, so we can always solve for $n$ in terms of the other parameters, i.e., $n\geq \widetilde{n}$ for some $\widetilde{n}$ depending only on $M_\alpha$, $\delta$, $r$, $\gamma$, $\alpha$ and the constants $C$ and $c$.
\item[(iii)] Condition $n\lambda^\alpha\ge\frac{M_\alpha}{\eta\log^2\frac{2}{\eta}}$ 
used in the first estimate above, which is equivalent to 
\[
n\ge \left(\frac{10M_\alpha}{\delta\log^2\frac{20}{\delta}}\right)^{\frac{2r+\gamma}{2r+\gamma-\alpha}}.
\]
\end{itemize}
Therefore, all the conditions on $n$ are satisfied whenever 
\[
n\ge n_0=\max\left\{1,\ \|L_K\|^{-(2r+\gamma)},\ 
\widetilde{n},\ 
\left(\frac{10M_\alpha}{\delta\log^2\frac{20}{\delta}}\right)^{\frac{2r+\gamma}{2r+\gamma-\alpha}}\right\}.
\]

We next verify the conditions on the number of random features $M$, i.e., 
\eqref{eq:M1} for $0<r<\frac12$ and \eqref{eq:M2} for $\frac12\le r\le1$ in 
Proposition \ref{prop:fluctuation}. With $\lambda=n^{-\frac{1}{2r+\gamma}}$ and 
$\eta=\frac{\delta}{10}$, we have 
\[
\lambda^{-\zeta}=n^{\frac{\zeta}{2r+\gamma}},\qquad
\lambda^{-\{\zeta+(1+\gamma-\zeta)(2r-1)\}}=n^{\frac{\zeta+(1+\gamma-\zeta)(2r-1)}{2r+\gamma}},
\]
and 
\[
\log\frac{cn}{\lambda\eta}=\log\left(10c\delta^{-1}n^{1+\frac{1}{2r+\gamma}}\right)
\le C\log\frac{cn}{\delta},
\]
Moreover, when $\zeta+\alpha=1$, 
\[
1+\log\frac{1+\|L_K\|}{\lambda}
=1+\log\bigl((1+\|L_K\|)n^{\frac{1}{2r+\gamma}}\bigr).
\]
Therefore, \eqref{eq:M1} and \eqref{eq:M2} are satisfied provided that 
\begin{align*}
M\ge 
\begin{cases}
C_0n^{\frac{\zeta}{2r+\gamma}}\log\frac{c_0n}{\delta}, 
& 0<r<\frac12,\ \zeta+\alpha\ne1,\\[2mm]
C_0n^{\frac{\zeta}{2r+\gamma}}\log\frac{c_0n}{\delta}
\Bigl[1+\log\bigl((1+\|L_K\|)n^{\frac{1}{2r+\gamma}}\bigr)\Bigr], 
& 0<r<\frac12,\ \zeta+\alpha=1,\\[2mm]
C_0n^{\frac{\zeta+(1+\gamma-\zeta)(2r-1)}{2r+\gamma}}\log\frac{c_0n}{\delta}, 
& \frac12\le r\le1,\ \zeta+\alpha\ne1,\\[2mm]
C_0n^{\frac{\zeta+(1+\gamma-\zeta)(2r-1)}{2r+\gamma}}\log\frac{c_0n}{\delta}
\Bigl[1+\log\bigl((1+\|L_K\|)n^{\frac{1}{2r+\gamma}}\bigr)\Bigr], 
& \frac12\le r\le1,\ \zeta+\alpha=1,
\end{cases}
\end{align*}
Thus we complete the proof.

\end{proof}

The proofs of Theorem \ref{thm1} and Corollary \ref{cor1} are the same as that of Theorem \ref{thm2} with $\zeta=1$ and $\zeta=\gamma$. Thus we omit the details.

\subsection{Proof of Theorem \ref{thm3}}

\begin{proof}
We follow a similar proof strategy as in Theorem \ref{thm1}. Plugging Propositions \ref{prop:mergedbudget}, \ref{prop:fluctuation}, and \ref{lem:noiseerror_h} into the error decomposition \eqref{eq:decompbound}, we obtain
\begin{align*}
\bigl\|\widehat f_{\lambda,M}-f_\rho\bigr\|_{L^2(\rho_X)} \le CR\left(
\sqrt{\frac{M_\alpha\lambda^{2r-\alpha}}{n\eta}}+\lambda^r
\right)+ C\left(
\frac{B^{\frac1p}\bigl(M_\alpha\lambda^{-\alpha}\bigr)^{\frac12-\frac1p}{\cal N}(\lambda)^{\frac1p}}{\eta^{\frac1p}n^{1-\frac1p}}
+\sigma\sqrt{\frac{{\cal N}(\lambda)\log\frac{c}{\eta}}{n}}
\right)
\end{align*}  
with probability at least $1-10\eta$.

It remains to control the additional term arising from the heavy-tailed noise. By Assumption \ref{ass4} that ${\cal N}(\lambda)\le Q^2\lambda^{-\gamma}$ and 
$\frac1p>0$, we have ${\cal N}(\lambda)^{\frac1p}\le Q^{\frac2p}\lambda^{-\frac\gamma p}$. 
Moreover, $\lambda=n^{-\frac{1}{2r+\gamma}}$ gives $\frac{1}{n^{1-\frac1p}}=n^{-\frac{p-1}{p}}
=\lambda^{\frac{(2r+\gamma)(p-1)}{p}}$, ignoring the constant factors, we have
\[
\begin{aligned}
\frac{\bigl(\lambda^{-\alpha}\bigr)^{\frac12-\frac1p}
{\cal N}(\lambda)^{\frac1p}}{\eta^{\frac1p}n^{1-\frac1p}}
\le \lambda^{-\frac{\alpha(p-2)}{2p}-\frac\gamma p+\frac{(2r+\gamma)(p-1)}{p}}\eta^{-\frac1p}=
\lambda^{r+\frac{(p-2)(2r+2\gamma-\alpha)}{2p}}
\eta^{-\frac1p}\le \lambda^r,
\end{aligned}
\]
where the equality uses 
\[
-\frac{\alpha(p-2)}{2p}-\frac\gamma p+\frac{(2r+\gamma)(p-1)}{p}-r
=\frac{(p-2)(2r+2\gamma-\alpha)}{2p},
\]
and the last inequality holds provided that 
\[
\lambda^{\frac{(p-2)(2r+2\gamma-\alpha)}{2p}}\eta^{-\frac1p}\le 1.
\]
Since $p>2$ and $2r+2\gamma-\alpha>2r+\gamma-\alpha>0$, the exponent 
$\frac{(p-2)(2r+2\gamma-\alpha)}{2p}$ is positive. Substituting 
$\lambda=n^{-\frac{1}{2r+\gamma}}$, the above condition is equivalent to 
\begin{equation}\label{eq:heavybudget}
n\ge \left(\eta^{-\frac1p}\right)^{\frac{2p(2r+\gamma)}{(p-2)(2r+2\gamma-\alpha)}}=\left(\frac{10}{\delta}\right)^{\frac{2(2r+\gamma)}{(p-2)(2r+2\gamma-\alpha)}}.
\end{equation}

Under the following condition on the sample size $n$, i.e.,
$$
n\ge n_1=\max\left\{1,\ \|L_K\|^{-(2r+\gamma)},\ 
\widetilde{n},\ 
\left(\frac{10M_\alpha}{\delta\log^2\frac{20}{\delta}}\right)^{\frac{2r+\gamma}{2r+\gamma-\alpha}},\ \left(\frac{10}{\delta}\right)^{\frac{2(2r+\gamma)}{(p-2)(2r+2\gamma-\alpha)}}\right\}\geq n_0.
$$ 
we can directly control the additional term arising from the heavy-tailed noise. Other terms in the error decomposition can be handled using techniques for light-tailed noise. Thus we can obtain the same error bound as in the light-tailed case with some different constants. 

\end{proof}

The proofs of Corollaries \ref{cor2} and \ref{cor3} are the same as that of Theorem \ref{thm3} with $\zeta=1$ and $\zeta=\gamma$. Thus we omit the details.

\section{Concentration Inequalities in Hilbert Spaces}\label{sec:concentration}

In this section, we give two concentration inequalities for sums of independent random variables taking values in a separable Hilbert space. They serve to bound the noise term $\left\|C_{M,\lambda}^{-1/2}\widehat{S}_{M}^*\widehat{\boldsymbol{\varepsilon}}\right\|_2$ under two different moment conditions on the noise. The first one, a Bennett inequality, applies to light-tailed noise and gives a bound with logarithmic dependence on the confidence level $\eta$. The second one, a Fuk-Nagaev inequality, only requires finite moments of order $p\ge3$ and is therefore suited to heavy-tailed noise. Its bound contains an additional polynomial term in $\eta$.

\subsection{Concentration Inequalities in Hilbert Spaces under Light-tailed Conditions}

The following Bennett inequality for vector-valued random variables is from \cite{smale2007learning}, and it can be derived from Theorem 3.4 in \cite{pinelis1994optimum} and the elementary inequality $t\log(1+t)\geq 2t-2\log(1+t)$ for any $t>0$.

\begin{lemma}[Bennett inequality for sum of random vectors]\label{lem:bennett}
Let $z_1,\ldots,z_n$ be a sequence of i.i.d. random variables on a separable Hilbert space $\cal H$. If there exist $\widetilde{\sigma},\widetilde{B} \geq 0$ such that
\begin{align}\label{eq:bennettmoment}
\mathbb{E}\|z-\mathbb{E}[z]\|_{\cal H}^p\leq \frac{1}{2}p!\widetilde{B}^{p-2}\widetilde{\sigma}^{2}, \quad \forall p\geq 2,
\end{align}
then for any $\eta \in (0,1)$, there holds
\[
\left\|\frac{1}{n}\sum_{i=1}^nz_i-\mathbb{E}z_i\right\|_{\cal H} \leq \frac{2\widetilde{B}\log (2/\eta)}{n}+\sqrt{\frac{2\widetilde{\sigma}^2\log (2/\eta)}{n}},
\]
with probability at least $1-\eta$. In particular, \eqref{eq:bennettmoment} holds if
\[
\|z\|_{\cal H}\leq \widetilde{B}/2, \text{ a.s.}\quad \text{and}\quad \mathbb{E}\|z\|_{\cal H}^2 \leq \widetilde{\sigma}^2.
\]
\end{lemma}

This lemma relies on a Bernstein-type moment condition \eqref{eq:bennettmoment} on the random variables. A sufficient condition for \eqref{eq:bennettmoment} is that the variables are almost surely bounded or have light-tailed distributions (e.g., Gaussian or sub-Gaussian). In our analysis it is applied to the whitened noise vectors $C_{M,\lambda}^{-1/2}{\phi}_{M}(\bfx_i)\widehat{\boldsymbol{\varepsilon}}$ and yields the high-probability bound for the noise term in Proposition \ref{lem:noiseerror}, which is one of the three error terms in the decomposition of Proposition \ref{lem:decomp}.

\subsection{Concentration Inequalities in Hilbert Spaces under Heavy-tailed Conditions}

The following Fuk-Nagaev inequality for vector-valued random variables is from \cite{yurinsky2006sums} and \cite{mollenhauer2025regularized}, which is a generalization of the classical real-valued Fuk-Nagaev inequality \citep{dkh1973some,nagaev1979large} to separable Hilbert spaces.

\begin{lemma}[Fuk-Nagaev inequality for sum of random vectors]\label{lem:fuknagaev}
Let $z_1,\ldots,z_n$ be a sequence of i.i.d. random variables on a separable Hilbert space $\cal H$, such that
\[
\mathbb{E}[z]=0,\quad \mathbb{E}[\|z\|_{\cal H}^2]\leq \widetilde{\sigma}^2, \quad \text{and} \quad\mathbb{E}[\|z\|_{\cal H}^p]\leq \widetilde{B},
\]
for some constants $\widetilde{\sigma}, \widetilde{B}> 0$ and $p\in \mathbb{N}$, $p\geq 3$. Then there exist some constants $c_p, c_p^{\prime}>0$ depending on $p$ such that for any $t>0$, we have
\begin{align}\label{eq:fuknagaev}
\mathbb{P}\left[\left\|\frac{1}{n}\sum_{i=1}^nz_i\right\|_{\cal H}>t\right]\leq c_p\left(\frac{\widetilde{B}}{t^p n^{p-1}}+\exp\left(-c_p^{\prime}\frac{t^2n}{\widetilde{\sigma}^2}\right)\right).
\end{align}
Equivalently, for any $\eta\in(0,1)$, we have
\begin{align}\label{eq:fuknagaevprob}
\left\|\frac{1}{n}\sum_{i=1}^nz_i\right\|_{\cal H}\leq \left(\frac{2c_p\widetilde{B}}{\eta n^{p-1}}\right)^{1/p}+\widetilde{\sigma}\sqrt{\frac{\log(2c_p/\eta)}{c_{p}^{\prime}n}}
\end{align}
with probability at least $1-\eta$.
\end{lemma}

Compared to Lemma \ref{lem:bennett}, this result assumes only finite $p$-th moments for some $p \geq 3$, and thus can be applied to heavy-tailed random variables. It consists of two terms on the right-hand side of \eqref{eq:fuknagaevprob}: a polynomial term (in $n$ and $\eta$), capturing heavy tails, and a sub-Gaussian term with logarithmic dependence on $1/\eta$, analogous to the Bennett-type bound in Lemma \ref{lem:bennett}. We use this inequality in place of Lemma \ref{lem:bennett} to control the same noise term under the heavy-tail assumption, which gives the noise bound in  Proposition \ref{lem:noiseerror_h}. Furthermore, once the sample size $n$ is sufficiently large, the polynomial term is dominated by the exponential terms, so the estimator attains the same rate as in the light-tailed case. For completeness, we give the proof of this result, following \cite{yurinsky2006sums} and \cite{mollenhauer2025regularized}. Throughout,
$C_p,c_p,c'_p$ denote positive constants depending only on the
fixed integer $p\geq 3$, their values may change from line to
line.

\begin{proof}
We first prove that, for every $x>0$,
\begin{equation}\label{eq:fn_sum_tail_proof}
 \mathbb P\Bigl\{\Bigl\|\sum_{i=1}^n z_i\Bigr\|_{\mathcal H}>x\Bigr\}
 \leq C_p\left(\frac{n\widetilde B}{x^p}
       +\exp\!\left[-c'_p\frac{x^2}{n\widetilde\sigma^2}\right]\right).
\end{equation}

\paragraph{A centered bounded-sum inequality.}
Let $Y_1,\ldots,Y_n$ be independent, centered,
$\mathcal H$-valued random variables satisfying
$\|Y_i\|_{\mathcal H}\leq L$ almost surely,
$\sum_i\mathbb E\|Y_i\|_{\mathcal H}^2\leq n\widetilde\sigma^2$,
and $\sum_i\mathbb E\|Y_i\|_{\mathcal H}^p\leq 2^pn\widetilde B$.
The centered Hilbert-space cosh inequality (Pinelis, 1994; see
also Mollenhauer et al., 2025, Appendix D.3.1) gives, for $h>0$,
\begin{align*}
 \mathbb E\cosh\!\left(h\Bigl\|\sum_iY_i\Bigr\|_{\mathcal H}\right)
 \leq\prod_i \mathbb E\Bigl[e^{h\|Y_i\|_{\mathcal H}}
                             -h\|Y_i\|_{\mathcal H}\Bigr]
 \leq\exp\!\left(\sum_i\mathbb E\Bigl[e^{h\|Y_i\|_{\mathcal H}}
      -1-h\|Y_i\|_{\mathcal H}\Bigr]\right).
\end{align*}
For integers $2\leq k\leq p$, moment interpolation together
with H\"older's inequality on the disjoint union of the
probability spaces yields
\[
 \sum_i\mathbb E\|Y_i\|_{\mathcal H}^k
 \leq\bigl(n\widetilde\sigma^2\bigr)^{\frac{p-k}{p-2}}
    \bigl(2^pn\widetilde B\bigr)^{\frac{k-2}{p-2}}.
\]
Consequently, by the weighted arithmetic--geometric mean
inequality, the contribution of the $k$th power-series term is
bounded by $C_p(h^2n\widetilde\sigma^2+h^pn\widetilde B)$. For
$k>p$, the bound $\|Y_i\|_{\mathcal H}\leq L$ gives
$\sum_i\mathbb E\|Y_i\|_{\mathcal H}^k\leq L^{k-p}2^pn\widetilde B$.
Summing the exponential series shows
\begin{equation}\label{eq:fn_cosh_bound_proof}
 \mathbb E\cosh\!\left(h\Bigl\|\sum_iY_i\Bigr\|_{\mathcal H}\right)
 \leq\exp\!\left(C_ph^2n\widetilde\sigma^2
                +C_ph^pn\widetilde Be^{hL}\right),
\end{equation}
where the constants $C_p$ are independent of $n$,
$\widetilde\sigma^2$, $\widetilde B$, and $L$. Since
$e^u\leq 2\cosh(u)$ for $u\geq 0$, Markov's inequality applied
to \eqref{eq:fn_cosh_bound_proof} gives
\begin{equation}\label{eq:fn_chernoff_proof}
 \mathbb P\left\{\Bigl\|\sum_iY_i\Bigr\|_{\mathcal H}>u\right\}
 \leq 2\exp\!\left(-hu+C_ph^2n\widetilde\sigma^2
             +C_ph^pn\widetilde Be^{hL}\right)
 \qquad (u,h>0).
\end{equation}

\paragraph{Truncation and correction of the mean.}
Fix $x>0$ and set $\Delta=n\widetilde B/x^p$. Choose
$K\geq 16$, depending only on $p$ and to be specified below,
and truncate at $L=x/K$:
\[
 \widetilde z_i=z_i\mathbf 1_{\{\|z_i\|_{\mathcal H}\leq L\}},
 \qquad Y_i=\widetilde z_i-\mathbb E\widetilde z_i.
\]
In contrast to $\widetilde z_i$, the variables $Y_i$ are
centered. By Markov's inequality and a union bound,
\begin{equation}\label{eq:fn_big_jump_proof}
 \mathbb P\Bigl\{\sum_{i=1}^n z_i\neq
       \textstyle\sum_i\widetilde z_i\Bigr\}
 \leq\sum_i\frac{\mathbb E\|z_i\|_{\mathcal H}^p}{L^p}
 \leq K^p\Delta.
\end{equation}
Since $\mathbb Ez_i=0$,
\begin{equation}\label{eq:fn_mean_correction_proof}
 \Bigl\|\sum_i\mathbb E\widetilde z_i\Bigr\|_{\mathcal H}
 \leq \sum_i\mathbb E\Bigl[\|z_i\|_{\mathcal H}
      \mathbf 1_{\{\|z_i\|_{\mathcal H}>L\}}\Bigr]
 \leq\frac{n\widetilde B}{L^{p-1}}=K^{p-1}\Delta x.
\end{equation}
Furthermore, $\|Y_i\|_{\mathcal H}\leq 2L$,
\[
 \sum_i\mathbb E\|Y_i\|_{\mathcal H}^2\leq n\widetilde\sigma^2,
 \qquad
 \sum_i\mathbb E\|Y_i\|_{\mathcal H}^p\leq 2^pn\widetilde B.
\]
The first inequality follows from the Hilbert-space variance
identity $\mathbb E\|Y-\mathbb EY\|_{\mathcal H}^2
=\mathbb E\|Y\|_{\mathcal H}^2-\|\mathbb EY\|_{\mathcal H}^2$;
the second follows from Jensen's inequality and
$(a+b)^p\leq 2^{p-1}(a^p+b^p)$.

Choose a fixed $\Delta_0\leq\min\{e^{-1},(4K^{p-1})^{-1}\}$.
If $\Delta\geq\Delta_0$, then \eqref{eq:fn_sum_tail_proof}
follows from the trivial bound
$\mathbb P\bigl\{\bigl\|\sum_i z_i\bigr\|_{\mathcal H}>x\bigr\}
\leq 1\leq\Delta/\Delta_0$
after enlarging $C_p$. We may thus assume $\Delta<\Delta_0$.
Then the right-hand side of \eqref{eq:fn_mean_correction_proof}
is at most $x/4$. Combining \eqref{eq:fn_big_jump_proof} with
\eqref{eq:fn_chernoff_proof}, applied at $u=3x/4$ and with the
truncation level $2L$, yields, for every $h>0$,
\begin{equation}\label{eq:fn_two_parameter_proof}
 \mathbb P\Bigl\{\Bigl\|\sum_{i=1}^n z_i\Bigr\|_{\mathcal H}>x\Bigr\}
 \leq K^p\Delta+
 2\exp\!\left(-\frac34hx+C_ph^2n\widetilde\sigma^2
              +C_ph^pn\widetilde Be^{2hx/K}\right).
\end{equation}

\paragraph{Optimization in two deviation regimes.}
Write $H=x^2/(n\widetilde\sigma^2)$ and
$G=\log(1/\Delta)\geq 1$. We take $D\geq 64C_p$ and then choose
$K$ (which may be enlarged) so that $K\geq 16$ and
$K\geq 4aD$, where $a=\min\{1,3/(16C_p)\}$.

If $H\leq DG$, choose $h=ax/(n\widetilde\sigma^2)$. Then
\[
 -\tfrac34hx+C_ph^2n\widetilde\sigma^2\leq-\tfrac9{16}aH,
 \qquad
 C_ph^pn\widetilde Be^{2hx/K}
 \leq C_pa^pD^pG^pe^{-G/2}\leq C'_p,
\]
where the last bound is finite since
$\sup_{G\geq1}G^pe^{-G/2}<\infty$. Thus
\eqref{eq:fn_two_parameter_proof} gives
\[
 \mathbb P\Bigl\{\Bigl\|\sum_{i=1}^n z_i\Bigr\|_{\mathcal H}>x\Bigr\}
 \leq K^p\Delta+C''_pe^{-c'_pH}.
\]

If $H>DG$, choose $h=4G/x$. Then
\[
 -\tfrac34hx=-3G,\qquad
 C_ph^2n\widetilde\sigma^2\leq\frac{16C_p}{D}G\leq\frac14G,
\]
and, since $K\geq 16$,
\[
 C_ph^pn\widetilde Be^{2hx/K}
 =C_p 4^p G^p \Delta e^{8G/K}
 \leq C_p4^pG^pe^{-G/2}\leq C''_p.
\]
Consequently, the exponential term in
\eqref{eq:fn_two_parameter_proof} is at most
$C_pe^{-11G/4}\leq C_p\Delta$, and we again obtain
\[
 \mathbb P\Bigl\{\Bigl\|\sum_{i=1}^n z_i\Bigr\|_{\mathcal H}>x\Bigr\}
 \leq C_p\left(\Delta+e^{-c'_pH}\right).
\]
Combining the two regimes proves \eqref{eq:fn_sum_tail_proof}.

Finally, take $x=nt$. Then
$n\widetilde B/x^p=\widetilde B/(n^{p-1}t^p)$ and
$x^2/(n\widetilde\sigma^2)=nt^2/\widetilde\sigma^2$, which
proves \eqref{eq:fuknagaev} after renaming the constants and
taking $c_p\geq 1$.

For the confidence bound, set
\[
 u_1=\left(\frac{2c_p\widetilde B}{\eta n^{p-1}}\right)^{\!1/p},
 \qquad
 u_2=\widetilde\sigma\sqrt{\frac{\log(2c_p/\eta)}{c'_pn}},
 \qquad u=\max\{u_1,u_2\}.
\]
Evaluated at $t=u$, each of the two terms on the right-hand
side of \eqref{eq:fuknagaev} is at most $\eta/2$. Thus
$\bigl\|n^{-1}\sum_{i=1}^n z_i\bigr\|_{\mathcal H}\leq u$ with
probability at least $1-\eta$. Since $u\leq u_1+u_2$, this
proves \eqref{eq:fuknagaevprob}.
\end{proof}

\section{Embedding Condition}\label{app:embedding_condition}

\begin{lemma}\label{lem:embed}
Under Assumption \ref{ass5}, then for all $\lambda >0$ and $\alpha_0<\alpha\le 1$, the following maximal capacity condition holds
$$
{\cal N}_{\infty}(\lambda) = \operatorname*{esssup}_{x}
\sum_j\frac{\mu_j e_j(\bfx)^2}{\mu_j+\lambda} \leq M_{\alpha}\lambda^{-\alpha}.
$$
\end{lemma}

\begin{proof}
Recall the definition of the embedding condition \eqref{embedding_property}, which states that there exists a constant $M_\alpha$ for any $\alpha_0<\alpha$ such that
$$
\sum_j \mu_j^{\alpha} e_j(\bfx)^2 \leq M_\alpha, \quad \text{for $\rho_X$-almost every $x$}.
$$
Without loss of generality, we assume that the constant $M_\alpha\ge 1$, since we can replace it with $\max\{M_\alpha,1\}$ if necessary. Then, for any $\lambda > 0$, we have
$$
\sum_j\frac{\mu_j e_j(\bfx)^2}{\mu_j+\lambda} \leq \sum_j \frac{\mu_j^{1-\alpha} \mu_j^{\alpha} e_j(\bfx)^2}{\mu_j+\lambda} \leq \sum_j \mu_j^{\alpha} e_j(\bfx)^2 \sup_j \frac{\mu_j^{1-\alpha}}{\mu_j+\lambda} \leq M_\alpha \lambda^{-\alpha},
$$
where the last inequality follows from $\sup_{t\ge 0} \frac{t^{s}}{t+\lambda} \leq \lambda^{s-1}$ for $0\le s \le 1$.

\end{proof}

\section{Useful Lemmas}

\begin{lemma}[Scalar Bernstein inequality]\label{lem:sbern}
Let $U_1,\dots,U_n$ be independent real-valued random variables such that,
\[
0\leq U_i\leq b,\qquad \mathbb E U_i\leq m,\qquad
\operatorname{Var}(U_i)\leq v
\]
almost surely for some constants $b,m,v>0$. Then for every $u>0$,
\begin{equation}\label{eq:sbern}
\mathbb P\Bigl(\frac1n\sum_{i=1}^n\bigl(U_i-\mathbb E U_i\bigr)>u\Bigr)
\leq\exp\Bigl[-\frac{nu^2}{2\bigl(v+bu/3\bigr)}\Bigr].
\end{equation}
\end{lemma}

\begin{lemma}[Matrix Bernstein inequality]\label{lem:mbern}
Let $Z_1,\dots,Z_n \in \mathbb{R}^{d\times d}$ be independent self-adjoint random matrices such that, 
\[
0 \le \|Z_i\|\le B,
\qquad \mathbb E Z_i=0,
\qquad
\Bigl\|\frac1n\sum_{i=1}^n\mathbb E\bigl[Z_i^{2}\bigr]\Bigr\|\le V
\]
almost surely for some constants $B,V>0$, where $\|\cdot\|$ denotes the operator norm.
Then for every $U>0$,
\begin{equation}\label{eq:mbern}
\mathbb P\Bigl(\Bigl\|\frac1n\sum_{i=1}^nZ_i\Bigr\|>U\Bigr)
\le2d\exp\Bigl[-\frac{nU^{2}}{2\bigl(V+BU/3\bigr)}\Bigr].
\end{equation}
\end{lemma}

\vskip 0.2in
\bibliography{sample}

\end{document}